%% file: acl_latex.tex
\pdfoutput=1

\documentclass[11pt]{article}

\usepackage[]{acl}

\usepackage{times}
\usepackage{latexsym}

\usepackage[T1]{fontenc}

\usepackage[utf8]{inputenc}

\usepackage{microtype}

\usepackage{inconsolata}

\usepackage{graphicx}

\usepackage{amsmath}
\usepackage{amssymb}
\usepackage{multirow}
\usepackage{array}
\usepackage{booktabs}
\usepackage{CJKutf8}
\usepackage{makecell}
\usepackage{subfigure}
\usepackage{float}
\usepackage{enumitem}
\usepackage{verbatim}
\usepackage{xcolor}
\usepackage[figuresright]{rotating}
\newenvironment{metaverbatim}{\verbatim}{\endverbatim}
\newcommand{\revisedText}[1]{{}{#1}}
\newcommand{{\Name}}{CodeMixBench}
\title{CodeMixBench: Evaluating Code-Mixing Capabilities of LLMs\\ Across 18 Languages}

\author{Yilun Yang \\
  NUC\\
  \texttt{jeromeyluck@gmail.com} \\\And
  Yekun Chai\thanks{Corresponding author.} \\
  ETH Zurich \\
  \texttt{yechai@ethz.ch} }

\begin{document}
\begin{CJK}{UTF8}{gkai}
\maketitle
\begin{abstract}
Code-mixing, the practice of switching between languages within a conversation, poses unique challenges for traditional NLP. 
Existing benchmarks are limited by their narrow language pairs and tasks, failing to adequately assess large language models' (LLMs) code-mixing abilities. Despite the recognized importance of code-mixing for multilingual users, research on LLMs in this context remains sparse. Additionally, current techniques for synthesizing code-mixed data are underdeveloped to generate code-mixing. In response, we introduce {\Name}, a comprehensive benchmark covering eight tasks, including three specific to LLMs and five traditional NLP tasks, and 18 languages across seven language families. We also propose a new method for generating large-scale synthetic code-mixed texts by combining word substitution with GPT-4 prompting. Our evaluation reveals consistent underperformance of LLMs on code-mixed datasets involving different language families. Enhancements in training data size, model scale, and few-shot learning could improve their performance. 
The code and dataset are available at \url{https://github.com/Jeromeyluck/CodeMixBench}.
\end{abstract}

\section{Introduction}
\label{ref:introduction}
\input{section/introduction}

\section{Related Work}
\label{ref:related_work}
\input{section/related_work}

\section{CodeMixBench}
\label{ref:CodeMixBench}
\input{section/CodeMixBench}

\section{Experiments}
\label{ref:experiments}
\input{section/experiment}

-
\section{Conclusion}
\label{ref:conclusion}
\input{section/conclusion}

\section*{Limitations}
We introduce {\Name}, a collection of 22 synthetic datasets and 30 open-source datasets, each with potential quality issues. Our synthesis method generates large-scale code-mixed datasets with detailed filtering, but unexpected quality problems can still occur. Furthermore, our benchmark includes 18 languages, making it challenging to maintain consistent quality control. Furthermore, it could be promising to evaluate and mitigate the potential bias in code-mixing scenarios~\cite{ravfogel2020null,peng2025debiasing}.

\section*{Acknowledgments}
We would like to thank all anonymous reviewers for
their insightful comments and feedback.

\bibliography{custom}

\appendix
\input{section/appendix}

\end{CJK}
\end{document}

%% file: section/introduction.tex
Code-mixing is a linguistic phenomenon where multilingual speakers switch or mix two or more languages within a single utterance or conversation. This typically occurs due to a lack of suitable vocabulary or expressions in one language,  the presence of untranslatable terms, or contextual factors such as interlocutors, situational context, messages, attitudes, and emotions \citep{kim2006reasons}. With the global rise of social media,  there has been a substantial increase in code-mixed content \citep{rijhwaniEstimatingCodeSwitchingTwitter2017}, prompting extensive interest from linguists and NLP researchers \citep{winataDecadesProgressCodeSwitching2023}. However, several key issues remain unresolved.

Existing studies are difficult to compare directly because they focus on different downstream tasks and language pairs. To address this issue, LinCE \citep{aguilarLinCECentralizedBenchmark2020} and GLUECoS \citep{khanujaGLUECoSEvaluationBenchmark2020} introduced two benchmarks, but they only cover a limited number of language pairs and traditional NLP tasks. LinCE addresses four language pairs and five traditional NLP tasks, including language identification (LID), part-of-speech tagging (POS), named entity recognition (NER), sentiment analysis (SA), and machine translation (MT), while GLUECoS covers only two language pairs and six traditional NLP tasks, \emph{i.e.}, LID, POS, NER, SA, question answering (QA), and natural language inference (NLI). These traditional NLP tasks are insufficient to evaluate LLM performance comprehensively. 

Despite strong multilingual performance on various benchmarks, LLMs' capabilities with code-mixing remain underexplored. Limited studies suggest that LLMs often perform worse than smaller, fine-tuned models on code-mixing tasks \citep{zhangMultilingualLargeLanguage2023a}, and multilingual users prefer chatbots that handle code-mixing well \citep{BawaMultilingualPreferCode-mix}. Thus, incorporating code-mixing into LLM evaluation is crucial.

Creating new code-mixed datasets for LLMs involves using synthesis techniques. Some studies \cite{bhatGrammaticalConstraintsIntrasentential2016, pratapaLanguageModelingCodeMixing2018b} focused on generating synthetic code-mixed data to solve the scarcity of code-mixed data, using methods based on the Equivalence Constraint theory \citep{POPLACK+1980+581+618}, a linguistic theory that restricts the occurrences of code-mixing. However, the quality of these outputs heavily depends on the performance of word alignment and syntactic parsing tools. 
Recent efforts to generate code-mixed text using data-driven models still face challenges related to dataset size, quality, or linguistic diversity \citep{yangAlternatingLanguageModeling2020, hsuCodeSwitchedTextSynthesis2023}. Also, initial attempts to use LLMs for generating code-mixed data did not fully leverage their instruction-following capabilities \citep{yongPromptingMultilingualLarge2023a}.

In response to these issues, we introduce the {\Name}, a code-mixing evaluation benchmark including eight tasks--three for evaluating LLMs (knowledge reasoning, mathematical reasoning, and truthfulness) and five for traditional NLP tasks (LID, POS, NER, SA, and MT). They span 18 languages from seven language families, covering high-resource, medium-resource, and low-resource languages. 
\revisedText{Our benchmark largely expands language pair and task coverage compared to LinCE and GLUECoS (Appendix \ref{app:versus}).}
We also propose a novel synthetic code-mixing approach using word substitution within GPT-4 prompting to generate large-scale code-mixed texts from parallel corpora.

Our contributions are summarized as follows:
\begin{enumerate}[noitemsep, left=0pt, labelsep=4pt, topsep=1pt, partopsep=0pt]
    \item We present {\Name}, the first comprehensive benchmark for evaluating the performance of LLMs on multilingual code-mixing. We have synthesized 22 datasets and, through extensive research, compiled 30 open-source code-mixed datasets to integrate into our benchmark. In total, the benchmark encompasses eight tasks and 18 languages from seven language families (\S \ref{ref:CodeMixBench}).
    \item We propose a novel pipeline for large-scale synthesis of multilingual code-mixing data, integrating word substitution with LLM prompts for the first time. The synthetic results validate the efficiency of our approach in generating substantial multilingual code-mixed data. (\S \ref{sec:parallel corpus}). 
    \item We evaluate three families of LLMs on {\Name}, revealing consistent underperformance across all models on code-mixing datasets involving language pairs from different language families. However, enhancements in training data size, model scale, post-training, and few-shot learning can improve LLM performance on code-mixing datasets (\S \ref{ref:experiments}).
\end{enumerate}

%% file: section/related_work.tex
\paragraph{Code-Mixing Challenge}
Early research employed linguistic rules and statistical methods \citep{liCodeSwitchLanguageModel2012b, liLanguageModelingFunctional2014, bhatGrammaticalConstraintsIntrasentential2016, rijhwaniEstimatingCodeSwitchingTwitter2017} for code-mixing modeling. Subsequently, research shifted towards neural network models like RNNs and LSTMs \citep{adelCombinationRecurrentNeural2013, adelRecurrentNeuralNetwork2013, wangCodeSwitchedNamedEntity2018, winataBilingualCharacterRepresentation2018}, and more recently towards pre-trained language models such as mBERT and XLM-R \citep{winataAreMultilingualModels2021a, malmasiMultiCoNERLargescaleMultilingual2022a, perezRoBERTuitoPretrainedLanguage2022}. These methodologies have been applied to various code-mixing-related downstream tasks, including language identification \citep{solorioOverviewFirstShared2014, molinaOverviewSecondShared2016}, named entity recognition \citep{aguilarNamedEntityRecognition2018a}, part-of-speech tagging \citep{singhTwitterCorpusHindiEnglish2018, sotoJointPartofSpeechLanguage2018}, sentiment analysis \citep{patraSentimentAnalysisCodeMixed2018, patwaSemEval2020TaskOverview2020a}, machine translation \citep{srivastavaPHINCParallelHinglish2020, chenCALCS2021Shared2022}, natural language inference \citep{khanujaNewDatasetNatural2020}, question answering  \citep{chanduLanguageInformedModeling2018a}, and multilingual code generation~\cite{chai-etal-2023-ernie, peng-etal-2024-humaneval-xl}. Benchmarks, such as GLUECoS \cite{khanujaGLUECoSEvaluationBenchmark2020} and LinCE \cite{aguilarLinCECentralizedBenchmark2020} primarily focus on traditional NLP tasks and are restricted to a limited number of languages.
Recent research by \citet{zhangMultilingualLargeLanguage2023a} on the performance of multilingual LLMs in code-switching contexts indicates that, despite their strong capabilities across various monolingual tasks, they still yield inferior performance compared to fine-tuned smaller models.

\paragraph{Synthesis of Code-Mixed Data} 
Early research synthesized code-mixed data based on linguistic rules. Following the EC theory, \cite{bhatGrammaticalConstraintsIntrasentential2016, pratapaLanguageModelingCodeMixing2018b} utilized word alignment tools and syntactic parsers to enable the structural substitution and integration of lexical elements within aligned parse trees. 
Subsequently, researchers trained generative models to produce code-mixed data, such as a sequence-to-sequence model with a Pointer-Generator \citep{winataCodeSwitchedLanguageModels2019, guptaSemisupervisedApproachGenerate2020}, Generative Adversarial Networks \citep{changCodeSwitchingSentenceGeneration2019,cclgan21, moegan23}, and Variational AutoEncoders \citep{samantaDeepGenerativeModel2019}. 
An increasing number of works \citep{samantaImprovedSentimentDetection2019, yangAlternatingLanguageModeling2020, aroraCoMixGuideTransformers2023a, hsuCodeSwitchedTextSynthesis2023} focused on extending pre-trained models for code-mixed data generation. 
\citet{yongPromptingMultilingualLarge2023a} examined the ability of LLMs to generate code-mixed text in Southeast Asian languages. 
Instead of using LLMs to directly generate code-mixed text, we revisit the EC theory and integrate its core principles into the prompt. Based on parallel corpora, we instruct the LLM to replace lexical elements between parallel sentences, thereby generating grammatically coherent code-mixed text.

%% file: section/CodeMixBench.tex
\input{tables/languages}
\subsection{Overview}
To evaluate LLMs' comprehension of multilingual code-mixed texts, we introduce {\Name}, a benchmark comprising eight tasks across 18 languages. Table~\ref{tab:languages} details the speaker population and resource ratio on CommonCrawl\footnote{https://commoncrawl.github.io/cc-crawl-statistics/plots/languages.html} for each language, identified by their ISO 639 codes. The chosen languages exhibit diversity in language families, resource availability, and speaker populations. Motivated by \citet{bang-etal-2023-multitask,lai-etal-2023-chatgpt,laiOkapiInstructiontunedLarge2023a}, five languages (zh, es, fr, de, nl) are categorized as high-resource (\textit{CC} >1\%), three (ar, hi, bn) as mid-resource (0.1\%-1\%), and four (mr, ne, ta, ml) as low-resource (<0.01\%). 

Our benchmark comprises synthesized datasets targeting knowledge reasoning, mathematical reasoning, and truthfulness tasks, along with LID, POS, NER, SA, and MT tasks, which have been adapted from open-source studies. For knowledge reasoning, we developed the code-mixed MMLU (CM-MMLU) based on the MMLU test set \citep{hendrycksMeasuringMassiveMultitask2021}, featuring multiple-choice questions from 57 subjects to assess the model's comprehensive knowledge reasoning abilities. For mathematical reasoning, we created the code-mixed GSM8K (CM-GSM8K), derived from the GSM8K test set \citep{cobbeTrainingVerifiersSolve2021}, which evaluates mathematical reasoning capabilities with each question including step-by-step solutions. For truthfulness assessment, we constructed the code-mixed TruthfulQA (CM-TruthfulQA) using 817 multiple-choice questions from the TruthfulQA test set \citep{linTruthfulQAMeasuringHow2022}. Details of the collected datasets are provided in Appendix \ref{app:collected}.

Figure~\ref{fig:pipeline} demonstrates the entire process of constructing our synthetic dataset, including a real example. The original datasets undergo three phases to be transformed into code-mixed datasets: First, collecting existing multilingual parallel corpora or constructing them via translation (detailed in Section \ref{sec:parallel corpus}). Second, instructing GPT to generate code-mixed datasets in various language pairs based on the parallel corpus (detailed in Section \ref{sec:synthesis}). Third, evaluating and filtering the synthetic dataset at word-level, semantic-level, and human-level (detailed in Section \ref{sec:evaluation}). We finally synthesized 11 code-mixed language pairs for CM-MMLU with 12,156 question-option-answer combinations, 4 pairs for CM-TruthfulQA with 3,122 multiple-choice instances, 4 pairs for CM-GSM8K with 4,367 math problems, and 3 pairs for MT with 2,711 code-mixed sentences. The datasets encompass 12 languages from six families: Germanic (en, de, nl), Romance (es, fr), Sino-Tibetan (zh), Afro-Asiatic (ar), Indo-Aryan (hi, bn, mr, ne), and Dravidian (ta). Linguistic diversity enables assessing the impact of multilingual code-switching on model performance. Detailed statistics for the synthetic datasets are provided in Appendix \ref{app:statistics}.

\begin{figure*}[t]
\vspace{-2em}
\centering
   \includegraphics[width=0.95\textwidth]{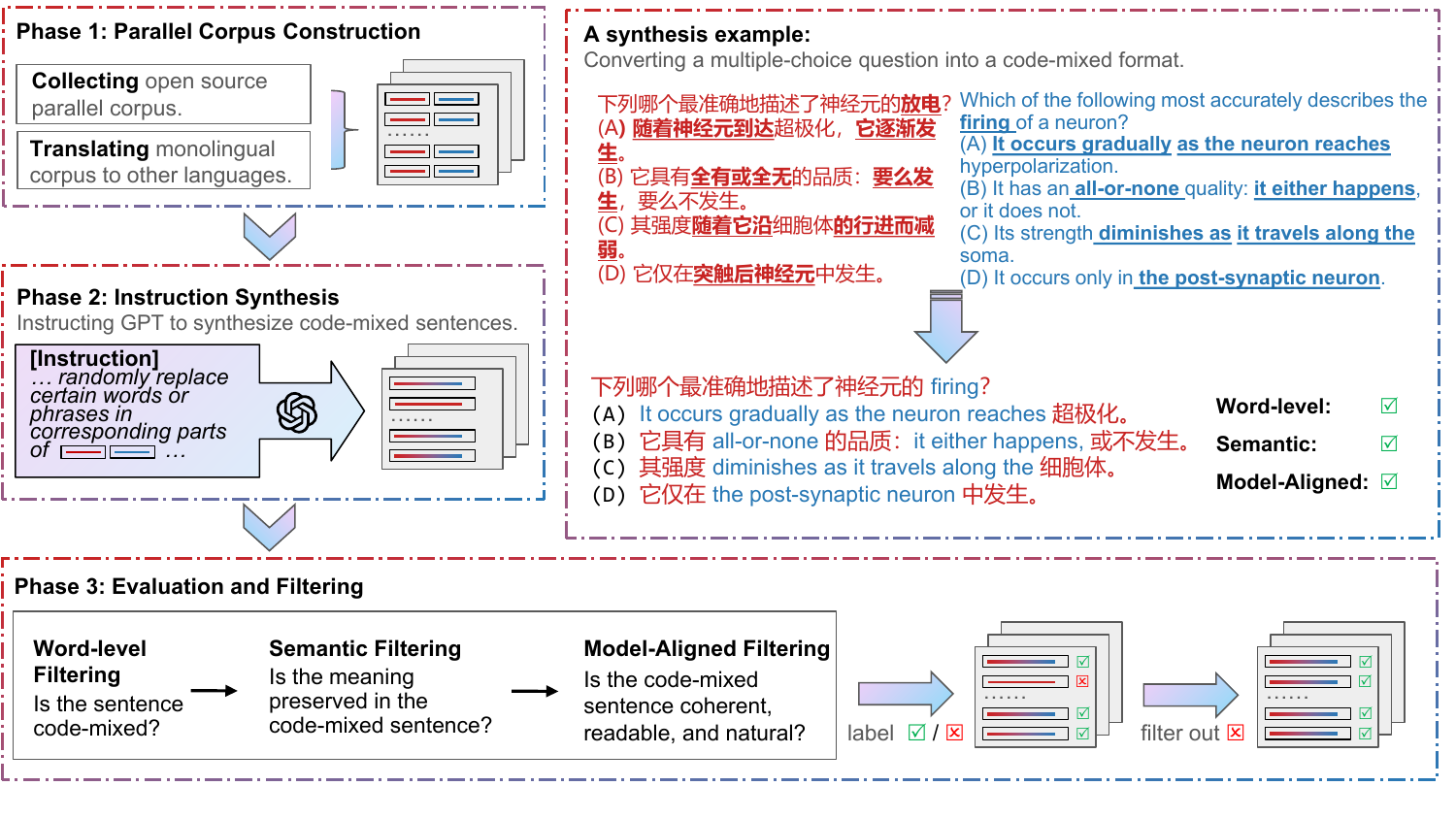}
 \vspace{-1em}
  \caption {\textbf{Illustration of the synthesis pipeline.}}
  \label{fig:pipeline}
  \vspace{-1em}
\end{figure*}

\subsection{Parallel Corpus Construction}
\label{sec:parallel corpus}
In first phase, we construct four parallel corpora for synthesizing code-mixed datasets. Using the multilingual MMLU test set from Opaki \citep{laiOkapiInstructiontunedLarge2023a}, we develop a parallel corpus of 4,018 multiple-choice questions, each available in 12 languages (en, zh, es, fr, ar, de, nl, hi, bn, mr, ne, ta). Additionally, we utilized GPT-4 Turbo to translate the English-only GSM8K and TruthfulQA datasets into four languages (zh, es, hi, ar), resulting in two parallel corpora with 1319 and 817 samples, respectively. To enhance linguistic diversity in machine translation tasks, we extracted a 4,344-sample parallel corpus (en, zh, es, ar) from the TED2013 dataset in OPUS \citep{tiedemannParallelDataTools2012}.

\subsection{Instruction Synthesis}
\label{sec:synthesis}
In second stage, we instruct GPT-4 Turbo to synthesize code-mixed sentences based on the parallel corpora. 
\revisedText{Code-mixing appears as a random alternation between and within sentences, but it is actually constrained by linguistic factors. \citet{POPLACK+1980+581+618} states code-mixing happens where the grammatical structures of both languages align. By ensuring that each language fragment is syntactically correct according to its own rules and that switches occur at structurally compatible points, word substitution between parallel corpus helps create coherent mixed-language sentences. Based on this idea, we devise a prompt (shown in Appendix \ref{app:synthesis}) for GPT-4 Turbo to randomly select and replace words or phrases in equivalent places where the surface structures of two sentences align. }

This method effectively embeds one language into another, implementing intra-sentential and inter-sentential code-mixing. Furthermore, we prompt GPT-4 Turbo to respond with the chosen words and their corresponding parts in another language. 

\subsection{Evaluation and Filtering} 
\label{sec:evaluation}
We implement a series of evaluation and filtering processes for the generated data.

\paragraph{Word-Level Filtering} We use the Multilingual Index  (M-index) \citep{doi:10.1177/13670069000040020101} and the Probability of Switching (I-index) \citep{Guzmn2017MetricsFM} as word-level evaluation metrics. Based on word-level language tagging (annotation strategy details in Appendix~\ref{app:annotation}), we calculated the M-index and I-index for each code-mixed text. Two code-mixing metrics are defined as:
    
    \begin{equation}
    M\text{-index} = \frac{1 - \sum p_j^2}{(k - 1) \sum p_j^2} \nonumber
    \end{equation}
    where \( p_j \) represents the proportion of the \( j \)-th category. \( k \) is the total number of language categories;
    \begin{equation}
    I\text{-index} = \frac{\sum_{1 \le i \le n-1} S(i, i+1)}{n-1} \nonumber
    \end{equation}
    where \( S(i, i+1) = 1 \) if the \( i \)-th and \( i+1 \)-th tokens of a sentence belongs to different languages; otherwise, \( S(i, i+1) = 0 \). \( n \) represents the total number of tokens in a sentence.
    
    The M-index ranges from 0 (monolingual text) to 1 (perfectly balanced code-mixed text with equal contributions from each language). Similarly, the I-index ranges from 0 (monolingual text) to 1 (optimal code-mixed text with alternating tokens from different languages). 
    \revisedText{To ensure dataset quality, we set the thresholds for the M-index and I-index to 0.1 to filter out monolingual sentences and those with low mixing or switching frequencies.}
    
\paragraph{Semantic Filtering}
To ensure the semantic consistency between the generated text and the original text, we computed the sentence similarity metrics for both texts. Additionally, We evaluated sentence similarity across two parallel corpora to assess the quality of the original parallel texts. First, we used LaBSE \citep{fengLanguageagnosticBERTSentence2022} (Appendix \ref{app:labse}) to project the original two monolingual texts (\textit{L1}, \textit{L2}) from the parallel corpus and the synthesized code-mixed text (\textit{CM}) into a common vector space. Subsequently, we calculate the pairwise cosine similarities among these three texts (\textit{CM}, \textit{L1}, \textit{L2}), resulting in three similarity scores ranging from 0 to 1. The score between \textit{CM} and \textit{L1/L2} partially reflects the synthesis quality of our method, while the score between \textit{L1} and \textit{L2} indicates the translation quality of the original parallel corpus. 
We determine that a similarity score below 0.8 suggests potential issues in the synthesis result or parallel corpus, necessitating the exclusion of such samples.

\paragraph{Model-Aligned Filtering}
To ensure the naturalness, coherence, and readability of synthesized sentences, we employ a highly human-aligned GPT-4 Turbo model (Appendix \ref{app:gpt4valid}) for automated evaluation. We prompt the model to assess synthetic results on naturalness, coherence, and readability, assigning scores to each criterion. Each criterion is rated on a scale from 1 to 3 (poor, fair, good), with detailed definitions provided for each level, shown in Appendix~\ref{app:humanlevel}. We filter out synthesized sentences if any score equals 1, indicating deficiencies in naturalness, coherence, or readability. 

%% file: tables/languages.tex

\begin{table}[t]
\centering
\resizebox{0.75\linewidth}{!}{%
\begin{tabular}{p{1.8cm}p{1.4cm}>{\centering\arraybackslash}p{0.7cm}>{\centering\arraybackslash}p{0.7cm}>{\centering\arraybackslash}p{0.7cm}}
\toprule
\multirow{2}{*}{\textbf{Family}} & \multirow{2}{*}{\textbf{Language}} & \textbf{ISO code} & \textbf{Pop.} (M) & \textbf{CC} (\%) \\ \midrule
\multirow{4}{*}{Germanic} & English & es & 1456 & 45.51 \\
 & German & de & 133 & 5.263 \\
 & Dutch & nl & 30 & 1.910 \\
 & Frisian & fy & 0.6 & \textbackslash{} \\ \midrule
\multirow{2}{*}{Sino-Tibetan} & Chinese & zh & 1138 & 4.423 \\ 
 & Hokkien & hok & 50 & \textbackslash{} \\ \midrule
\multirow{2}{*}{Romance} & Spanish & es & 559 & 4.594 \\
 & French & fr & 310 & 4.307 \\ \midrule
\multirow{3}{*}{Afro-Asiatic} & Arabic & ar & 380 & 0.617 \\
 & MSA & msa & 330 & \textbackslash{} \\ 
 & EA & ea & 103 & \textbackslash{} \\ \midrule
\multirow{4}{*}{Indo-Aryan} & Hindi & hi & 610 & 0.185 \\
 & Bengali & bn & 273 & 0.106 \\
 & Marathi & mr & 99 & 0.024 \\
 & Nepali & ne & 32 & 0.044 \\ \midrule
\multirow{2}{*}{Dravidian} & Tamil & ta & 87 & 0.042 \\ 
 & Malayalam & ml & 37 & 0.022 \\ \midrule
\multirow{1}{*}{Tupian} & Guarani & gn & 6.5 & \textbackslash{} \\ \bottomrule
\end{tabular}%
}
\caption{
    \textbf{Statistics of 18 languages from 7 families}. Each language is assigned a unique code in this paper based on the ISO 639. The \textit{Pop.} indicates the population in millions of speakers. The \textit{CC} indicates ratios of languages in the CommomCrawl. The \textit{MSA} and \textit{EA} stand for Modern Standard Arabic and Egyptian Arabic. 
  }
  \vspace{-1em}
\label{tab:languages}
\end{table}

%% file: section/experiment.tex
\subsection{Experiment Setup} 
\paragraph{Evaluation Settings}
For \textbf{CM-MMLU} and \textbf{CM-TruthfulQA}, we prompt models to select the correct option for multiple-choice questions. We use chain-of-thought (CoT) evaluation for \textbf{CM-GSM8K} task and parsed the model's response using regular regex to obtain the final solution. We report accuracy as the evaluation metric. For above three tasks, we also provide the model performance of English-only evaluation (\textit{en only}) for reference. 
For \textbf{LID, POS, NER, and SA} tasks, we prompt the models to generate the answers. Specifically, we provide the LLMs with all possible tags in the prompt and instruct models to generate in JSON format. In the \textbf{MT} task, we instructed models to translate code-mixed sentences. We use accuracy for LID, POS, NER, and SA tasks, and the BLEU score for MT assessment. All evaluations are under one-shot settings. We present the prompts for all 8 tasks in the Appendix \ref{app:experimentPrompt}. 

\paragraph{Models}
We selected LLMs from three different families for the comparison evaluation. For the GPT family, we evaluated GPT-3.5 Turbo-instruct, GPT-3.5 Turbo,  GPT-4 Turbo \cite{openaiGPT4TechnicalReport2024} and GPT-4o.
For the LLaMA family, we evaluated LLaMA2-Chat (7B, 13B, 70B) \cite{touvronLlamaOpenFoundation2023}, LLaMA3-Base (8B), and LLaMA3-Instruct (8B, 70B).
For the Mistral family, we evaluated Mistral 7B \cite{jiangMistral7B2023}, Mixtral 8x7B \cite{jiangMixtralExperts2024b}, and Mixtral 8x22B. 
We set the top-p to 0.95 and temperature to 0.8 for GPT, and used greedy decoding for LLaMA and Mistral models.

\input{tables/one-shot}

\begin{figure*}[h]
  \vspace{-2em}
  \includegraphics[width=\textwidth]{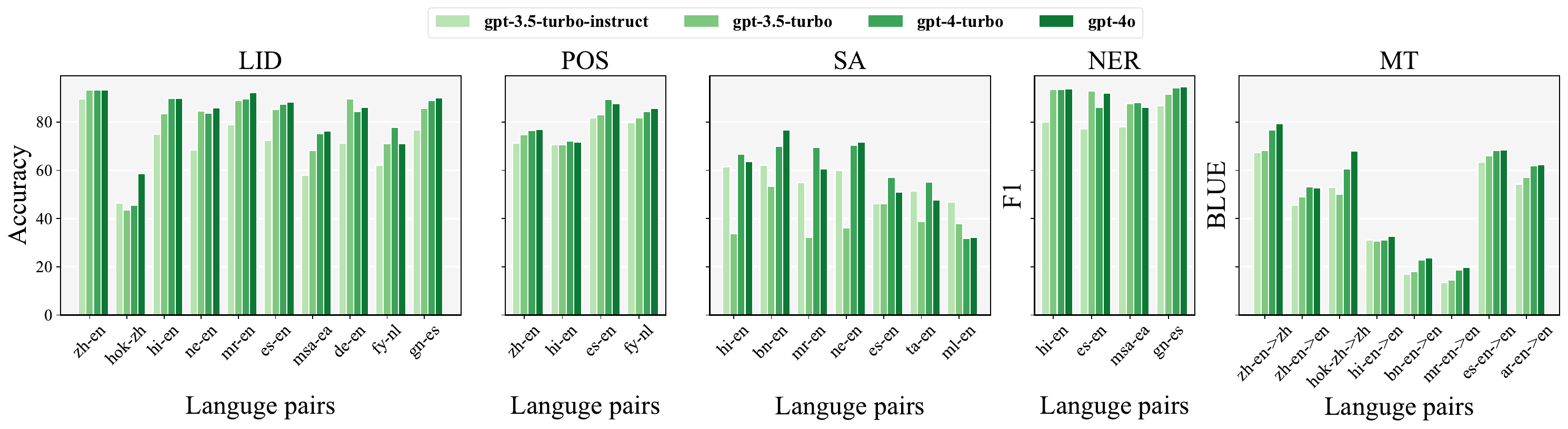}
  \vspace{-2em}
  \caption {\textbf{One-shot accuracy versus language pairs for GPT models on LID, POS, SA, NER and MT.}}
  \label{fig:tasks_2}
\end{figure*}

\subsection{Main Results}
Table~\ref{tab:one-shot_1} presents the experimental results of the selected models across the CM-MMLU, CM-GSM8K, and CM-TruthfulQA. Due to space constraints, the performance of the GPT family on LID, POS, NER, SA, and MT tasks is detailed in Appendix \ref{app:one-shot}, with visualizations provided in Figure~\ref{fig:tasks_2}.

\paragraph{Larger models excel on {\Name}} In Table~\ref{tab:one-shot_1}, GPT-4o achieves the highest scores across all language pairs in the CM-MMLU task, while  GPT-4 Turbo attains the highest scores for each language pair in the CM-GSM8K and CM-TruthfulQA tasks. This suggests that GPT-4o excels in comprehensive knowledge reasoning, whereas  GPT-4 Turbo is superior in mathematical reasoning and truthfulness. Additionally, within the LLaMA2, LLaMA3, and Mistral model families, the highest scores across all datasets consistently come from the largest models. Therefore, increasing model size enhances performance on multilingual code-mixed datasets.

\paragraph{GPT-3.5-Turbo-Instruct vs. GPT-3.5 Turbo} In Table~\ref{tab:one-shot_1}, GPT-3.5 Turbo outperforms GPT-3.5-Turbo-Instruct by an average of 2.07 points in CM-MMLU, 14.54 points in CM-GSM8K, and 7.23 points in CM-TruthfulQA. Table \ref{tab:one-shot_2} in the Appendix \ref{app:one-shot} shows GPT-3.5 Turbo scored higher on LID (+9.51\%), POS (+1.68\%), NER (+10.99\%), and MT (+1.07\%), but was 14.98 points lower on SA. This may be due to differing focuses during instruction tuning, with  GPT-3.5 Turbo emphasizing conversational completion and GPT-3.5-Turbo-Instruct focusing on instruction completion, leading to different training corpora. Thus, GPT-3.5 Turbo excelled over GPT-3.5-Turbo-Instruct in all {\Name} tasks except SA.

\paragraph{LLaMA3-8B-Base vs. LLaMA3-8B-Instruct} In Table~\ref{tab:one-shot_1}, LLaMA3-8B-Instruct performs comparably to LLaMA3-8B-Base on CM-MMLU and CM-TruthfulQA but outperforms it by 7.73 points on CM-GSM8K. This is likely due to the increased complexity of the mathematical reasoning required by CM-GSM8K. The improved performance on CM-GSM8K can be attributed to high-quality prompts during continued post-training stages, including supervised fine-tuning and alignment tuning, followed by the pre-training of LLaMA3.

\paragraph{LLaMA2 vs. LLaMA3} Table~\ref{tab:one-shot_1} shows that LLaMA3-8B outperforms LLaMA2-7B-Chat with average gains of 25.29, 57.29, and 23.43 points on CM-MMLU, CM-GSM8K, and CM-TruthfulQA, respectively. Additionally, LLaMA3-70B surpasses LLaMA2-70B with improvements of 25.75, 45.80, and 35.24 points on the same benchmarks. These enhancements may be due to the training dataset for LLaMA3 containing over 15T tokens, a size seven times larger than that used for LLaMA2.

\paragraph{Mistral 7B vs. Mixtral 8x7B} We also observe from Table~\ref{tab:one-shot_1} that  Mixtral 8x7B outperforms Mistral 7B by 12.14, 19.73, and 8.98 points on CM-MMLU, CM-GSM8K, and CM-TruthfulQA, correspondingly. This improvement is likely due to the scaling of model parameters in Mixture of Experts (MoE) architecture and the substantial increase in multilingual training compute for Mixtral 8x7B.

\begin{figure*}[h]
  \vspace{-1em}
  \includegraphics[width=\textwidth]{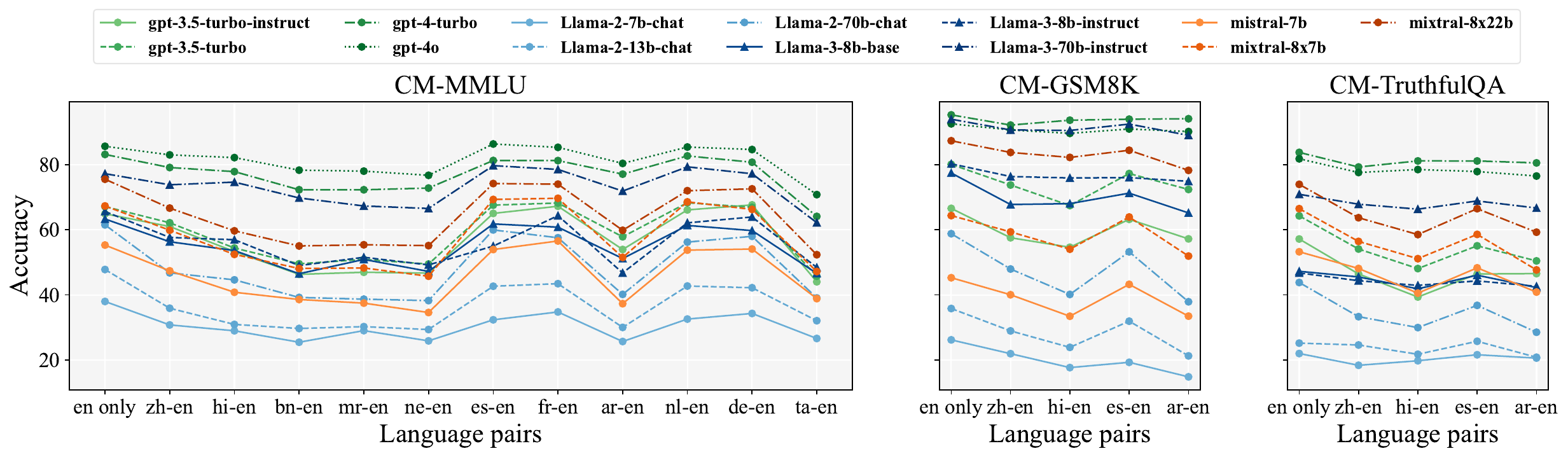}
  \vspace{-1em}
  \caption {\textbf{Accuracy versus language pairs for models on CM-MMLU, CM-GSM8K and CM-TruthfulQA.}}
  \vspace{-1em}
  \label{fig:one_shot}
\end{figure*}

\subsection{Analysis across Languages}
Figure~\ref{fig:one_shot} illustrates the accuracy variations of LLMs from three families on the CM-MMLU, CM-GSM8K, and CM-TruthfulQA tasks across different language pairs.

\vspace{0.5mm} \noindent \textbf{Cross-family code-mixing can impair the performance of LLMs.} Figure~\ref{fig:one_shot} shows significant fluctuations in \textit{zh-en}, \textit{hi-en}, \textit{bn-en}, \textit{mr-en}, \textit{ne-en}, \textit{ar-en}, and \textit{ta-en} language pairs, while \textit{es-en}, \textit{fr-en}, \textit{de-en}, and \textit{nl-en} pairs perform similarly to English-only scenario. 
\revisedText{This similarity may be attributed to English, German, Dutch, Spanish, and French having similar word order features according to WALS \citep{wals}, along with their common Indo-European family and geographic proximity.}
Therefore, code-mixing between languages with substantial linguistic differences can significantly hinder the performance of LLMs.

\vspace{0.5mm} \noindent \textbf{Models exhibit consistent fluctuation patterns across different code-mixed language pairs.} Figure~\ref{fig:one_shot} reveals a notable trend: despite originating from three distinct institutions, the models display parallel accuracy fluctuations across different language pairs for the three tasks. For CM-MMLU, most models show a decline in accuracy from \textit{en only} to \textit{ne-en}, followed by a rebound for \textit{es-en} and \textit{fr-en}. This uniform impact on performance likely results from overlapping training data sourced from the internet, commonly used by three organizations during model training.

\vspace{0.5mm} \noindent \textbf{More low-resource data improves code-mixing comprehension.} Analyzing Figure~\ref{fig:one_shot} and Table~\ref{tab:one-shot_1}, the decrease for high-resource language and English code-mixing (\textit{zh-en}) was 2.63 points compared to English-only datasets. Medium-resource language code-mixing (\textit{hi-en}, \textit{ar-en}, \textit{bn-en}) showed declines of 3.47, 5.25, and 7.32 points, respectively, while low-resource language mixtures (\textit{mr-en}, \textit{ne-en}, \textit{ta-en}) experienced more substantial drops of 7.62, 8.9, and 14.83 points. This indicates the model has a better understanding of code-mixed data involving high-resource languages and English. Consequently, increasing training on low-resource language corpora could improve the model's comprehension of code-mixed data involving these languages.

\subsection{$K$-shot Analysis}

\begin{figure*}[t]
\vspace{-3em}
\centering
\subfigure[CM-MMLU]{
  \includegraphics[width=\textwidth]{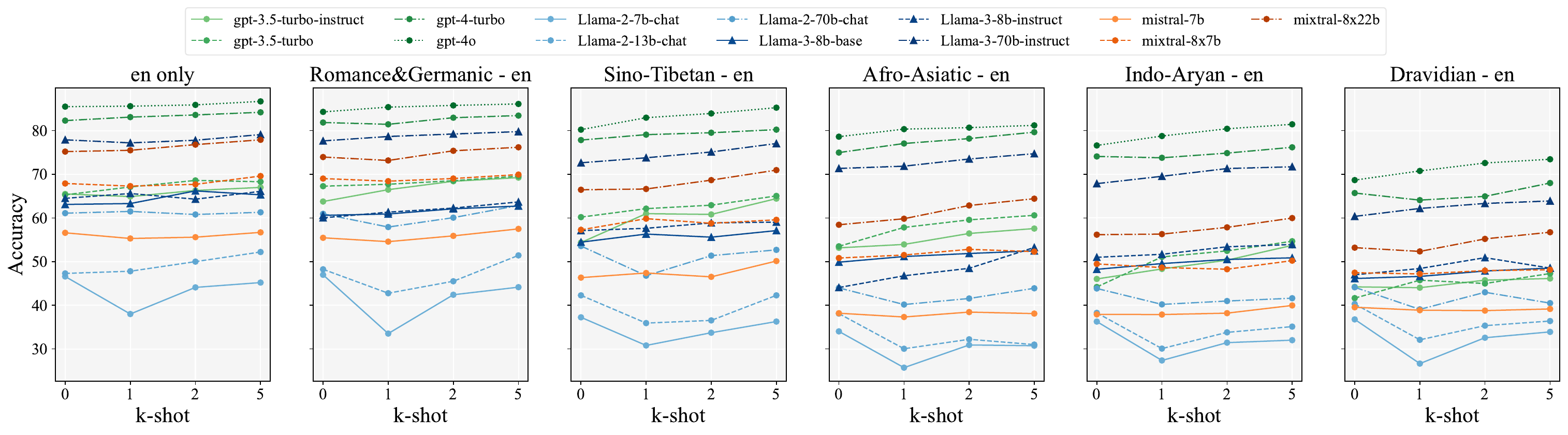}
  \label{fig:mmlu_kshot}
}
\subfigure[CM-GSM8K]{
  \includegraphics[width=0.9\textwidth]{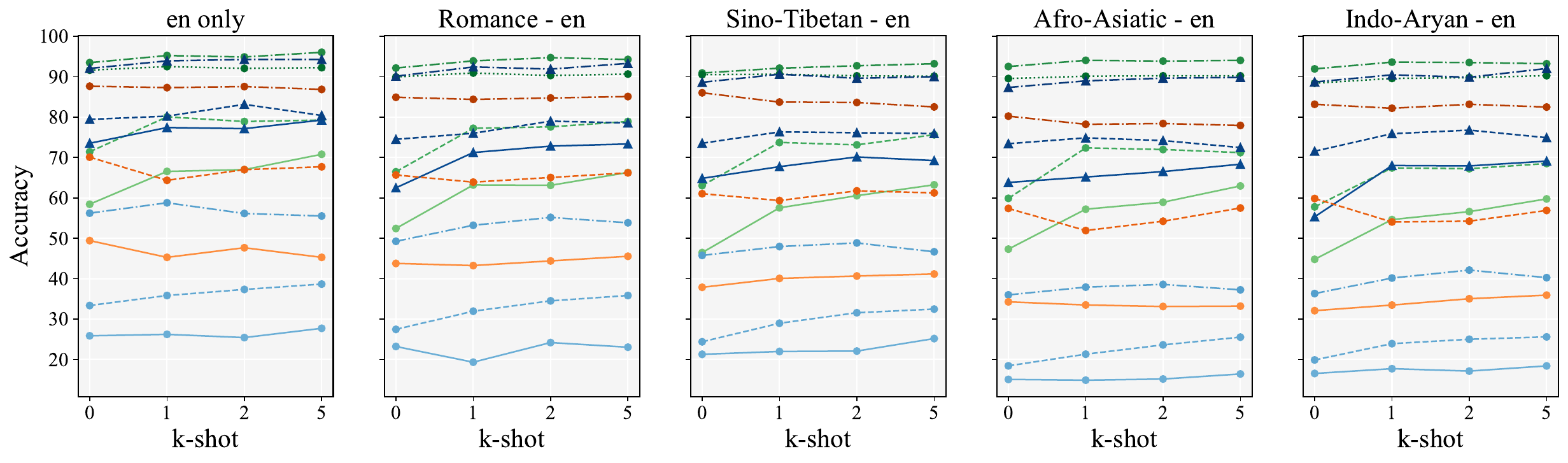}
  \label{fig:gsm8k_kshot}

}
\subfigure[CM-TruthfulQA]{
  \includegraphics[width=0.9\textwidth]{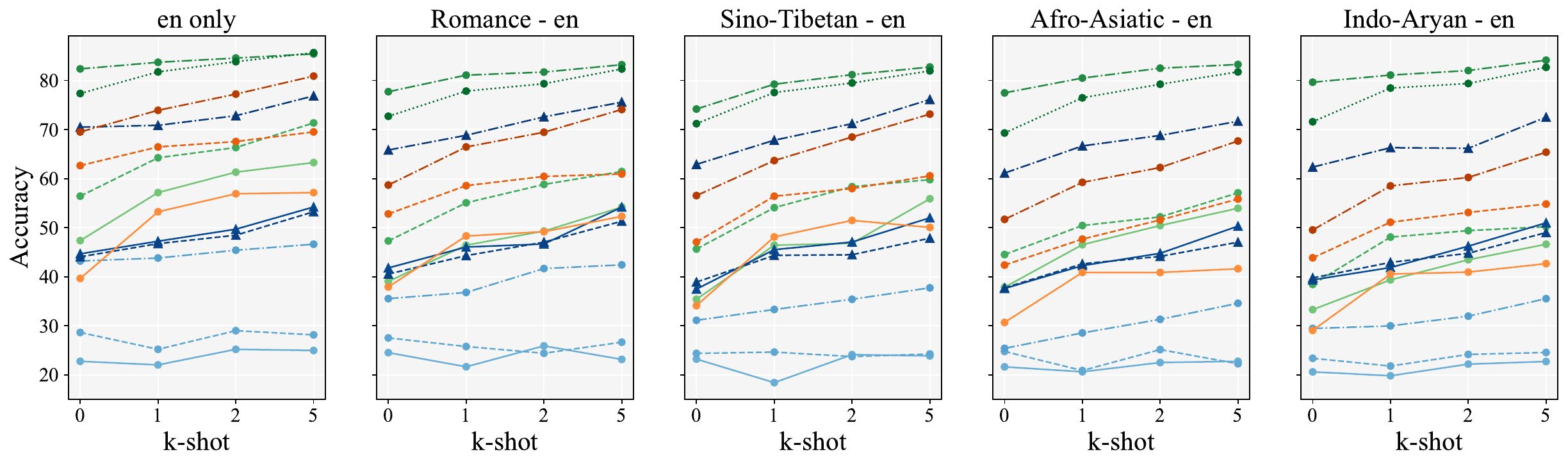}
  \label{fig:truthfulqa_kshot}

}
\vspace{-1em}
\caption {\textbf{Accuracy of $k$-shot evaluation for three model families on three tasks.}}
\label{fig:kshot}
\vspace{-1em}
\end{figure*}

To further investigate the impact of varying quantities of code-mixed examples on model performance, we conducted $k$-shot evaluations ($k \in \{ 0, 1, 2, 5\}$) on the CM-MMLU, CM-GSM8K, and CM-TruthfulQA datasets. English-only (\textit{en only}) served as a control group, allowing us to compare performance trends between the \textit{en only} and various code-mixed scenarios across different language families. Results were averaged by language family and visualized in Figure~\ref{fig:kshot}, with full results in Appendix \ref{app:k-shot}. Figure~\ref{fig:kshot} shows that models like GPT-4 Turbo, GPT-4o, and LLaMA3-70B-Instruct, which have higher average accuracy scores, maintain more stable $k$-shot accuracy trends as $k$ increases. This indicates their robust multilingual and few-shot learning abilities. In contrast, other models often experience sudden drops in accuracy for certain language pairs as $k$ increases.

\vspace{0.5mm} \noindent \textbf{Advanced models excel at few-shot learning on knowledge and truthfulness reasoning.} Figure~\ref{fig:mmlu_kshot} indicates that the accuracy of GPT models, LLaMA3, and Mistral generally increases with higher $k$ on the CM-MMLU, whereas LLaMA2 models show a significant drop at one-shot before recovering. LLaMA2-13B-Chat and LLaMA2-70B-Chat demonstrate a positive correlation between accuracy and $k$ values in \textit{en only} datasets, indicating their few-shot learning capabilities. In contrast, for code-mixed datasets, one-shot and two-shot accuracies are lower than zero-shot, with even five-shot performance lagging behind zero-shot for \textit{Sino-Tibetan - en}, \textit{Afro-Asiatic - en}, \textit{Indo-Aryan - en}, and \textit{Dravidian - en}. This suggests code-mixing hinders the one-shot and two-shot learning capabilities of these models, though performance can gradually recover at five-shot. Also in Figure~\ref{fig:truthfulqa_kshot}, except for LLaMA2-7B-Chat and LLaMA2-13B-Chat, all models' accuracy scores increase with $k$ on CM-TruthfulQA. In summary, few-shot learning is effective for all selected models except LLaMA2 in knowledge and truthfulness reasoning.

\vspace{0.5mm} \noindent \textbf{Few-shot learning minimally enhances mathematical reasoning.} In Figure~\ref{fig:gsm8k_kshot}, the model’s $k$-shot accuracy on the CM-GSM8K task shows less variability compared to the other tasks, which we attribute to the higher complexity of CM-GSM8K relative to CM-MMLU and CM-TruthfulQA, posing greater challenges to a model's few-shot learning. However, GPT-3.5 Turbo, GPT-3.5-Turbo-Instruct, and LLaMA3-8B-Base exhibit significant accuracy improvements from zero-shot to one-shot. This is because these models initially fail to follow the required response format in zero-shot prompts, causing incorrect answers, while one-shot improves these models' output format adherence. Besides, Mixtral-7x22b demonstrates a decrease in accuracy as $k$ increases across all datasets, indicating its inadequate few-shot learning capability on CM-GSM8K. Overall, in the CM-GSM8K task, few-shot learning provides limited enhancement in mathematical reasoning for the GPT and LLaMA family and may negatively affect Mistral models.

%% file: tables/one-shot.tex
\begin{table*}[t]
\centering
\vspace{-2em}
\resizebox{0.95\textwidth}{!}{%
\begin{tabular}{@{}lccccccccccccc@{}}\toprule
\multicolumn{1}{c}{} & \multicolumn{1}{c|}{\begin{tabular}[c]{@{}c@{}}\textbf{GPT}\\ \textbf{-Instruct}\end{tabular}} & \multicolumn{3}{c|}{\textbf{GPT}} & \multicolumn{3}{c|}{\begin{tabular}[c]{@{}c@{}}\textbf{LLaMA2}\\ \textbf{-Chat}\end{tabular}} & \multicolumn{1}{c}{\begin{tabular}[c]{@{}c@{}}\textbf{LLaMA3}\\ \textbf{-Base}\end{tabular}} & \multicolumn{2}{c|}{\begin{tabular}[c]{@{}c@{}}\textbf{LLaMA3}\\ \textbf{-Instruct}\end{tabular}}  & \multicolumn{3}{c}{\textbf{Mistral \& Mixtral}} \\ \cmidrule(l){2-14}
 & \multicolumn{1}{c|}{3.5-T} & 3.5-T & 4-T & \multicolumn{1}{c|}{4o} & 7b & 13b & \multicolumn{1}{c|}{70b} & \multicolumn{1}{c}{8b} & 8b & \multicolumn{1}{c|}{70b} & \multicolumn{1}{c}{7b} & 8x7b & \multicolumn{1}{c}{8x22b} \\ \midrule
\multicolumn{14}{c}{\textit{CM-MMLU}} \\ 
en only & 64.90 & 66.30 & 83.10 & \textbf{85.60} & 38.00 & 47.80 & \textbf{61.50} & 63.30 & 65.60 & \textbf{77.20} & 55.3 & 67.30 & \textbf{75.50} \\
zh-en & 60.99 & 60.81 & 79.08 & \textbf{82.97} & 30.80 & 35.92 & \textbf{46.78} & 56.31 & 57.63 & \textbf{73.79} & 47.40 & 59.84 & \textbf{66.64} \\
hi-en & 53.32 & 55.37 & 77.83 & \textbf{82.13} & 29.00 & 30.96 & \textbf{44.63} & 53.61  & 56.93 & \textbf{74.61} & 40.82 & 52.44 & \textbf{59.67} \\
bn-en & 46.32 & 47.49 & 72.26 & \textbf{78.28} & 25.49 & 29.71 & \textbf{39.23} & 46.50  & 49.01 & \textbf{69.75} & 38.60 & 48.11 & \textbf{55.03} \\
mr-en & 46.95 & 49.67 & 72.26 & \textbf{77.98} & 29.05 & 30.27 & \textbf{38.71} & 50.89 & 51.55 & \textbf{67.29} & 37.49 & 48.27 & \textbf{55.39} \\
ne-en & 46.70 & 48.78 & 72.78 & \textbf{76.70} & 25.91 & 29.39 & \textbf{38.26} & 47.22 & 49.22 & \textbf{66.52} & 34.61 & 45.74 & \textbf{55.13} \\
es-en & 65.01 & 69.20 & 81.24 & \textbf{86.30} & 32.37 & 42.67 & \textbf{59.95} & 61.78  & 54.98 & \textbf{79.67} & 53.93 & 69.28 & \textbf{74.17} \\
fr-en & 67.21 & 68.83 & 81.21 & \textbf{85.28} & 34.78 & 43.45 & \textbf{57.54} & 60.79  & 64.32 & \textbf{78.50} & 56.55 & 69.65 & \textbf{73.98} \\
ar-en & 53.94 & 56.45 & 77.06 & \textbf{80.35} & 25.71 & 30.04 & \textbf{40.17} & 51.17  & 46.76 & \textbf{71.86} & 37.32 & 51.52 & \textbf{59.83} \\
ta-en & 44.03 & 45.75 & 64.09 & \textbf{70.77} & 26.65 & 32.09 & \textbf{39.06} & 46.61  & 48.42 & \textbf{62.18} & 38.87 & 47.18 & \textbf{52.34} \\
nl-en & 66.08 & 67.14 & 82.64 & \textbf{85.37} & 32.60 & 42.73 & \textbf{56.21} & 61.32  & 62.11 & \textbf{79.30} & 53.74 & 68.55 & \textbf{71.98} \\
de-en & 67.63 & 68.46 & 80.71 & \textbf{84.60} & 34.32 & 42.21 & \textbf{57.98} & 59.74   & 63.91 & \textbf{77.18} & 54.08 & 66.23 & \textbf{72.54} \\ 
Average & 56.20 & 58.27 & 76.47 & \textbf{80.97} & 29.70 & 35.40 & \textbf{47.14} & 54.18 & 54.99 & \textbf{72.79} & 44.85 & 56.98 & \textbf{63.34} \\ \midrule
\multicolumn{14}{c}{\textit{CM-GSM8K}} \\
en only & 66.55  & 80.05 & \textbf{95.23} & 92.50 & 26.21 & 35.83 & \textbf{58.78} & 77.41  & 80.23 & \textbf{93.91} & 45.28 & 64.34 & \textbf{87.29} \\
zh-en & 57.54  & 73.73 & \textbf{92.11} & 90.61 & 21.98 & 28.97 & \textbf{47.95} & 67.73  & 76.32 & \textbf{90.61} & 40.06 & 59.34 & \textbf{83.72} \\
hi-en & 54.63  & 67.42 & \textbf{93.60} & 89.57 & 17.72 & 23.92 & \textbf{40.16} & 68.01 & 75.89 & \textbf{90.45} & 33.46 & 54.04 & \textbf{82.19} \\
es-en & 63.20  & 77.23 & \textbf{93.91} & 90.91 & 19.33 & 31.95 & \textbf{53.22} & 71.23  & 75.99 & \textbf{92.41} & 43.25 & 63.90 & \textbf{84.38} \\
ar-en & 57.20  & 72.36 & \textbf{94.05} & 90.12 & 14.88 & 21.31 & \textbf{37.91} & 65.16  & 74.86 & \textbf{88.96} & 33.49 & 51.92 & \textbf{78.21} \\ 
Average & 58.14 & 72.68 & \textbf{93.42} & 90.30 & 18.47 & 26.54 & \textbf{44.81} & 68.03  & 75.76 & \textbf{90.61} & 37.57 & 57.30 & \textbf{82.12} \\ \midrule
\multicolumn{14}{c}{\textit{CM-TruthfulQA}} \\
en only & 57.16  & 64.26 & \textbf{83.72} & 81.76 & 22.03 & 25.21 & \textbf{43.82} & 47.25  & 46.76 & \textbf{70.87} & 53.24 & 66.46 & \textbf{73.93} \\
zh-en & 46.43  & 54.09 & \textbf{79.25} & 77.56 & 18.42 & 24.64 & \textbf{33.33} & 45.53  & 44.36 & \textbf{67.83} & 48.12 & 56.42 & \textbf{63.68 }\\
hi-en & 39.37  & 48.08 & \textbf{81.11} & 78.47 & 19.82 & 21.80 & \textbf{29.99} & 41.88  & 42.93 & \textbf{66.31} & 40.55 & 51.12 & \textbf{58.52} \\
es-en & 46.43  & 55.07 & \textbf{81.10} & 77.85 & 21.65 & 25.78 & \textbf{36.80} & 46.06  & 44.31 & \textbf{68.84} & 48.31 & 58.57 & \textbf{66.46} \\
ar-en & 46.54  & 50.44 & \textbf{80.50} & 76.48 & 20.63 & 20.88 & \textbf{28.55} & 42.26  & 42.64 & \textbf{66.67 }& 40.88 & 47.67 & \textbf{59.25} \\ 
Average & 44.69 & 51.92 & \textbf{80.49} & 77.59 & 20.13 & 23.28 & \textbf{32.17} & 43.93  & 43.56 & \textbf{67.41} & 44.47 & 53.45 & \textbf{61.98} \\
\bottomrule
\end{tabular}%
}
\caption{
\textbf{One-shot accuracy of selected models on CM-MMLU, CM-GSM8K and CM-TruthfulQA.} Where 3.5-T indicates GPT-3.5 Turbo, and 4-T indicates GPT-4 Turbo. The \textit{en only} stands for a dataset we randomly sample from the test set of the original dataset in English. To be compared with other code-mixed datasets, the \textit{en only} datasets for CM-MMLU, CM-GSM8K, and CM-TuthfulQA contain 1000, 1133, and 817 English instances each. The \textit{Average} represents the mean score of each model across various datasets (excluding \textit{en only} dataset) from a given task. For each model family, the scores of the top-performing models are highlighted in bold.
}
\label{tab:one-shot_1}
\vspace{-1em}
\end{table*}

%% file: section/conclusion.tex
This study introduces {\Name}, a comprehensive benchmark for evaluating code-mixing performance in LLMs, spanning eight tasks and 18 languages. We also adopt GPT-4 Turbo for constructing synthetic code-mixed data to address data scarcity issues.
Our findings show that while code-mixing challenges LLMs performance, improvements can be achieved through larger pre-training datasets, increased model scales, and few-shot learning. In the future, {\Name} holds great promise for evaluating the code-mixing capabilities of LLMs and inspiring further research in this area.


%% file: section/appendix.tex
\section{CodeMixBench vs. other benchmarks}
\label{app:versus}
\input{tables/versus}

\revisedText{As shown in Table \ref{tab:versus}, LinCE includes four language pairs and five NLP tasks: Language Identification (LID), Part of Speech (POS), Named Entity Recognition (NER), Sentiment Analysis (SA), and Machine Translation (MT). In contrast, GLUECoS covers two language pairs, lacks the MT task, but adds Question Answering (QA) and Natural Language Inference (NLI). Our review of recent code-mixing studies indicates that research extends beyond the language pairs used in LinCE and GLUECoS. Therefore, we expanded to 16 language pairs and introduced tasks better suited for evaluating LLMs, such as Multi-Choice, Math, and Truthfulness, resulting in a total of eight tasks.}

\section{Collected Datasets}
\label{app:collected}
In Table \ref{tab:collected}, we selected and reconstructed 30 datasets from existing open-source projects. To comprehensively evaluate the performance of large models on code-mixing, we aimed to encompass a diverse range of language families and tasks, prioritizing manually annotated datasets. Ultimately, we cover traditional NLP tasks such as Language Identification (LID), Named Entity Recognition (NER), Part-of-Speech tagging (POS), Sentiment Analysis (SA), and Machine Translation (MT), and cover 16 languages from seven language families: Germanic (en, de, nl, fy), Sino-Tibetan (zh, hok), Romance (es), Afro-Asiatic (msa, ea), Indo-Aryan (hi, bn, ne, mr), Dravidian (ta, ml), and Tupian (gn).

\subsection{Datasets of LID Task}
\input{tables/collected_datasets}

\paragraph{zh-en} \citet{calvilloSurprisalPredictsCodeSwitching2020b} collected data from the Chinese Students and Scholars Association Bulletin Board Systems (CSSA BBS) of Pennsylvania State University, Carnegie Mellon University, and the University of Pittsburgh. The dataset consists of posts from bilingual Chinese-English speakers who have studied in the US for several years. The dataset includes 3,022 samples, totalling 37,064 tokens, with 25,092 Chinese tokens, 7,228 English tokens, and 4,744 punctuation tokens. 

\paragraph{hok-zh} \citet{luExploringMethodsBuilding2022a} utilized a rule-based approach to synthesize parallel corpora into a Hokkien-Mandarin code-mixed corpus, ensuring dataset quality through subsequent post-processing steps. The parallel corpora are derived from iCorpus and the Ministry of Education's Taiwanese Southern Min Dictionary (MoeDict). The test set comprises 3,800 code-mixed sentences and 44,022 tokens, with the distribution as follows: 30,941 Hokkien tokens, and 13,081 Mandarin tokens. 

\paragraph{hi-en} LinCE  \citep{aguilarLinCECentralizedBenchmark2020} constructed the Hindi-English dataset based on \citet{maveLanguageIdentificationAnalysis2018b} and ICON 2016 competition \citet{sequieraPOSTaggingHindiEnglish2015}. We utilized the development set, comprising 744 social media posts from Twitter and Facebook, with the following token distribution: Hindi (8,997), English (3,306), language-independent tokens (2,231), mixed (5), named entities (875), unknown (2), foreign words (29), ambiguous (1). 

\paragraph{ne-en} The Nepali-English corpus, originally introduced by 2014 CALCS (Computational Approaches to Linguistic Code-Switching) workshop \citep{solorioOverviewFirstShared2014}, has been restructured by LinCE. We use the development set, comprising 1,332 tweets, with the following token distributions: Nepali (5,649), English (8,417), named entities (514), mixed (17), and ambiguous (13).

\paragraph{mr-en} We selected the test set from the MeLID dataset developed by \citet{chavanMyBoliCodemixed2023}, which includes 1,340 Marathi-English code-switched tweets annotated by four native Marathi speakers. It contains 11,485 Marathi tokens, 2,925 English tokens, and 1,535 tokens from other categories, intended to facilitate research in language identification tasks.

\paragraph{es-en} The Spanish-English corpus was obtained from the 2016 CALCS workshop \citep{molinaOverviewSecondShared2016}. LinCE provided new splits for this corpus, and we employ the development set, which comprises 3,332 tweets and 40,391 tokens. The token distribution in the development set is as follows: English tokens (16,712), Spanish tokens (14,955), language-independent tokens (7,830), tokens mixed in English and Spanish (6), named entities (815), unknown (32), foreign words (2), and ambiguous (39). 

\paragraph{msa-ea} LinCE restructured the Modern Standard Arabic (MSA)-Egyptian Arabic (EA) corpus from the 2016 CALCS workshop \citep{molinaOverviewSecondShared2016}. We choose the development set, comprising 1,332 tweets, with the following token distributions: MSA (13,317), EA (4,100), language-independent tokens (1,707), named entities (2,688), mixed (2), and ambiguous (164).

\paragraph{de-en} The TONGUESWITCHER \citep{sternerTongueSwitcherFineGrainedIdentification2023} project offers a substantial corpus of 25.6 million German-English code-switched tweets, annotated using both rule-based and neural network methods. We use the test set of the dataset which contains 1,252 tweets and 37,511 tokens, We utilize the test set from the dataset, comprising 1,252 tweets and 37,511 tokens: 34,190 in German, 3,175 in English, and 146 in German-English code-switching.

\paragraph{fy-nl} This dataset originated from broadcasts by Omrop Fryslân (Frisian Broadcasting Company), comprising approximately 18.5 hours of spontaneous interviews. \citet{braggaarChallengesAnnotatingParsing2021} randomly selected and annotated 400 utterances with LID tags. Among these, 67.8\% of the words are in Frisian, 26.1\% are in Dutch, and the rest comprise a mix of Frisian-Dutch, hesitation markers (e.g., "eh"), or other languages. We use the test subset of the dataset, comprising 280 samples, with the following token counts: Dutch (378), Frisian (1,955), Frisian-Dutch (15), and other tokens (8).

\paragraph{gn-es} \citet{chiruzzoOverviewGUASPAIberLEF2023a} built this dataset from news articles and tweets. It consists of approximately 25,000 tokens and is annotated in two stages by six annotators proficient in Spanish and with some knowledge of Guarani. We utilized the test set comprising 180 sentences and 2,857 tokens, categorized as follows: 1,193 Guarani tokens, 815 Spanish tokens, 47 mixed-language tokens, 8 foreign words, 331 named entities, and 463 tokens classified as other.

\subsection{Datasets of POS Task}
\paragraph{zh-en} In the zh-en dataset in the LID task, we introduced the dataset built by \citet{calvilloSurprisalPredictsCodeSwitching2020b}, which is based on the CSSA BBS. They employed the Stanford Parser to obtain POS tags for code-mixed sentences. We selected a dataset consisting of 2,909 sentences and 35,600 tokens. The distribution of POS tags is as follows: NN (9,990), PU (5,880), VC (604), CD (1,691), M (1,207), JJ (710), P (736), MSP (41), VV (5,331), VA (924), VE (716), DEG (677), CC (461), AD (3,236), PN (788), DT (493), NT (273), LC (324), DEC (447), SP (250), OD (67), NR (359), ETC (60), CS  (87), AS (180), DER (11), SB (10), BA (16), URL (1), DEV (13), IJ (15), and LB (2).

\paragraph{hi-en} LinCE proposed standard splits for a dataset comprising 1,489 tweets (33,010 tokens) annotated with POS tags \citep{singhTwitterCorpusHindiEnglish2018}. We select a development set of 160 tweets, with the following token counts per POS category: 
\begin{itemize}
    \item X (790) for all other categories such as abbreviations or foreign words. 
    \item VERB (669) is used for verbs. 
    \item NOUN (516) is used for nouns. 
    \item ADP (346) is used for prepositions and postpositions. 
    \item PROPN (271) is used for proper nouns. 
    \item ADJ (170) is used for adjectives. 
    \item PRON (159) is used for pronouns. 
    \item PART (145) is used for particles. 
    \item DET (116) is used for determiners and articles. 
    \item ADV (100) is used for adverbs. 
    \item CONJ (77) is used for coordinating conjunctions. This is represented by ‘CCONJ’ in the universal POS tagset. 
    \item PART\_NEG (43) is used for indicating negation.
    \item PRON\_WH (39) is used for interrogative pronouns (like where, why, etc.). 
    \item NUM (35) is used for numerals. 
\end{itemize}

\paragraph{es-en} The Spanish-English dataset is derived from the Miami Bangor corpus \citep{sotoCrowdsourcingUniversalPartofSpeech2017a}. LinCE stratified the dataset into training (27,893 sentences), development (4,298 sentences), and testing (10,720 sentences) sets. From the development set, 1,000 samples (totalling 7,712 tokens) were randomly selected. The token counts per part-of-speech tag are as follows: VERB (1,262), PUNCT (1,234), PRON (1,189), NOUN (676), DET (552), ADV (498), ADP (472), INTJ (362), CONJ (278), ADJ (254), AUX (243), SCONJ (238), PART (165), PROPN (150), NUM (86), and UNK (53).

\paragraph{fy-nl} \citet{braggaarChallengesAnnotatingParsing2021} also annotated 400 broadcast utterances with POS tags. We utilize the test set, which includes 280 samples. The token counts for each POS tag in this subset are as follows: NOUN (310), ADP (288), PRON (285), ADV (284), VERB (263), DET (232), PROPN (154), ADJ (142), AUX (111), INTJ (105), CCONJ (101), SCONJ (41), and NUM (40). 

\subsection{Datasets of NER Task}
\paragraph{hi-en} \citet{singhLanguageIdentificationNamed2018a} developed a dataset of 2,079 tweets annotated by three linguistic experts, and subsequently splits by LinCE. From this dataset, we selected a development set of 314 tweets, comprising 5,364 tokens. The token distribution includes 4,789 O tokens, 61 B-ORGANISATION tokens, 19 I-ORGANISATION tokens, 254 B-PERSON tokens, 112 I-PERSON tokens, 105 B-PLACE tokens, and 24 I-PLACE tokens.

\paragraph{es-en} The Spanish-English corpus, introduced at the 2018 CALCS workshop \citep{aguilarNamedEntityRecognition2018a} for NER, was used fairly splited by LinCE. We randomly sample 1,000 instances from the development set, comprising a total of 12,139 tokens. The distribution of entity tokens is as follows: 11,834 O tokens, 82 B-PER tokens, 25 I-PER tokens, 21 B-PROD tokens, 3 I-PROD tokens, 47 B-LOC tokens, 18 I-LOC tokens, 7 B-TIME tokens, 4 I-TIME tokens, 12 B-ORG tokens, 13 I-ORG tokens, 5 B-EVENT tokens, 7 I-EVENT tokens, 14 B-TITLE tokens, 18 I-TITLE tokens, 14 B-GROUP tokens, 7 I-GROUP tokens, 7 B-OTHER tokens, and 1 I-OTHER token.

\paragraph{msa-ea} This MSA-EA corpus was also introduced at the 2018 CALCS workshop. We utilized the development set, comprising 1122 samples with a total of 22742 tokens. The token distribution is as follows: O tokens: 20,031, B-PER tokens: 698, I-PER tokens: 415, B-GROUP tokens: 191, I-GROUP tokens: 112, B-LOC tokens: 358, I-LOC tokens: 116, B-PROD tokens: 55, I-PROD tokens: 26, B-ORG tokens: 149, I-ORG tokens: 114, B-TITLE tokens: 115, I-TITLE tokens: 143, B-EVENT tokens: 69, I-EVENT tokens: 52, B-TIME tokens: 61, I-TIME tokens: 18, B-OTHER tokens: 17, and I-OTHER tokens: 2.

\paragraph{gn-es} In LID task, we introduce the Guarani-Spanish dataset constructed by \citet{chiruzzoOverviewGUASPAIberLEF2023a}. In addition to LID labels, this dataset contains manually annotated NER tags. We selected the test set, which comprises the following token counts: 2526 overall tokens, 81 B-PER tokens, 89 B-ORG tokens, 34 I-PER tokens, 33 B-LOC tokens, 21 I-LOC tokens, and 73 I-ORG tokens.

\subsection{Datasets of SA Task}
\paragraph{hi-en} \citet{patraSentimentAnalysisCodeMixed2018} built this Hindi-English dataset, derived from the social media platform Twitter, has been manually annotated for sentiment, encompassing positive, negative, and neutral labels. For our study, we utilized a test set comprising 1,261 samples, which includes 385 positive, 290 negative, and 586 neutral instances.

\paragraph{bn-en} The TB-OLID dataset \citep{raihanOffensiveLanguageIdentification2023c}, designed for offensive language detection in code-mixed texts, comprises 5,000 Facebook comments, with English constituting 38.42\% of the content. All comments are manually annotated. For our benchmark, we utilized the test subset of 1,000 instances, consisting of 573 non-offensive and 427 offensive comments.

\paragraph{mr-en} \citet{chavanMyBoliCodemixed2023} also provided a Marathi-English dataset with manually annotated sentiment labels. We selected the test set containing 1,250 instances, distributed as 417 positive, 417 negative, and 416 neutral samples.

\paragraph{ne-en} The dataset consists of code-switched Nepali-English comments from YouTube, intended for sentiment analysis with manually annotated labels \citep{pahariLanguagePreferenceExpression2023}. The test set we used includes 1,070 samples, distributed as follows: 346 Positive, 359 Negative, and 365 Neutral.

\paragraph{es-en} We used the development set from the Spanish-English corpus provided in the SentiMix competition \citep{patwaSemEval2020TaskOverview2020a}, partitioned by LinCE. This set includes 1,859 instances, categorized as follows: 1,037 Positive, 305 Negative, and 517 Neutral.

\paragraph{ta-en} The TamilMixSentiment \citep{chakravarthi-etal-2020-corpus} dataset consists of manually annotated Tamil-English code-mixed comments from YouTube. The test set, which we utilized, comprises 3,049 instances with the following distribution: 2,075 Positive, 424 Negative, 173 Neutral, and 377 Mixed feelings.

\paragraph{ml-en} \citet{chakravarthiSentimentAnalysisDataset2020} curated this Malayalam-English dataset from comments on 2019 Malayalam movie trailers on YouTube, with sentiment annotations performed by at least three trained annotators. We employed their test set, which includes 1171 instances: 565 Positive, 138 Negative, 398 Neutral, and 70 Mixed feelings.

\subsection{Datasets of MT Task}
\paragraph{zh-en $\rightarrow$ zh} \citet{calvilloSurprisalPredictsCodeSwitching2020b} employed five bilingual Chinese-English speakers to translate 3,022 sentences from the previously introduced zh-en dataset in the LID task into Chinese. These translators, international Chinese undergraduates, match the language proficiency and cultural background of the CSSA BBS forum users.
\paragraph{hok-zh $\rightarrow$ zh} Given that the Hokkien-Mandarin dataset is synthesized from parallel corpora \citep{luExploringMethodsBuilding2022a}, it allows for the straightforward construction of a translation task utilizing both the synthesized data and the original data. As a result, we have developed a dataset comprising 3,800 samples, facilitating the translation of Hokkien-Mandarin into Mandarin.

\paragraph{hi-en $\rightarrow$ en} \citet{chenCALCS2021Shared2022} created a translation task from English to Hinglish at the 2021 CALCS workshop, using a subset of the CMU Document Grounded Conversations dataset. We utilized its development set and converted it into a Hinglish-to-English translation task, comprising 942 instances.

\paragraph{bn-en $\rightarrow$ en} \citet{vavreAdaptingMultilingualModels2022} proposed a dataset for translating Bengali-English texts to English, sourced from the Spoken Tutorial project. This dataset includes transcriptions from video lectures collected from the Spoken Tutorial educational website, as well as parallel sentences from the Samanantar project and other sources. On average, each sentence contains 11.32 Bengali tokens and 13.31 English tokens. We selected the ST-Hard subset for testing, which comprises 2000 sentences where the baseline model performed the poorest.

\paragraph{mr-en $\rightarrow$ en} \citet{vavreAdaptingMultilingualModels2022} also introduced a Marathi-English code-mixed to English translation task, sourced from the Spoken Tutorial project. Each sentence in this dataset averages 11.32 Marathi tokens and 13.00 English tokens. We similarly selected the ST-Hard subset, containing 2000 sentences.

\section{Automatic LID Annotation}
\label{app:annotation}
Our method for word-level Language Identification annotation is simple and effective, utilizing the GPT-4 Turbo model without relying on extra dictionaries. Based on the parallel sentences (\textit{L1}, \textit{L2}), we instruct the model to replace tokens from \textit{L1} with corresponding tokens from \textit{L2}, to synthesize code-mixed sentences (\textit{CM}). We identify tokens' LID tags as follows: tokens from \textit{L1} not present in \textit{L2} are marked as the first language, tokens from \textit{L2} not present in \textit{L1} are marked as the second language, and if tokens belonging to both \textit{L1} and \textit{L2} we consider this token to be language-independent and mark it as "other". This approach is particularly effective for languages with distinct character sets. However, for languages sharing the same script, such as English and French, this method may inaccurately label shared tokens as "other". To resolve this issue, we also instruct the model to return all the replaced tokens, forming set \textit{X}.  If a token in the code-mixed sentence comes from \textit{X}, we mark it as the second language. This automatic annotation technique is suitable for large-scale multi-language annotation tasks. We designed regular expressions to tokenize sentences into words, and for Chinese text, we use the Jieba\footnote{https://github.com/fxsjy/jieba} tokenizer.

\section{Semantic Filtering using LaBSE}
\label{app:labse}
\input{tables/labse}
\revisedText{The Language-agnostic BERT Sentence Encoder (LaBSE) is a BERT-based model trained for sentence embeddings in 109 languages. As shown in Table \ref{tab:labsescores}, LaBSE scores are high because GPT generated code-mixed sentences by replacing corresponding parts in parallel sentences, maintaining their original structure with minor linguistic changes. Our experiments demonstrated LaBSE's stability in computing semantic similarity scores for code-mixed sentences. We also sampled 20 examples from each of the 11 language pairs in CM-MMLU and manually verified LaBSE's evaluation of the synthetic data. Our manual reviews align closely with LaBSE's high scores, likely because the synthetic data was generated using simple word substitution by a powerful GPT model, which minimally impacted the source text's semantics. Consequently, we used LaBSE for batch evaluation of synthetic data quality.}

\section{Model Aligned Filtering using GPT-4}
\label{app:gpt4valid}
\input{tables/gpt4valid}
\revisedText{We employed the robust GPT-4 turbo, incorporating detailed scoring guidelines and the Chain-of-Thought (CoT) methodology within the prompt (see Appendix \ref{app:humanlevel}) to guide the model in performing analysis before assigning a final score, thereby enhancing the reliability of the assessment. We use the Mean Absolute Percentage Error (MAPE) formula to compute the differences between GPT and human scores across three dimensions: coherence, naturalness, and readability, where n equals 3. 
\[
\text{MAPE} = \frac{1}{n} \sum_{i=1}^{n} \left| \frac{\text{GPT\_score}_i - \text{human\_score}_i}{\text{human\_score}_i} \right| \times 100\%
\]
\[
\text{Agreement} = 1 - \text{MAPE}
\]
We sampled 20 instances from each language pair in the CM-MMLU dataset and manually reviewed the model-aligned evaluation results, achieving a 91.4\% average agreement rate. Table \ref{tab:gpt4valid} displays four randomly selected examples. While GPT cannot fully replace human evaluators, it can process large volumes of data in batches and achieve a high degree of consistency with human assessments.}

\section{Prompts for Building {\Name}}
\subsection{Synthesis Prompt}
\label{app:synthesis}
\begin{metaverbatim}

You will receive a pair of parallel mo-
nolingual sentences. Randomly replace 
certain words or phrases in the first 
sentence with corresponding parts from 
the second sentence to synthesize them 
into a codemixed sentence. Finally, ou-
tput the code-mixed sentence and words 
or phrases you have replaced in the 
following format:

Code-mixed sentence: …
Replaced parts: 
<<<word/phrase-->word/phrase>>>, 
<<<word/phrase-->word/phrase>>>, …
\end{metaverbatim}

\subsection{Model-Aligned Filtering Prompt}
\label{app:humanlevel}
\begin{metaverbatim}

You will be presented with a code-mixed 
sentence. Your task is to evaluate the 
sentence based on three separate met-
rics. Assuming the readers are people 
familiar with each language in the se-
ntence.

Evaluation Criteria:
Coherence (1-3): Assesses how well the 
sentence elements are connected and fl-
ow together, considering the mixing of 
languages.
1: Poor. The sentence lacks logical fl-
ow or connection between its parts, ma-
king it hard to understand.
2: Fair. The sentence has some logical 
connections between its parts, but the 
flow might be interrupted by awkward 
language mixing.
3: Good. The sentence demonstrates a 
clear and logical connection between 
its parts, with the mixing of langua-
ges not hindering understanding.

Naturalness (1-3): Evaluate the sent-
ence for its natural-sounding language 
use and integration of the code-mixed 
elements.
1: Poor. The sentence sounds unnatural 
or forced, with the mixing of languages 
seeming out of place.
2: Fair. The sentence sounds somewhat 
natural, though the integration of di-
fferent languages can occasionally 
feel awkward.
3: Good. The sentence sounds natural 
and the mixing of languages appears 
seamless and intentional.

Readability (1-3): Measures how easy 
it is to read and understand the sent-
ence, considering the impact of code-
mixing on readability.
1: Poor. The sentence is difficult to 
read, with the mixing of languages 
significantly hindering comprehension.
2: Fair. The sentence is readable, 
though the reader may need to pause 
to understand the mixed languages.
3: Good. The sentence is easy to re-
ad, with the code-mixing enhancing or 
not detracting from the ability to 
understand the content.

Output your evaluation following this 
format:
Concise and refined evaluation analysis: 
…
Scores (only scores): coherence score, 
nat-uralness score, readability score.
\end{metaverbatim}

\section{Prompts of Experiment}

\label{app:experimentPrompt}
\subsection{Prompt of LID, POS, NER Task}
\label{app:experimentPromptLID}

\begin{metaverbatim}

You are a smart and intelligent [INPUT 
TASK] system. You will receive a toke-
nized sentence code-mixed with [INPUT 
FIRST LANGUAGE] and [INPUT SECOND LAN-
GUAGE]. Label each token in the token-
ized sentence based on the categories: 
[tag_1, tag_2, … , tag_k]

You must tag every token in the token-
ized sentence in order, without skipp-
ing or missing any token for any reason. 
Fill in this JSON format: [{specific 
token_1: tag_k}, {specific token_2: 
tag_k}].

Please refer to the example: 
Tokenized sentence: [INPUT A CODE-MIXED 
SENTENCE]
Your answer: [JSON FORMAT].

Tokenized sentence: [INPUT A CODE-MIXED 
SENTENCE];
Your answer:
\end{metaverbatim}

\subsection{Prompt of SA Task}
\label{app:experimentPromptSA}
\begin{metaverbatim}

You are a smart and intelligent sent-
iment analysis (SA) system. I will gi-
ve you a code-mixed sentence that has 
been mixed with [INPUT FIRST LANGUAGE] 
and [INPUT SECOND LANGUAGE]. Assign 
the appropriate label from: [tag_1, 
tag_2, …, tag_k].

Please refer to the example: 
Sentence: [INPUT A CODE-MIXED SENTENCE]
Your answer: [INPUT A TAG]

Sentence: [INPUT A CODE-MIXED SENTENCE]
Your answer: 
\end{metaverbatim}

\subsection{Prompt of MT Task}
\label{app:experimentPromptMT}
\begin{metaverbatim}

You will receive a sentence code-mixed 
with [INPUT FIRST LANGUAGE] and [INPUT 
SECOND LANGUAGE]. Translate the given 
sentence into [INPUT TARGET LANGUAGE]. 

Please refer to the example: 
Sentence: [INPUT A CODE-MIXED SENTENCE]
Your answer: [INPUT TARGET SENTENCE]

Sentence: [INPUT A CODE-MIXED SENTENCE]
Your answer: 
\end{metaverbatim}

\subsection{Prompt of CM-MMLU Task}
\label{app:experimentPromptMMLU}
\begin{metaverbatim}

You are a system possessing knowledge 
in all subjects. You are skilled at 
selecting the correct answer based on 
multiple-choice questions. Do not in-
clude explanations in your answer. 

(k-shot setting here)
Question: [INPUT MULTIPLE-CHOICE QUE-
STION]
Answer: [INPUT ANSWER]
...

Question: [INPUT MULTIPLE-CHOICE QUE-
STION]
Answer: 
\end{metaverbatim}

\subsection{Prompt of CM-GSM8K Task}
\label{app:experimentPromptGSM8K}
\begin{metaverbatim}

You are skilled at solving mathemati-
cal problems. Output the solution and 
final answer for the next problem. 
The solution should include the entire 
process of calculating the final ans-
wer. The final answer to the problem 
is just one definite numerical value. 
Don't output the problem. 
Output in this format:
Solution:
Final answer: (one definite numerical 
value)

(k-shot setting here)
Problem: [INPUT MATH PROBLEM]
Solution: [INPUT CoT SOLUTION]
Final answer: [INPUT FINAL ANSWER]
...

Problem: [INPUT MATH PROBLEM]
Solution: 
Final answer: 
\end{metaverbatim}

\subsection{Prompt of CM-TruthfulQA Task}
\label{app:experimentPromptTruthfulQA}
\begin{metaverbatim}

You are skilled at selecting the cor-
rect answer based on multiple-choice 
questions. Do not include explanati-
ons in your answer.

(k-shot setting here)
Question: [INPUT MULTIPLE-CHOICE QUE-
STION]
Answer: [INPUT ANSWER]
...

Question: [INPUT MULTIPLE-CHOICE QUE-
STION]
Answer: 
\end{metaverbatim}

\section{Statistics of Synthetic Datasets}
\label{app:statistics}
\begin{figure}[t]
  \centering
  \includegraphics[width=0.8\columnwidth]{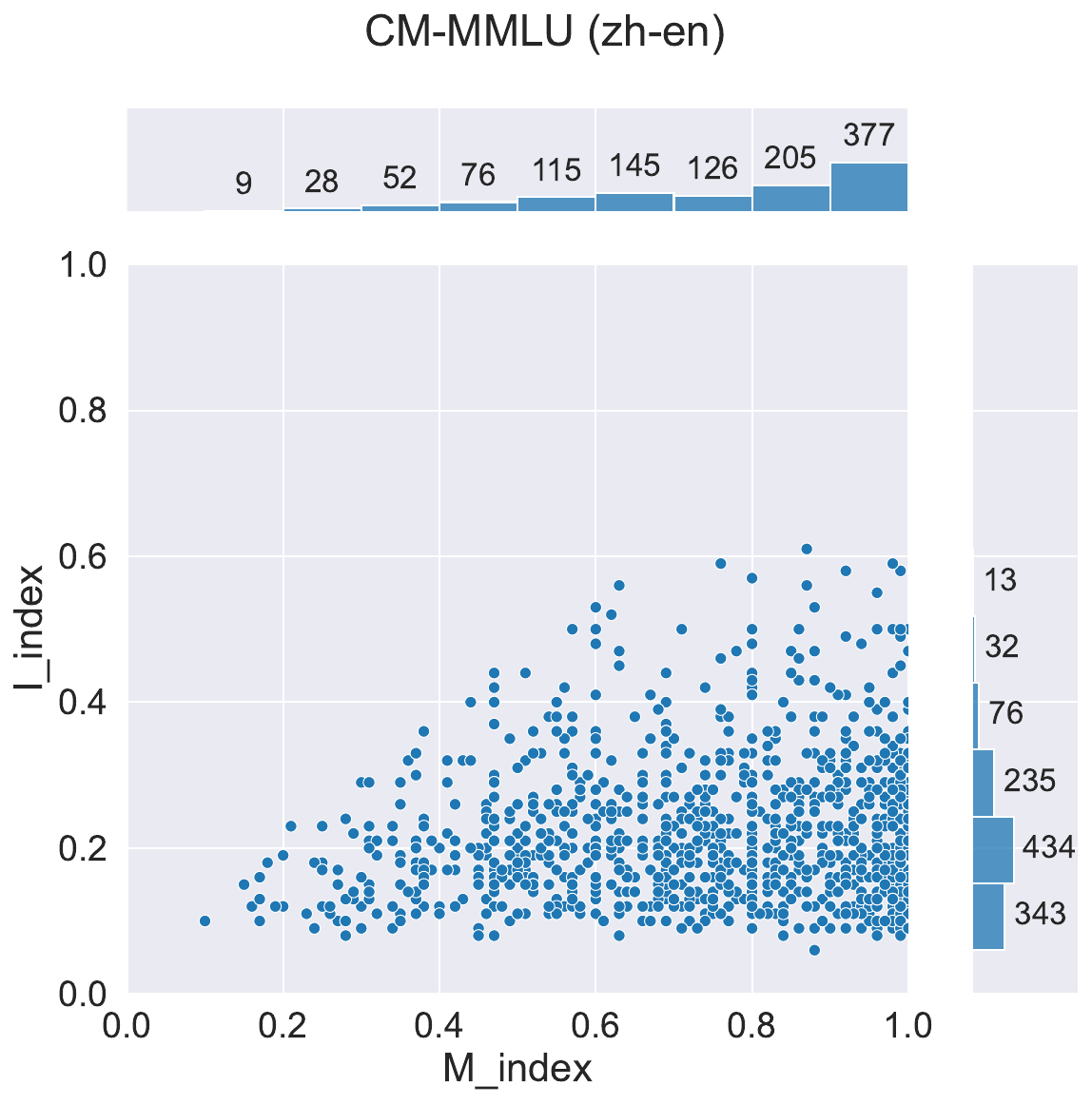}
  \caption{\textbf{The distribution of 1133 samples in the code-mixed (zh-en) MMLU.} Two histograms are added around the scatter plot in this figure. The scatter plot displays the M-index and I-index for each sample. The histograms represent the distributions of two metrics.}
  \label{fig:distribution}
\end{figure}

We synthesize CM-MMLU (11 language pairs), CM-GSM8K (4 pairs), CM-TruthfulQA (4 pairs), and MT tasks (3 pairs), detailed in Table~\ref{tab:statistics}. 
We observe that each dataset contains an average of 1,016 samples, with token counts of 24,543, 19,897, and 3,330 for two languages and language-independent tokens (i.e. punctuation, numerals, and formulas), respectively. Both Semantic and Model-Aligned evaluations show high scores. The weighted average M-index across 22 datasets is 0.81, indicating a balanced proportion of the two languages within the text. The average I-index of 0.25 meets our expectations, as a high I-index would not represent realistic code-mixing. Imagining a sentence code-mixed with Chinese and English like "我们 will 走 very 长的 journey, 所以 we 得 bring 足够的 food 和 water" (We will take a very long journey, so we need to bring enough food and water). The sentence has both the M-index and the I-index equal to 1 but is difficult to read and appears unrealistic.
For the single dataset, we analyzed the distributions of the M-index and I-index metrics within the dataset. One dataset (\textit{es-en} of CM-MMLU) is illustrated in Figure~\ref{fig:distribution} and others are shown in 
Figure \ref{fig:additionalDistribution}. In summary, our statistical analysis indicates that our synthesized dataset demonstrates sufficient code-mixing between pairs of languages while preserving coherence, naturalness, readability, and a high degree of similarity to the original task sentences. We spent a total cost of \$718.45 to construct these datasets.

\input{tables/statistics}


\begin{figure*}[h]
    \subfigure{
      \includegraphics[width=0.32\textwidth]{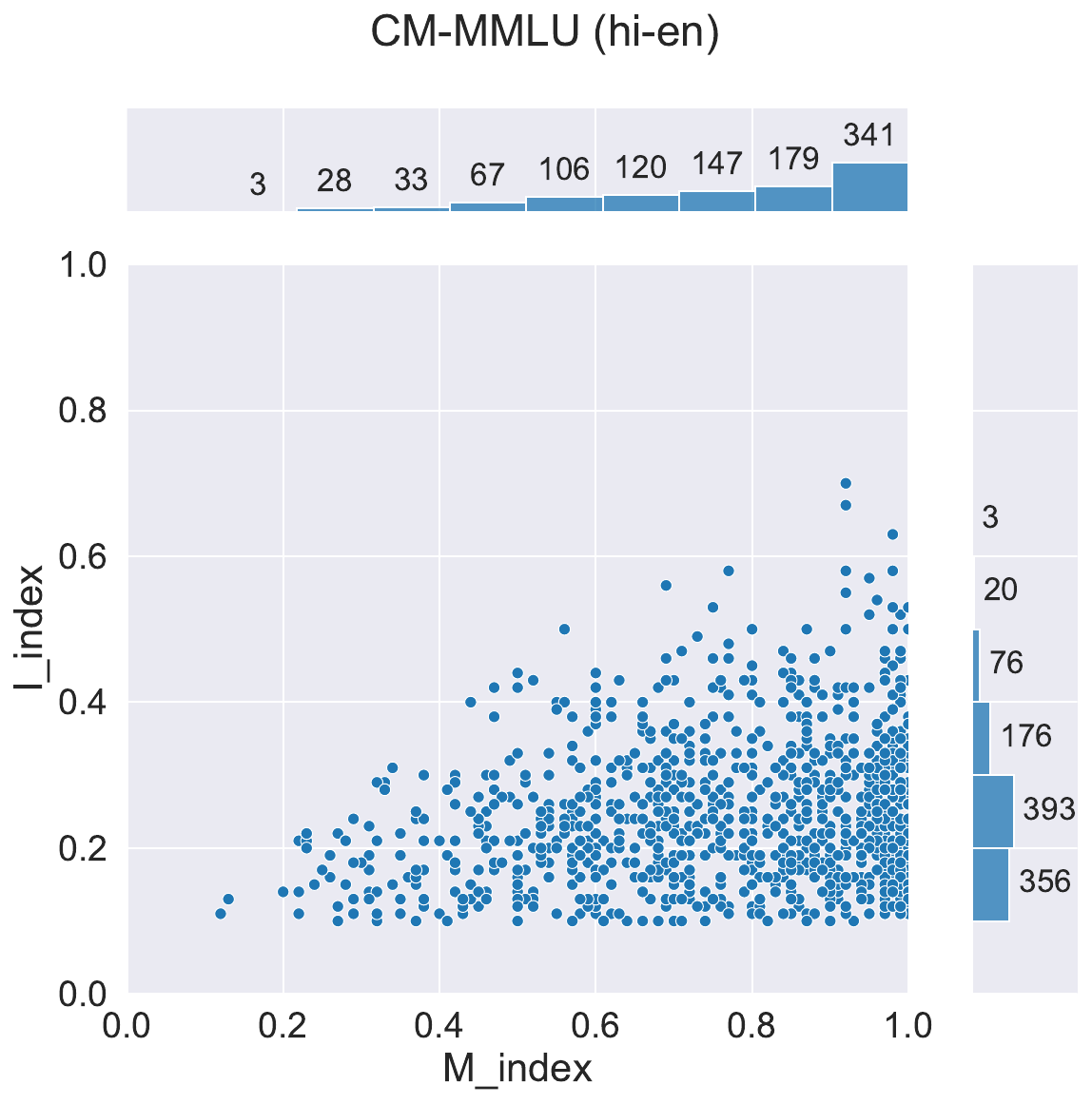}
      \label{fig:mmlu_hi}
    }
    \subfigure{
      \includegraphics[width=0.32\textwidth]{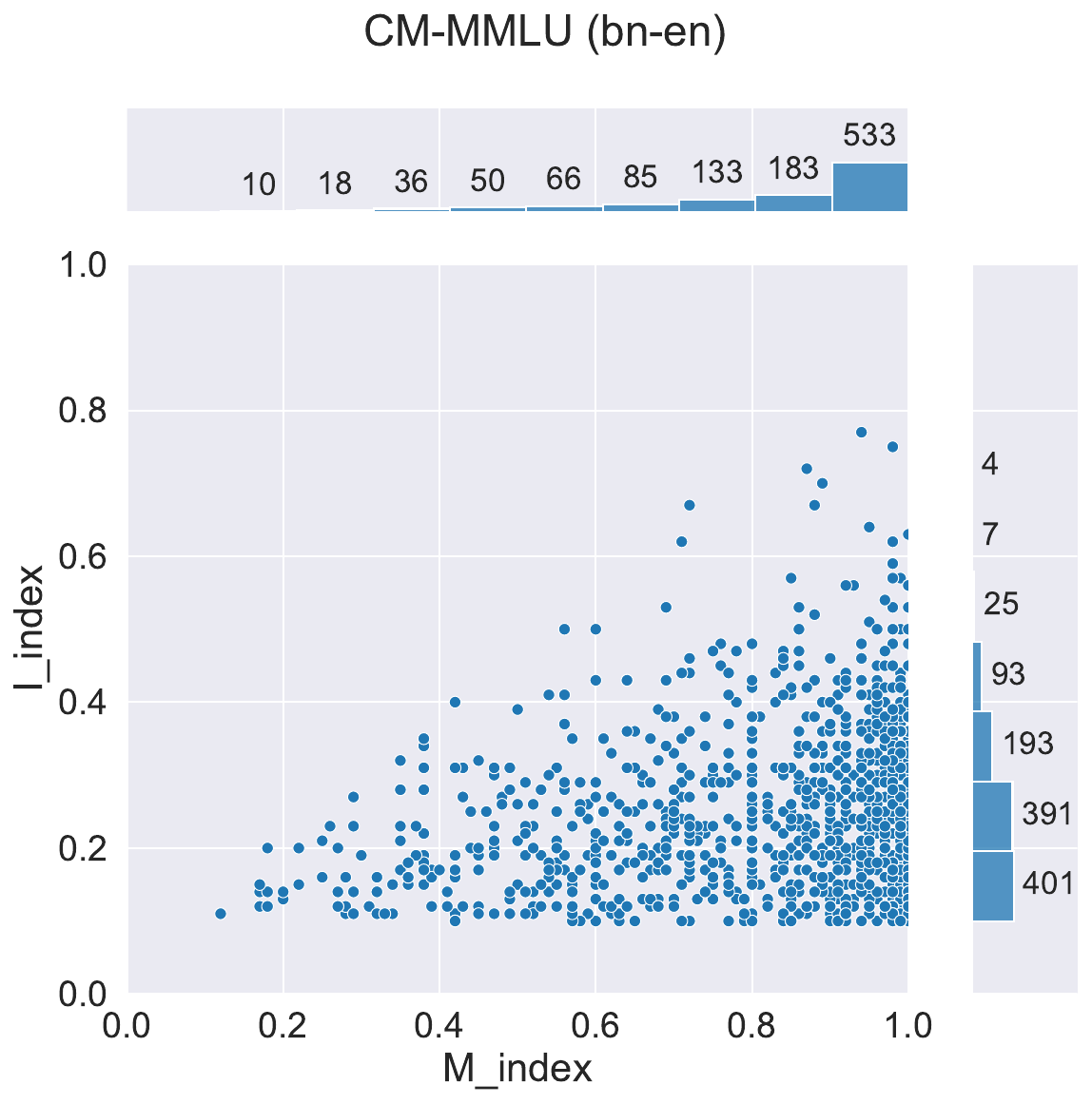}
      \label{fig:mmlu_bn}
    }
    \subfigure{
      \includegraphics[width=0.32\textwidth]{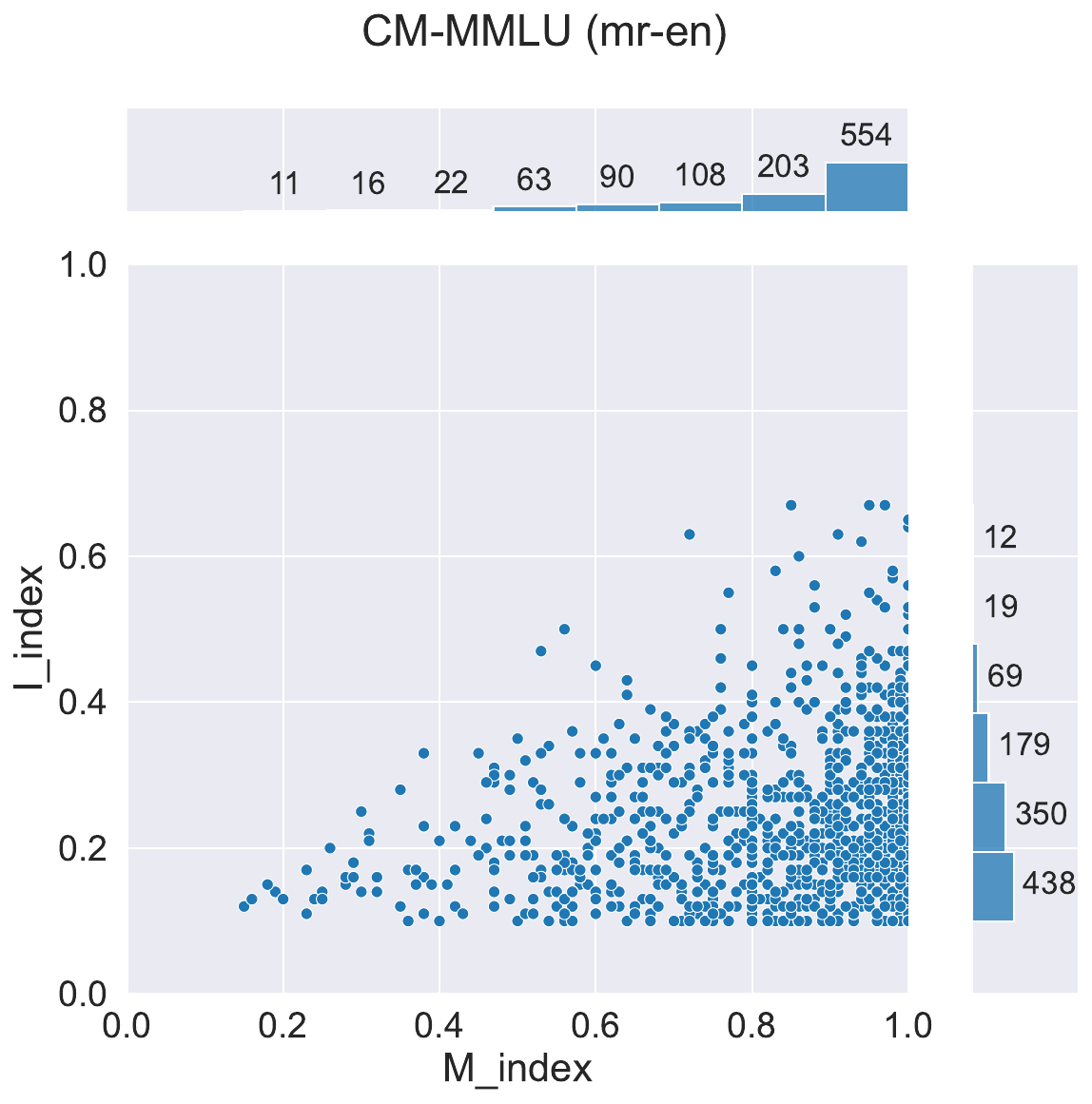}
      \label{fig:mmlu_mr}
    }
    \subfigure{
      \includegraphics[width=0.32\textwidth]{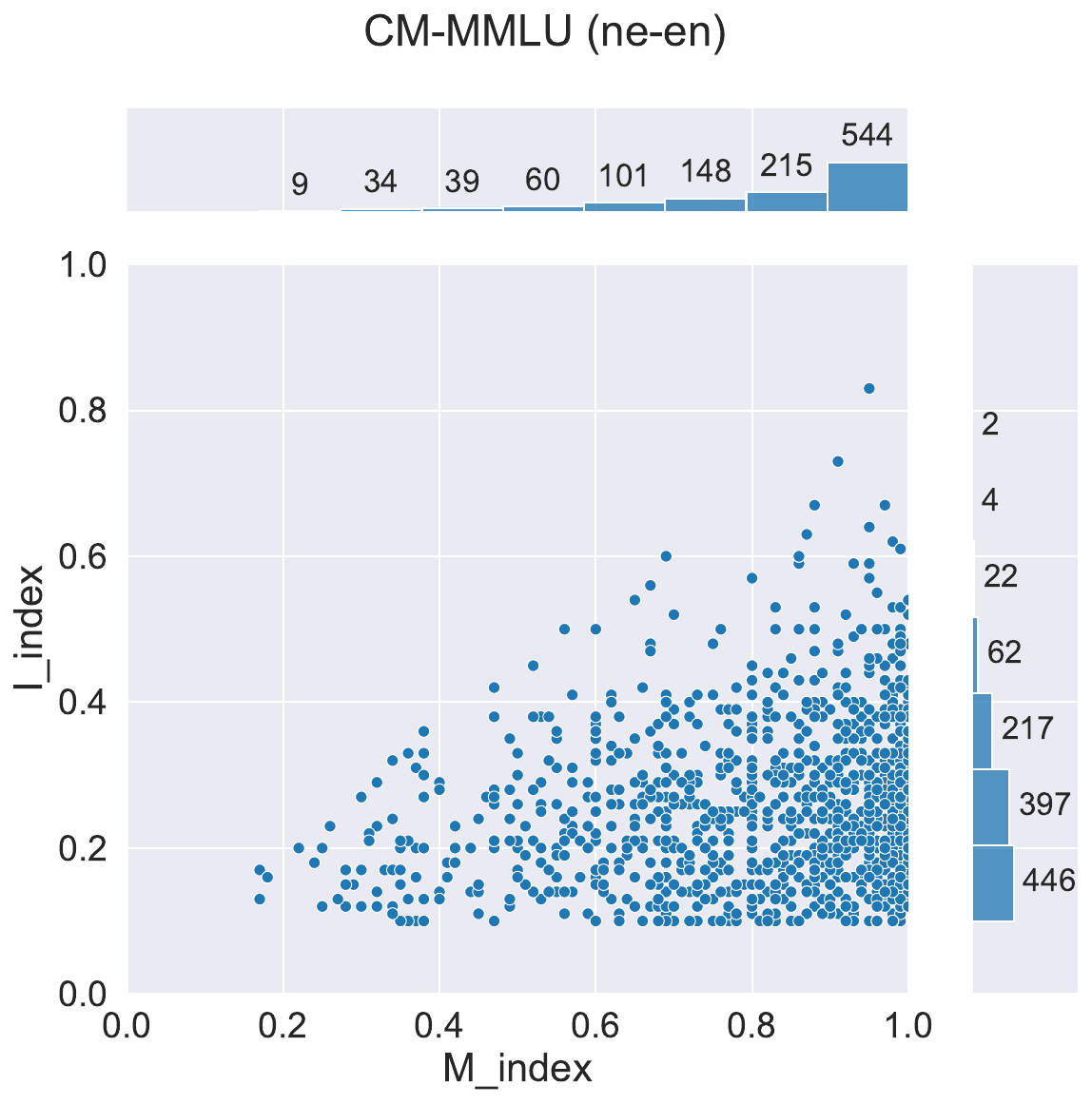}
      \label{fig:mmlu_ne}
    }
    \subfigure{
      \includegraphics[width=0.32\textwidth]{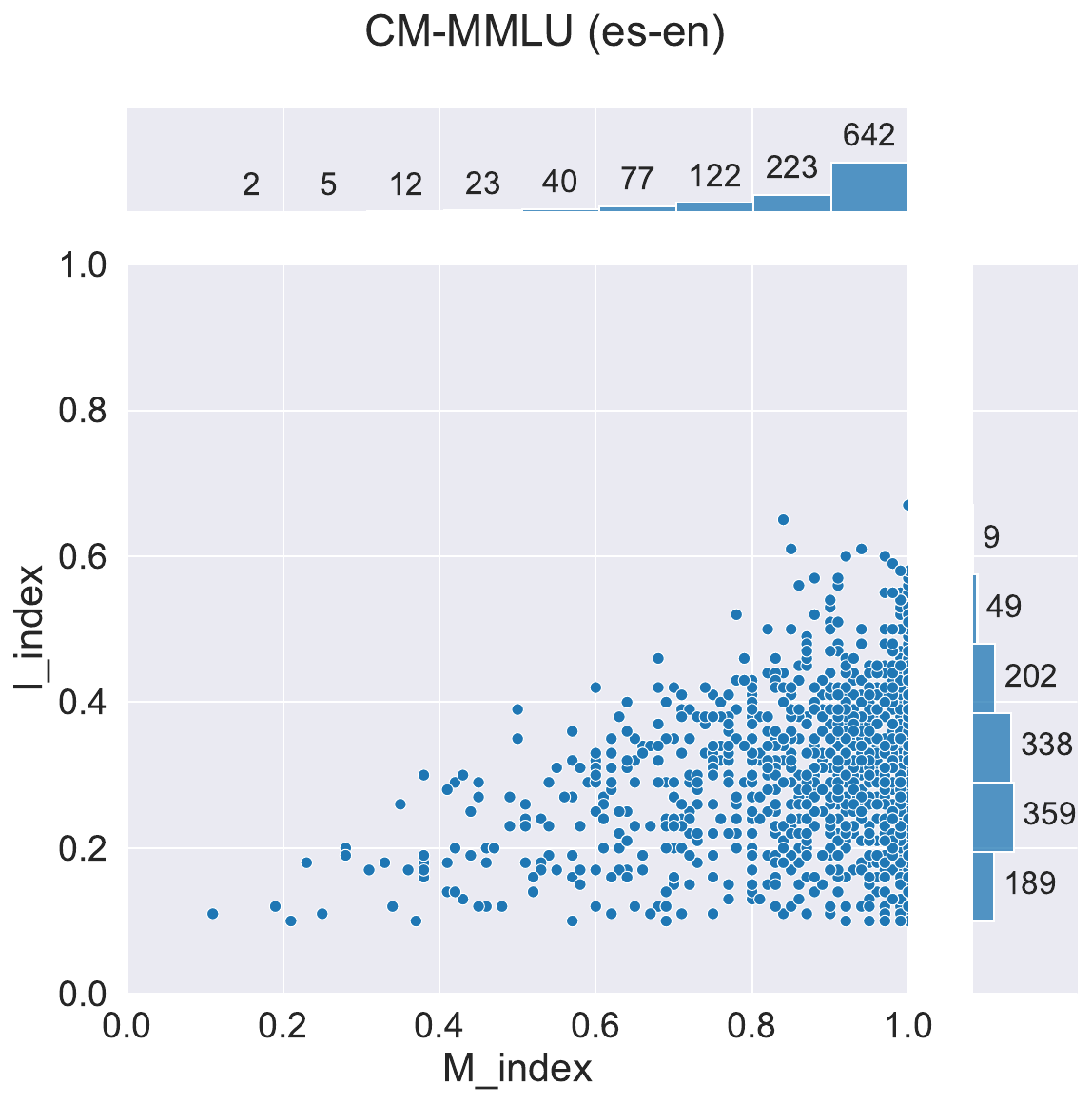}
      \label{fig:mmlu_es}
    }
    \subfigure{
      \includegraphics[width=0.32\textwidth]{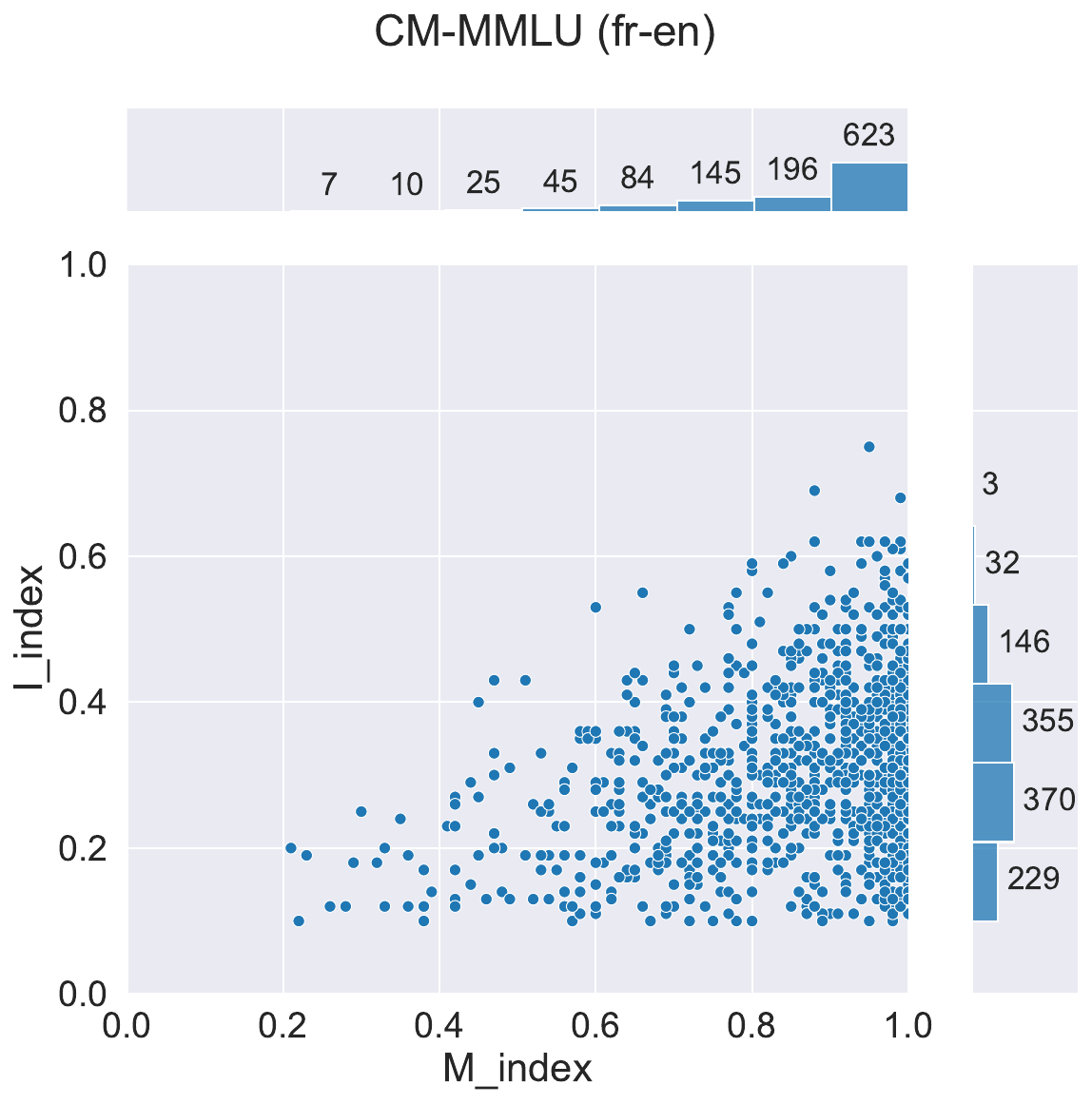}
      \label{fig:mmlu_fr}
    }
    \subfigure{
      \includegraphics[width=0.32\textwidth]{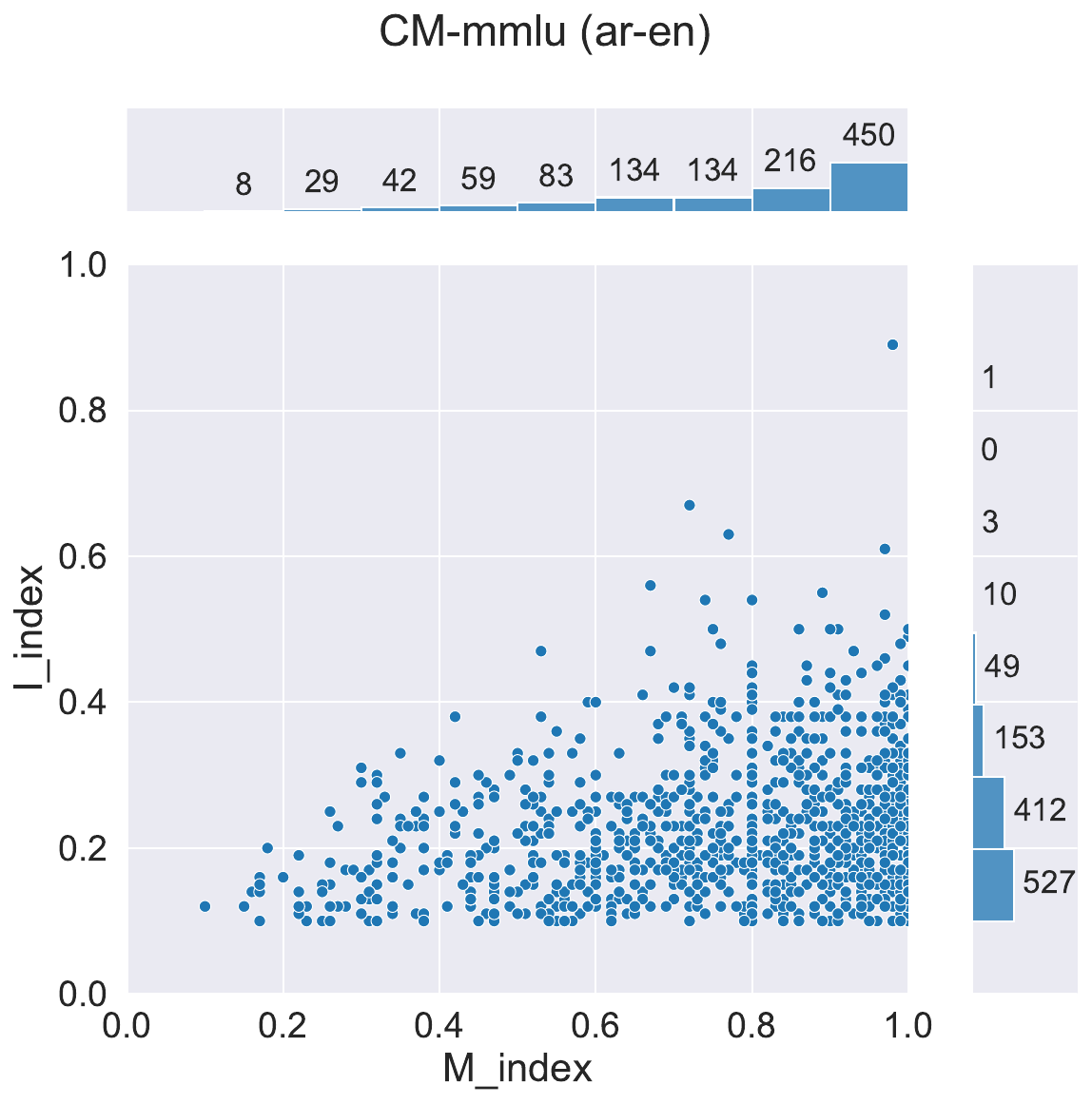}
      \label{fig:mmlu_ar}
    }
    \subfigure{
      \includegraphics[width=0.32\textwidth]{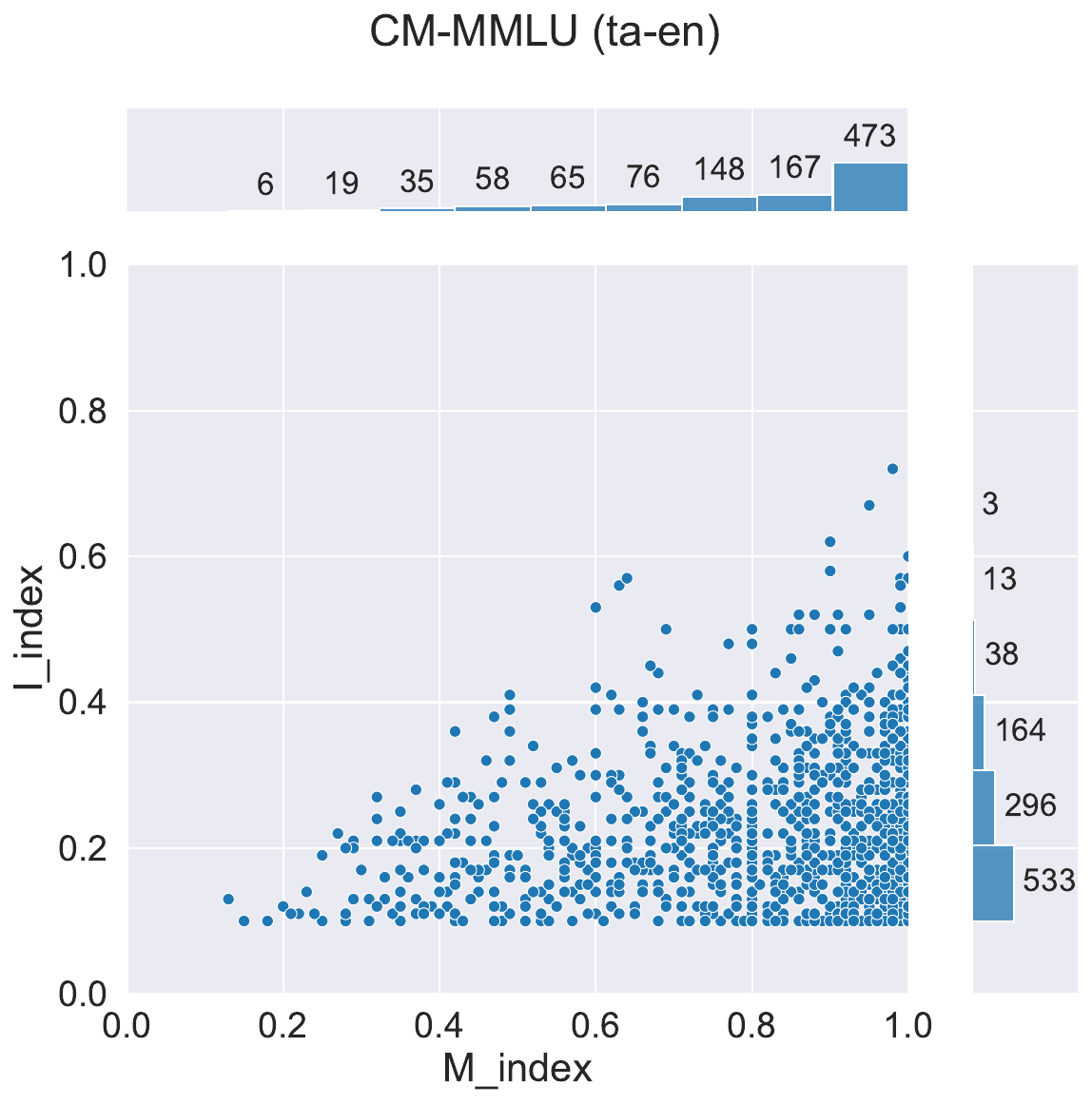}
      \label{fig:mmlu_ta}
    }
    \subfigure{
      \includegraphics[width=0.32\textwidth]{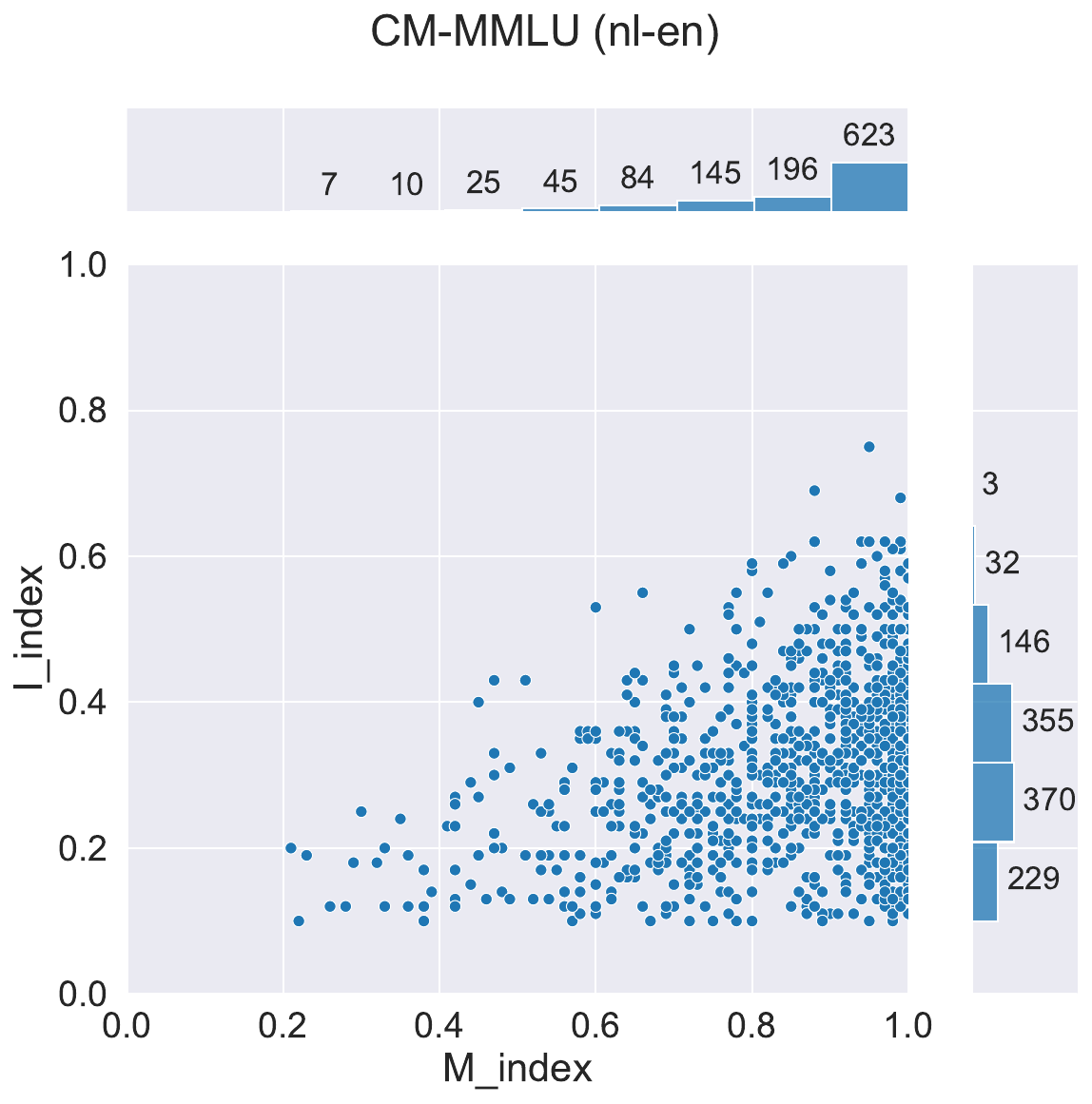}
      \label{fig:mmlu_nl}
    }
    \caption {\textbf{Distribution of Additional synthetic code-mixing datasets in {\Name}.}}
    \label{fig:additionalDistribution}
\end{figure*}
\label{app:distribution}
\begin{figure*}[h]
    \ContinuedFloat
    \subfigure{
      \includegraphics[width=0.32\textwidth]{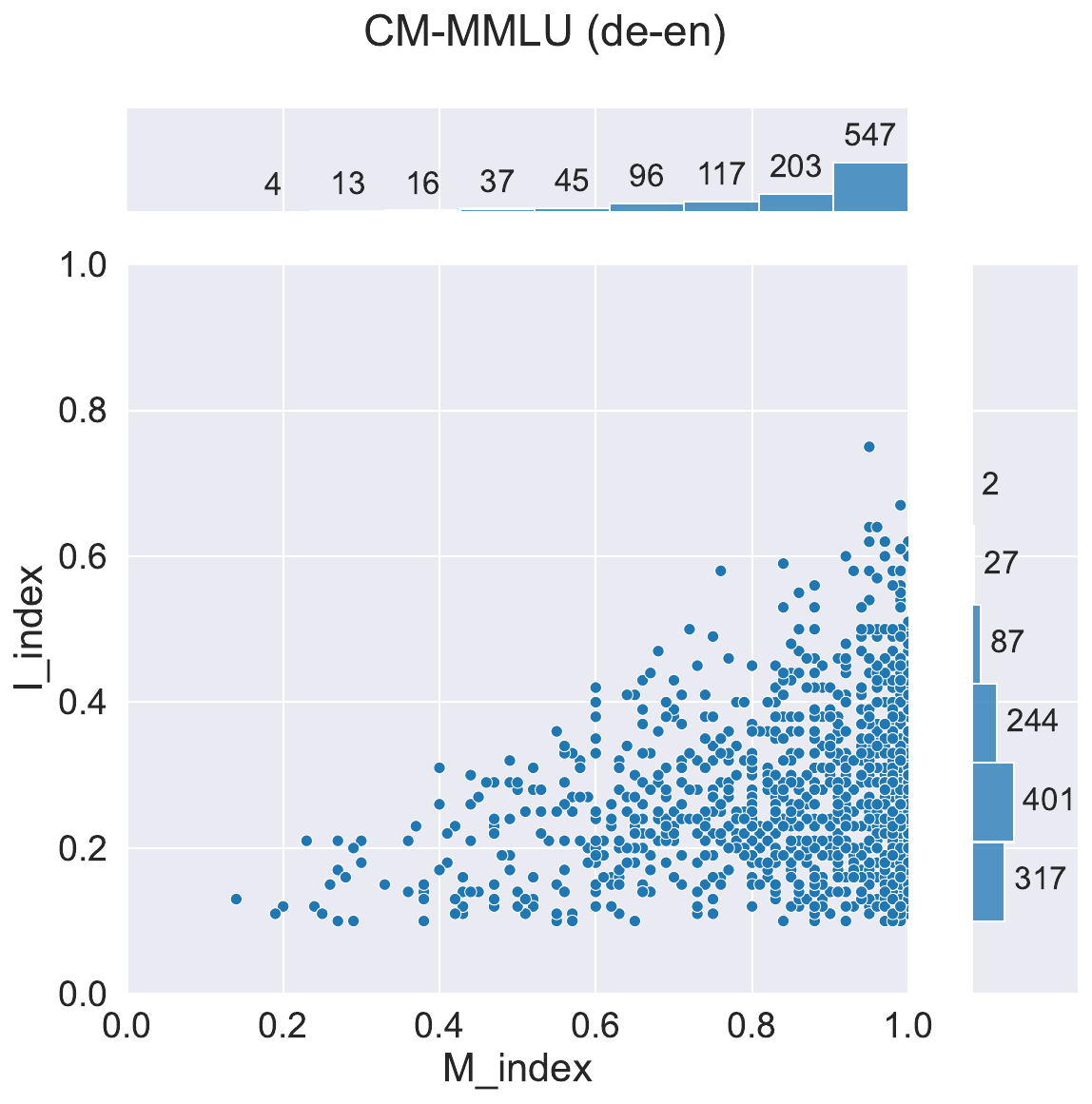}
      \label{fig:mmlu_de}
    }
    \subfigure{
      \includegraphics[width=0.32\textwidth]{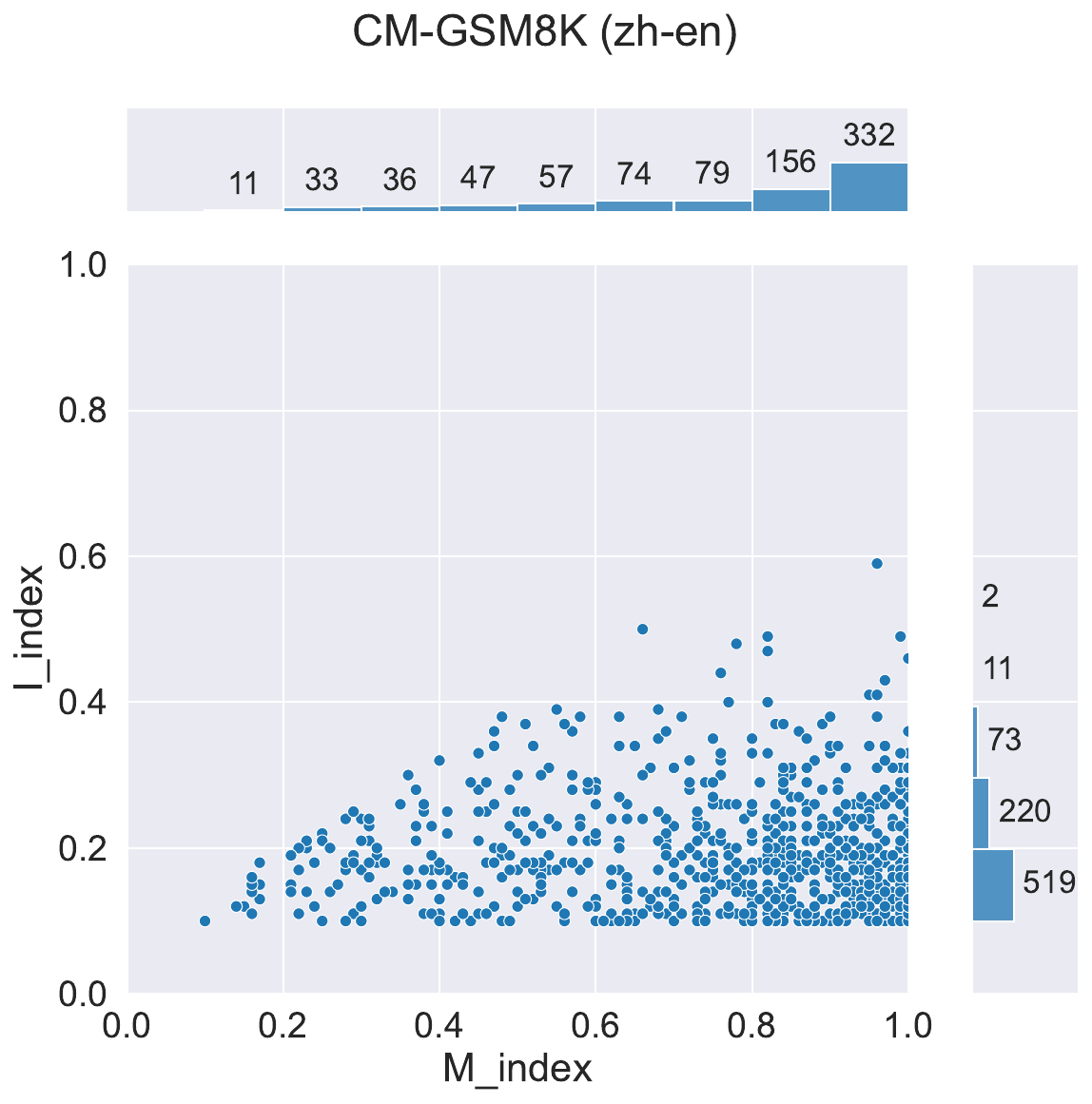}
      \label{fig:gsm8k_zh}
    }
    \subfigure{
      \includegraphics[width=0.32\textwidth]{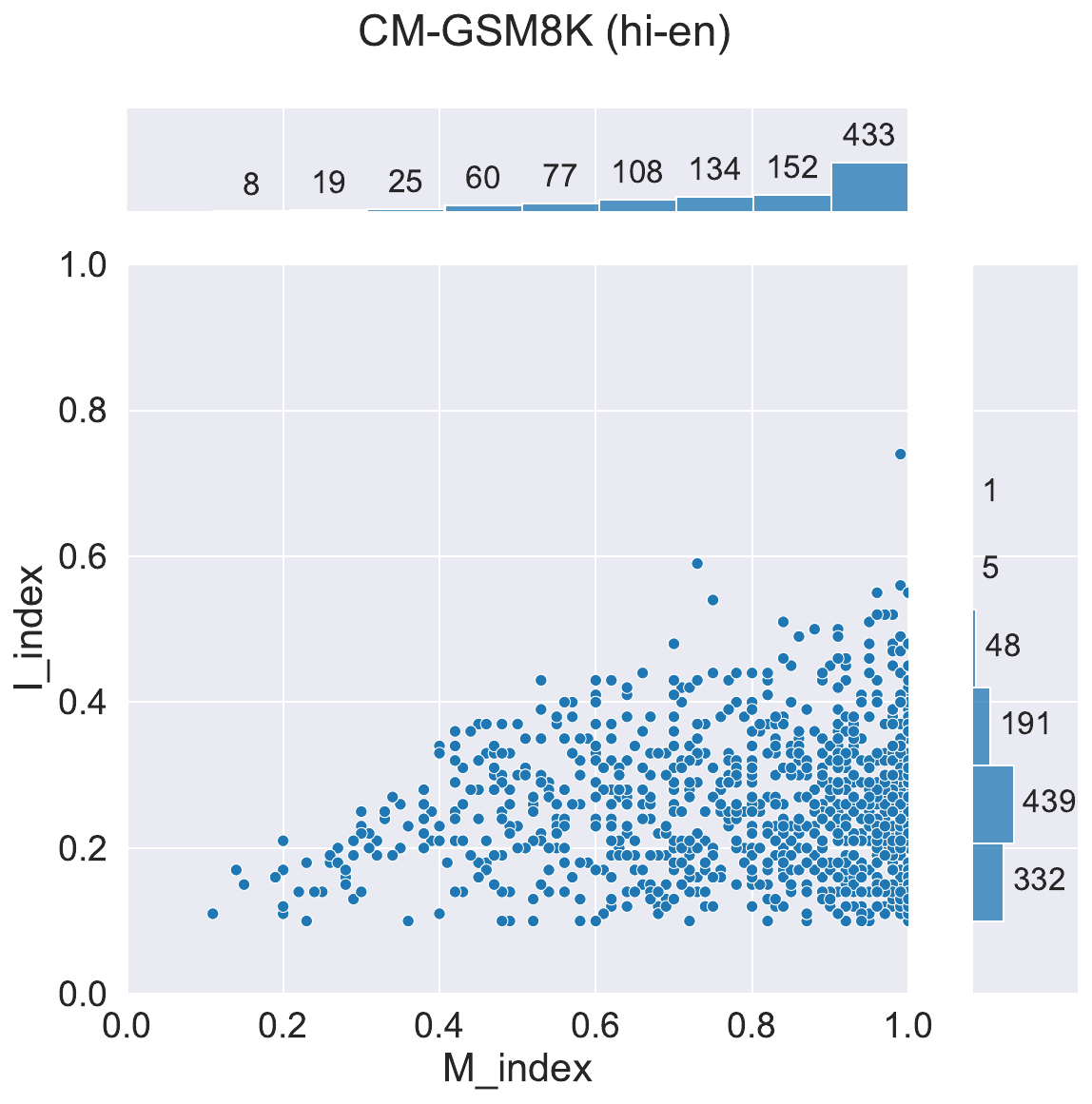}
      \label{fig:gsm8k_hi}
    }
    \subfigure{
      \includegraphics[width=0.32\textwidth]{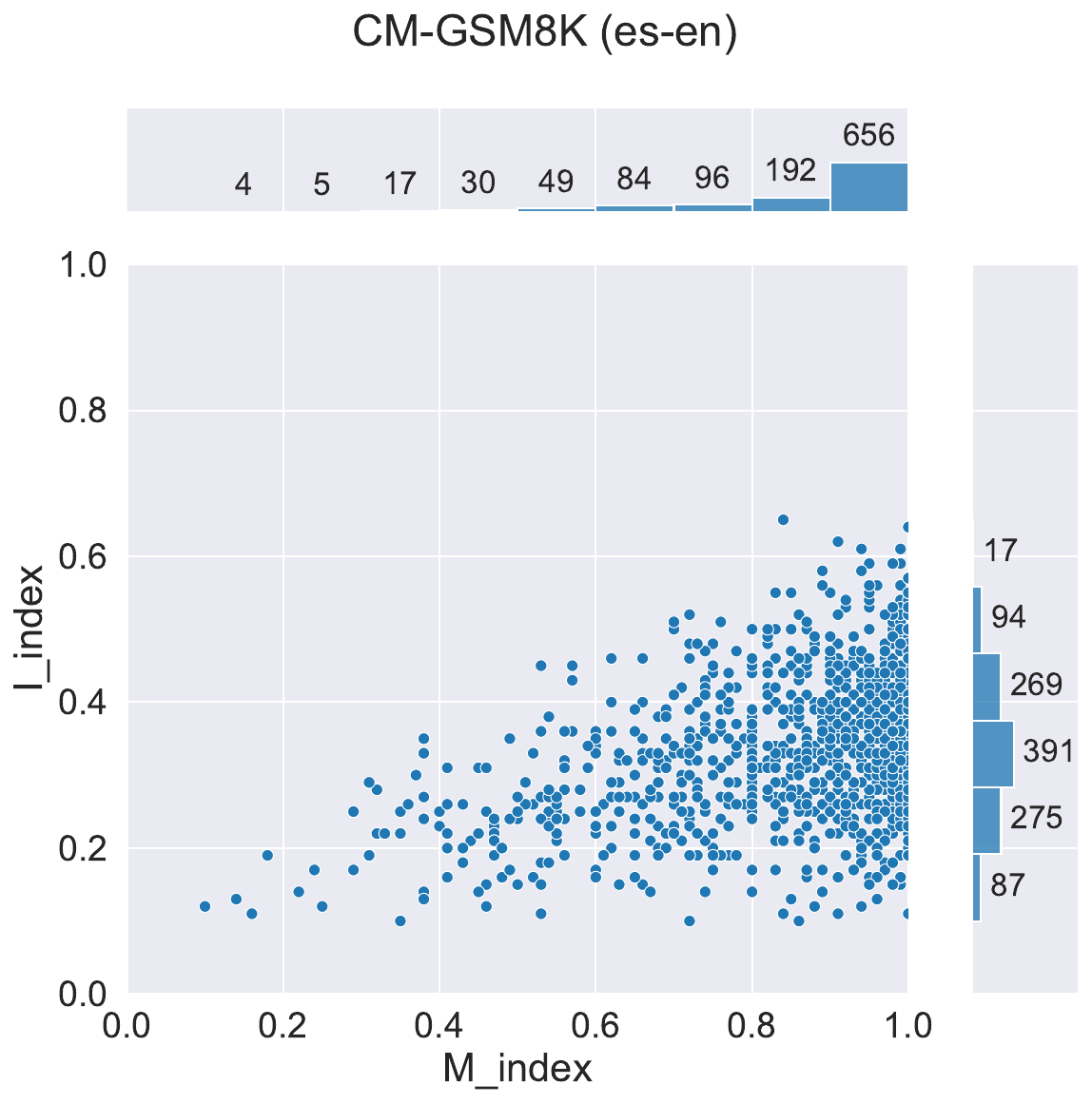}
      \label{fig:gsm8k_es}
    }
    \subfigure{
      \includegraphics[width=0.32\textwidth]{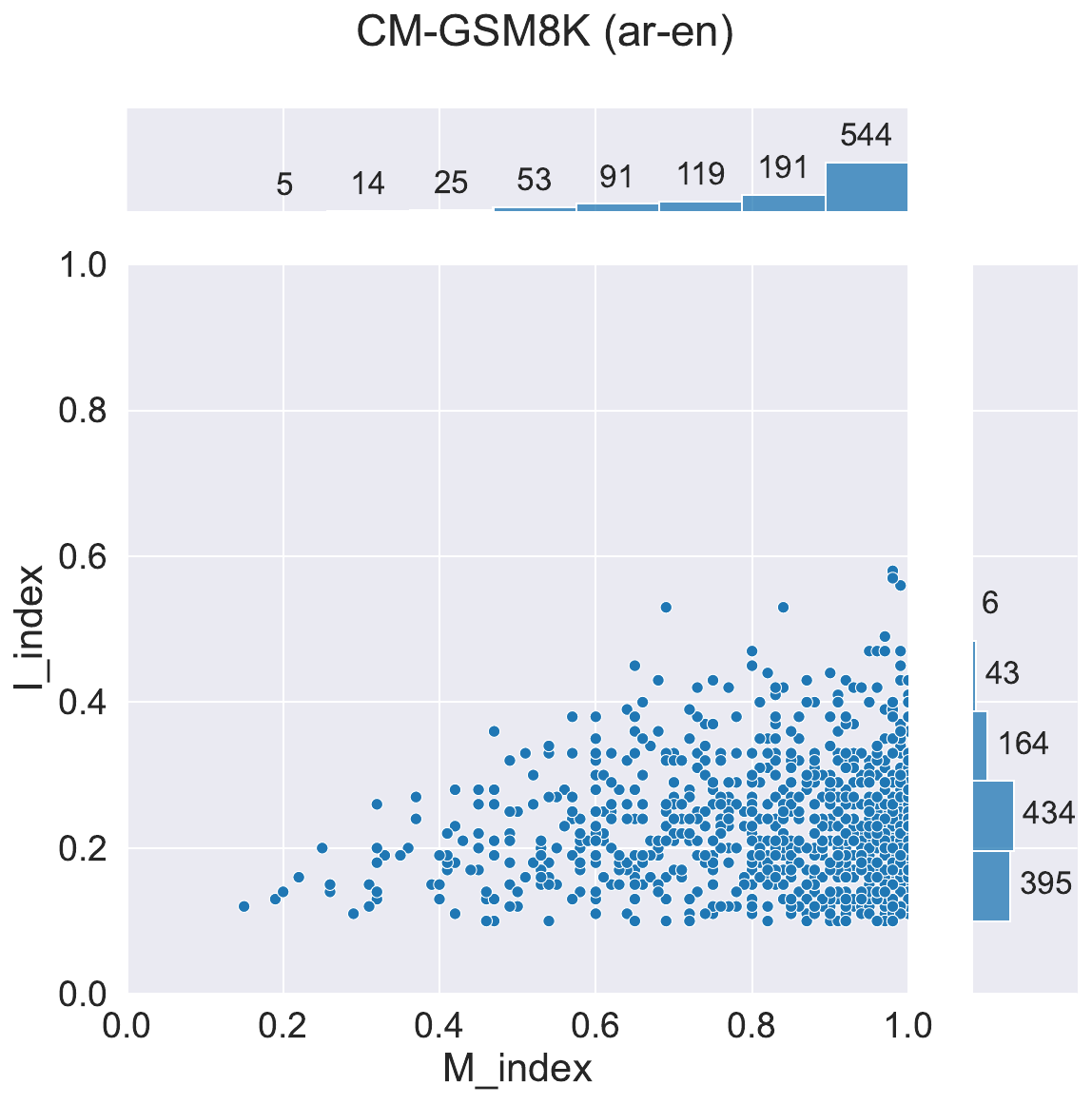}
      \label{fig:gsm8k_ar}
    }
    \subfigure{
      \includegraphics[width=0.32\textwidth]{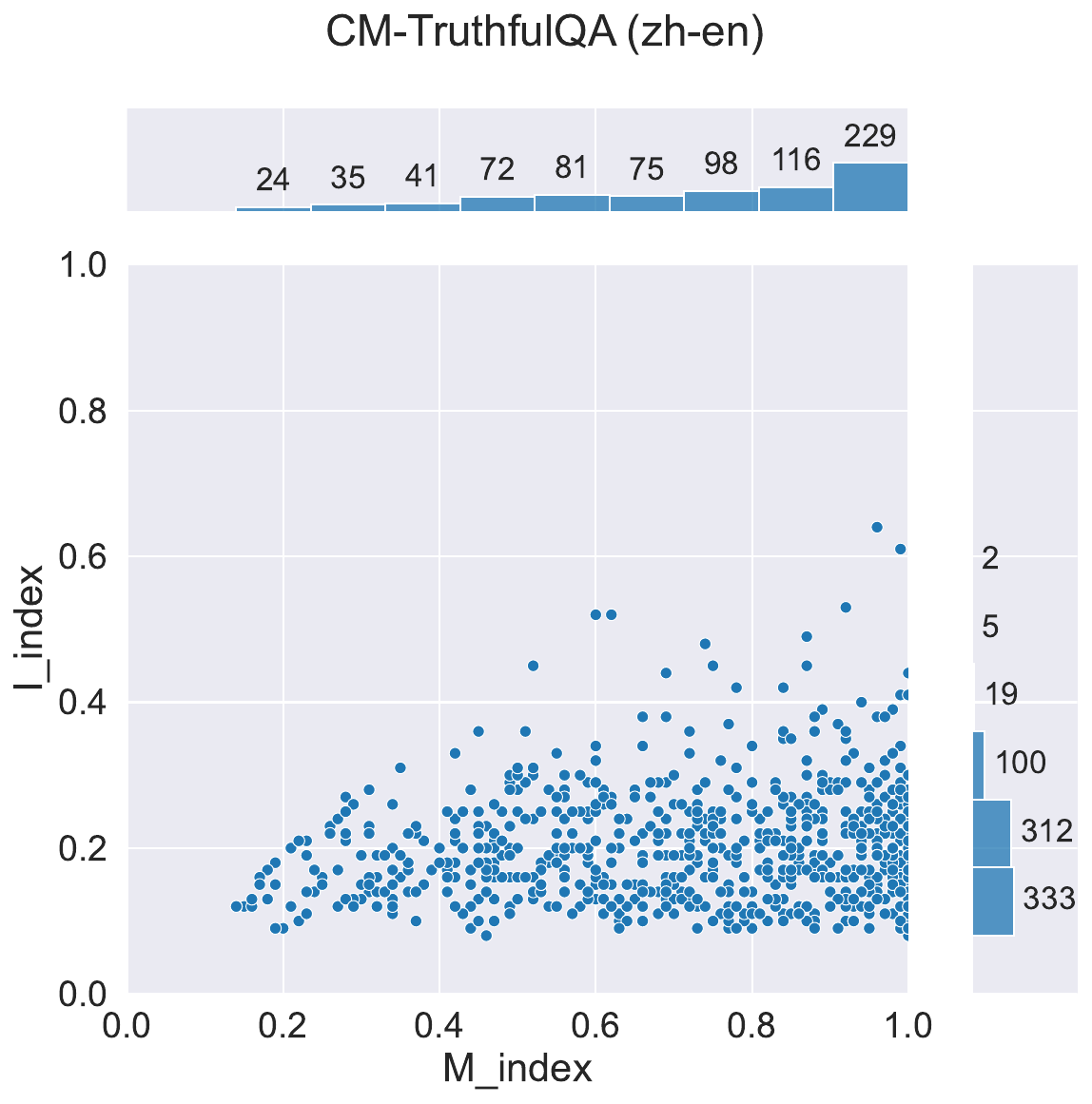}
      \label{fig:truthfulqa_zh}
    }
    \subfigure{
      \includegraphics[width=0.32\textwidth]{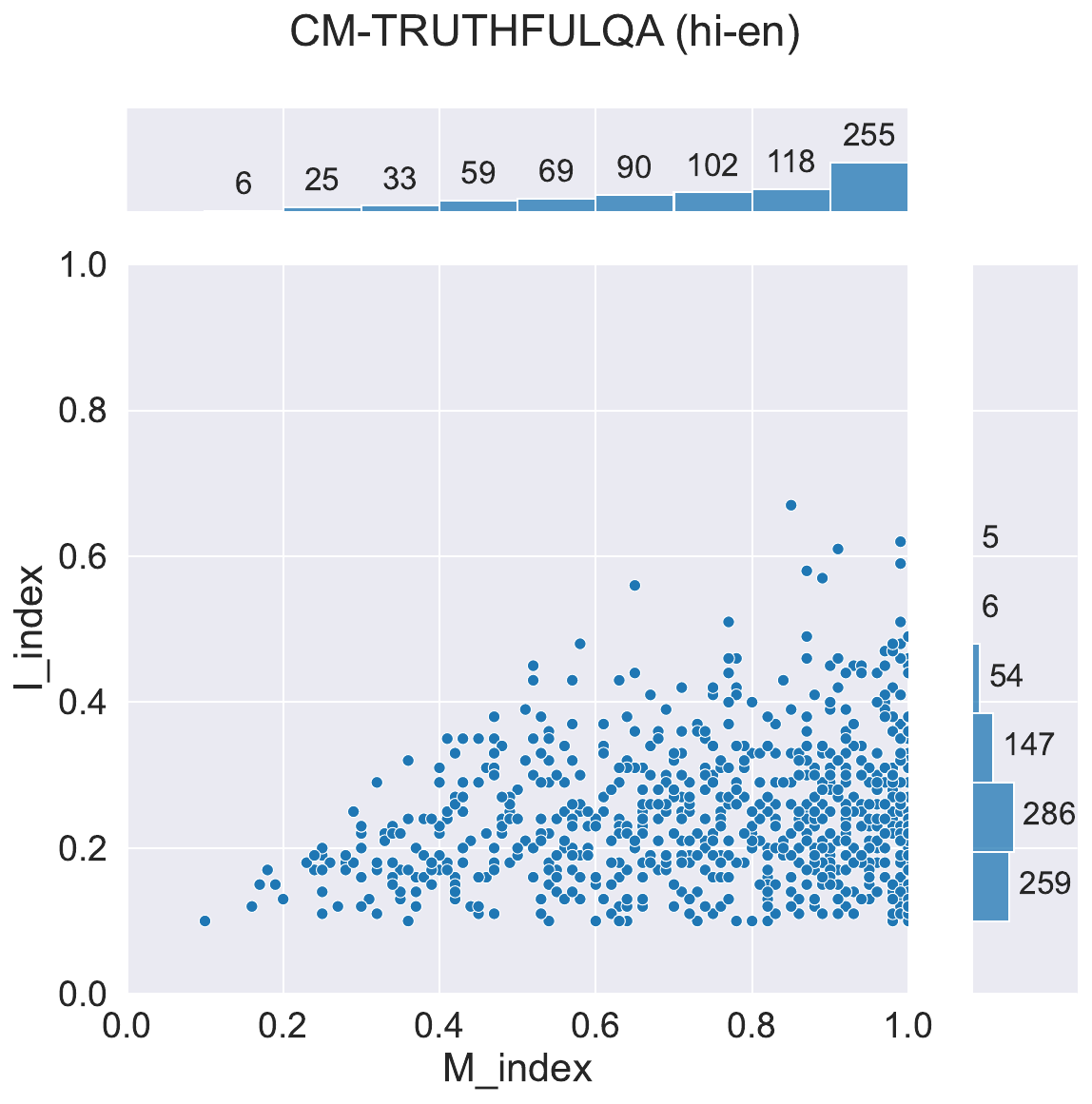}
      \label{fig:truthfulqa_hi}
    }
    \subfigure{
      \includegraphics[width=0.32\textwidth]{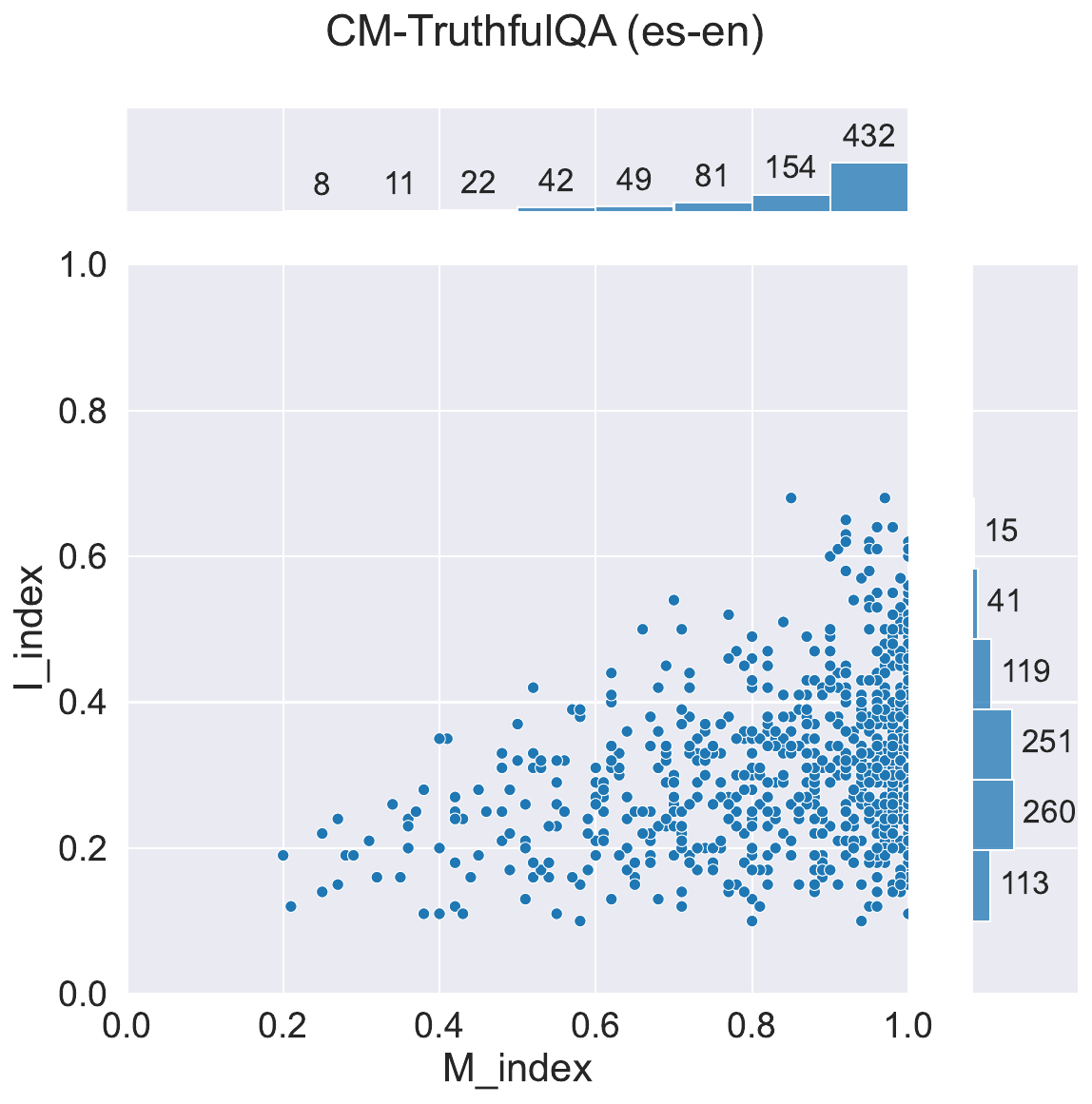}
      \label{fig:truthfulqa_es}
    }
    \subfigure{
      \includegraphics[width=0.32\textwidth]{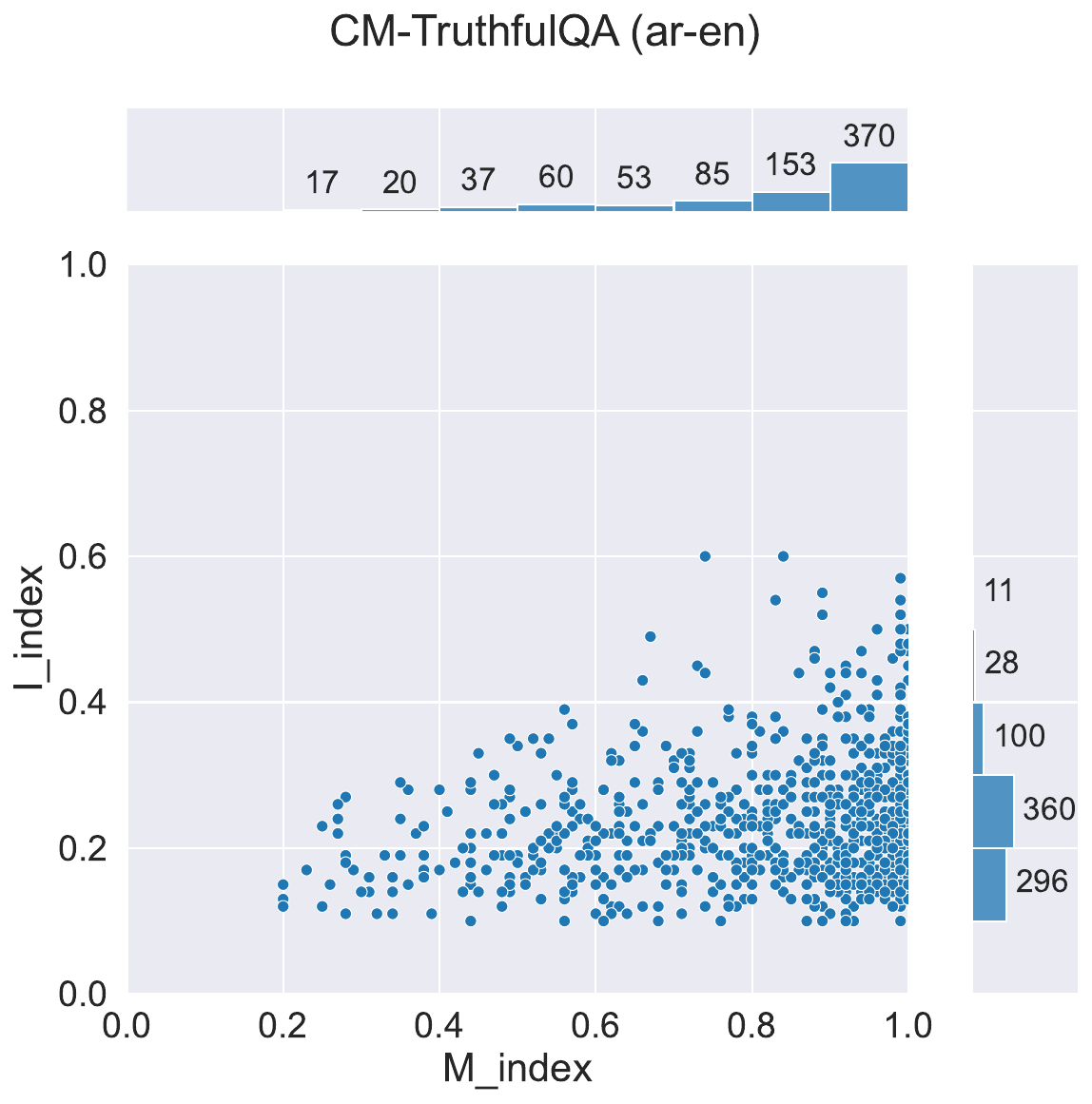}
      \label{fig:truthfulqa_ar}
    }
    \subfigure{
      \includegraphics[width=0.32\textwidth]{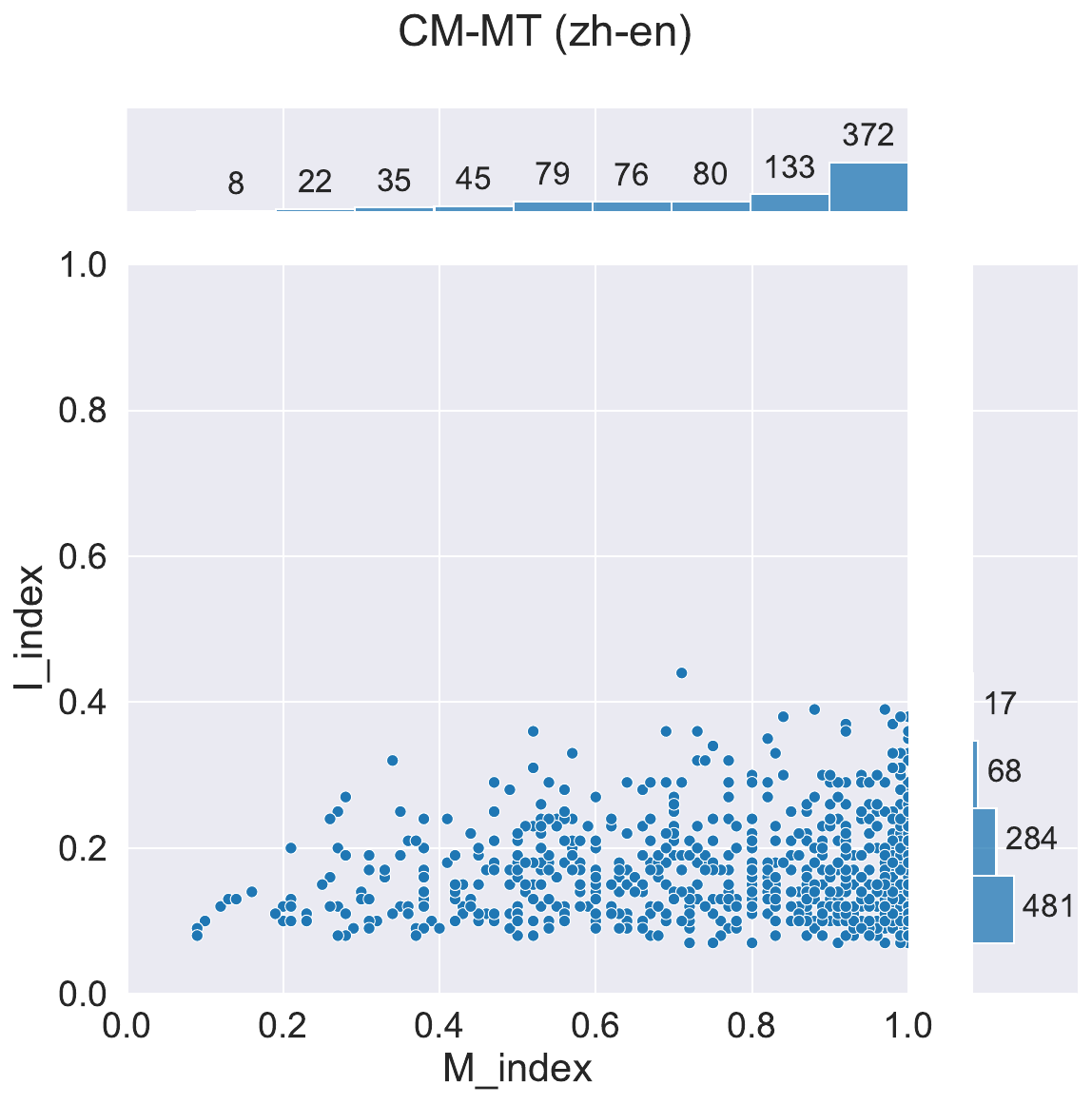}
      \label{fig:mt_zh}
    }
    \subfigure{
      \includegraphics[width=0.32\textwidth]{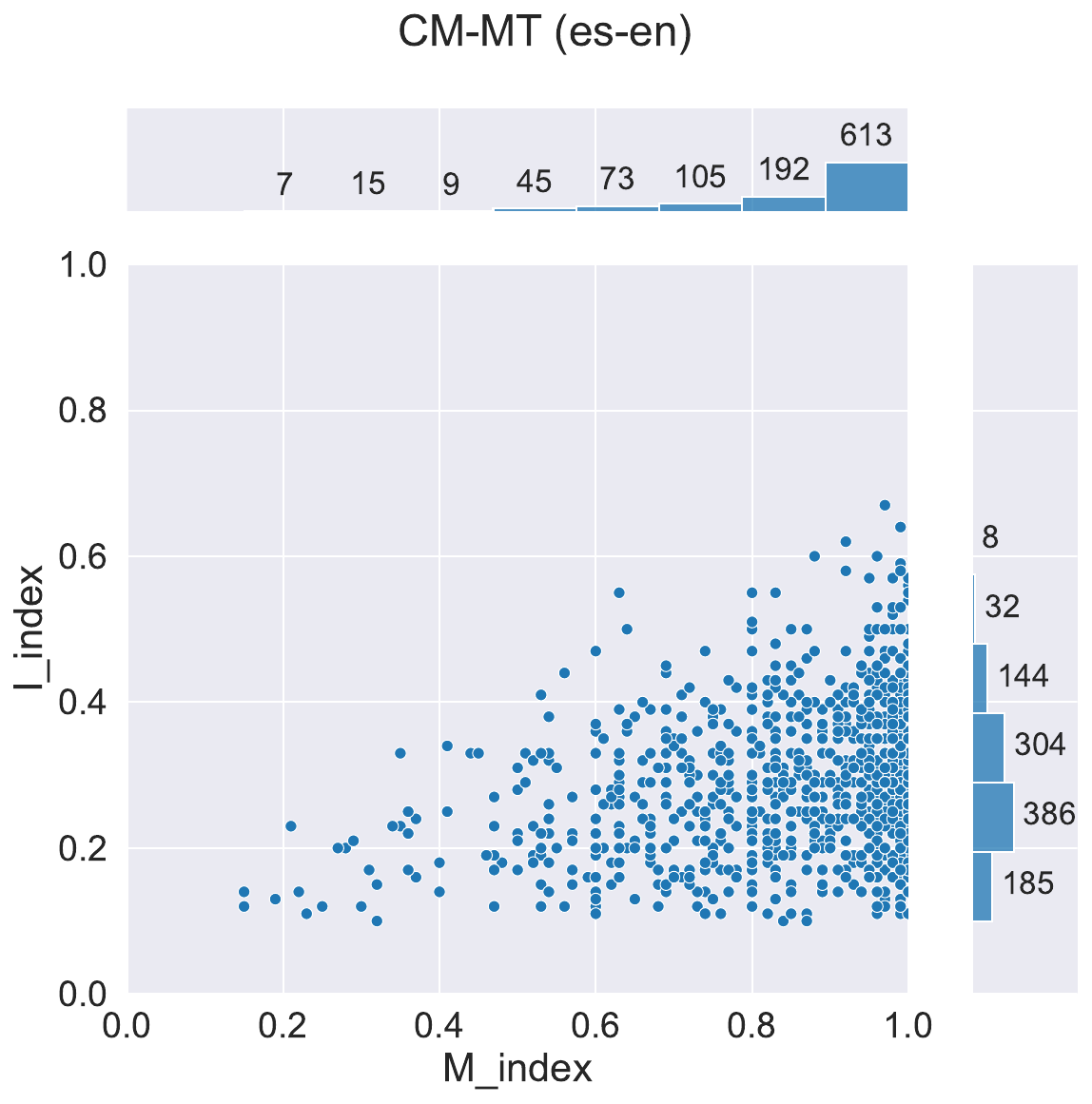}
      \label{fig:mt_es}
    }
    \subfigure{
      \includegraphics[width=0.32\textwidth]{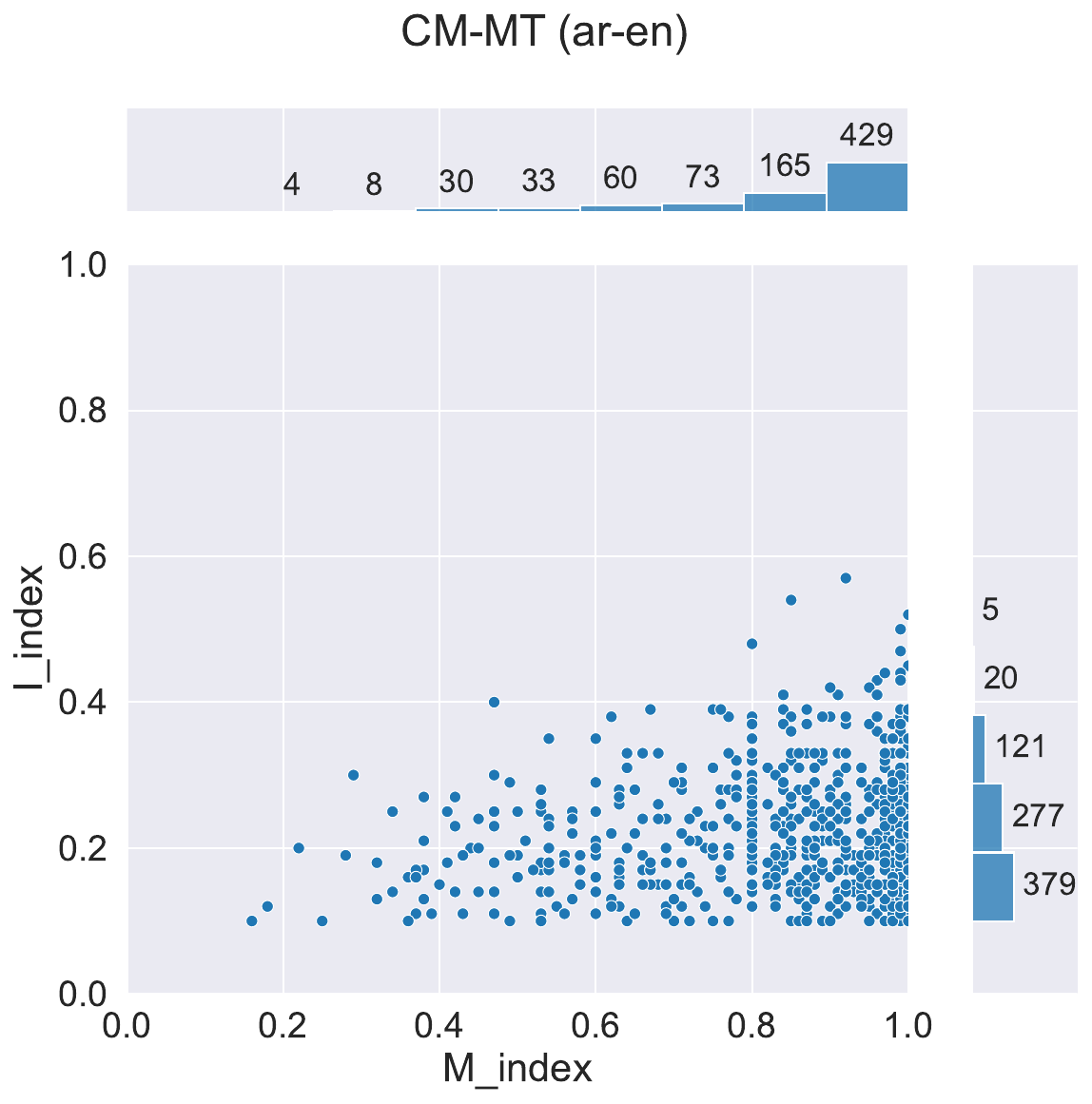}
      \label{fig:mt_ar}
    }
    \caption {\textbf{Distribution of Additional synthetic code-mixing datasets in {\Name}.}}
    \label{fig:additionalDistribution}
\end{figure*}

\onecolumn

\clearpage
\section{Experiment Results of Collected Datasets}
\label{app:one-shot}
\input{tables/one-shot_2}

\clearpage
\section{Results of K-shot Experiments across Language Pairs}
\label{app:k-shot}
\begin{figure*}[h]
    \subfigure{
      \includegraphics[width=\textwidth]{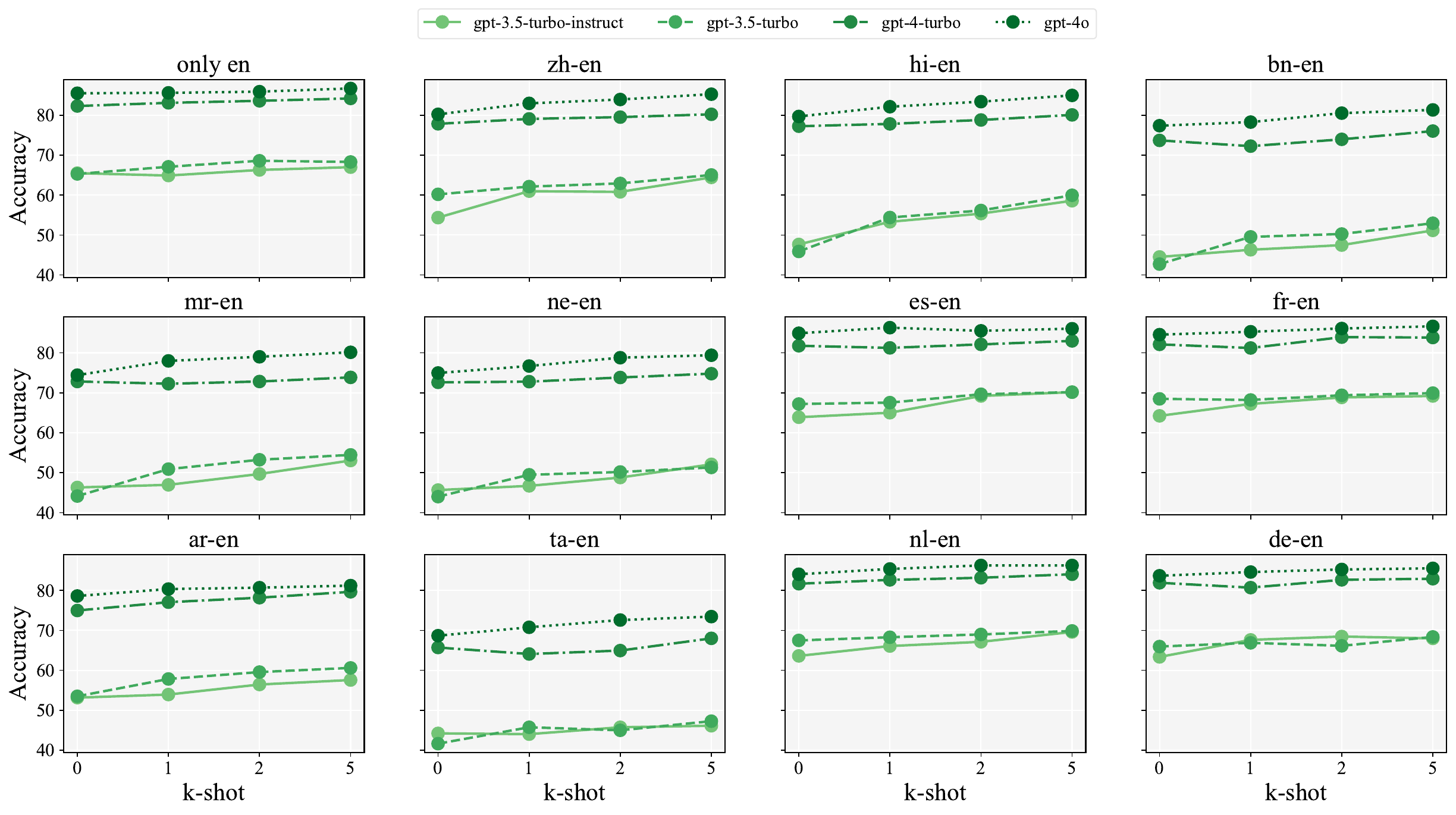}
      \label{fig:mmlu_gpt}
    }
    \subfigure{
      \includegraphics[width=\textwidth]{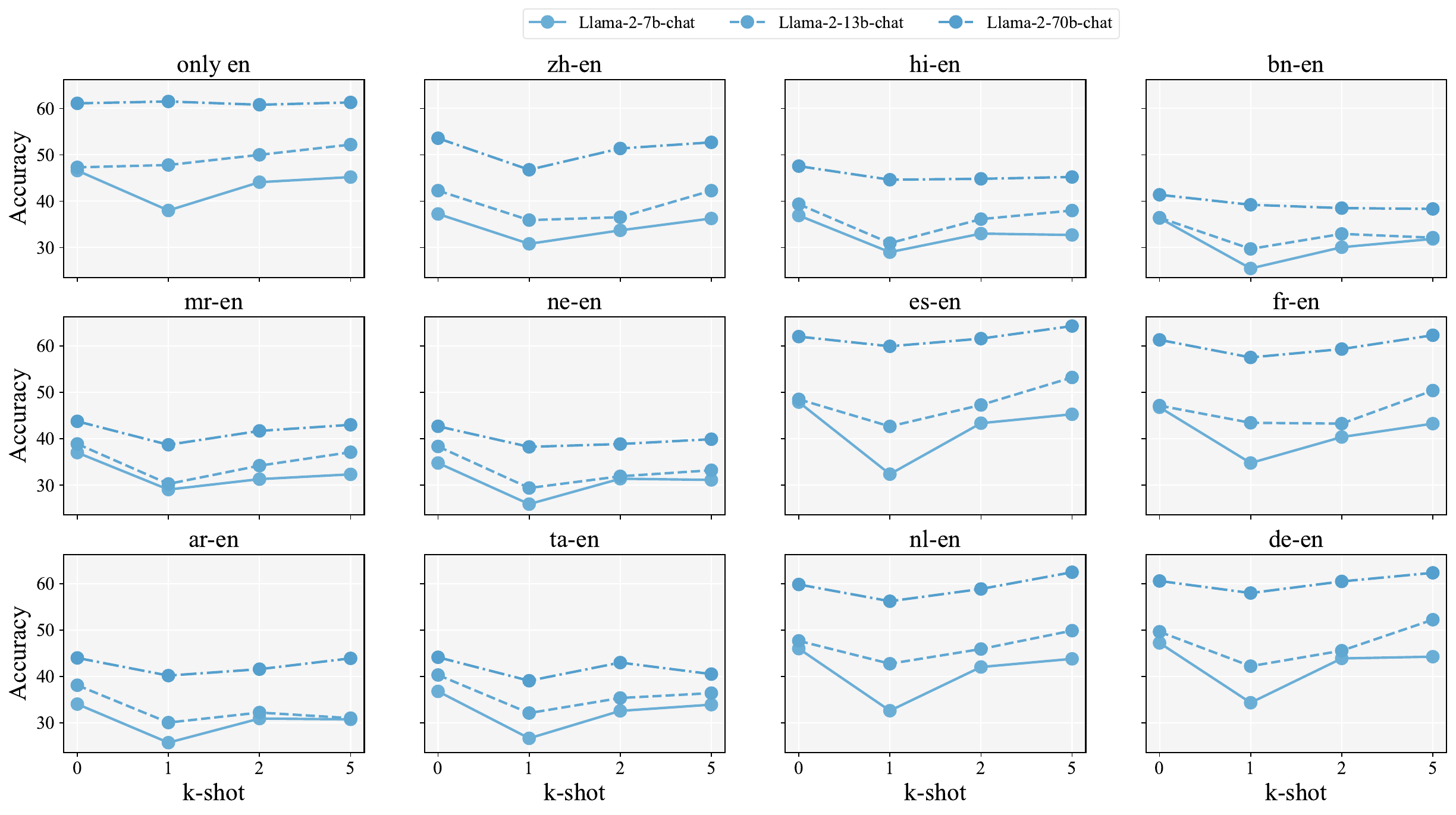}
      \label{fig:mmlu_llama}
    }
    \caption {\textbf{Accuracy of $K$-shot evaluation across three model families on CM-MMLU.}}
    \label{fig:mmlu_fewshot}
\end{figure*}

\label{app:k-shot}
\begin{figure*}[h]
    \ContinuedFloat
    \subfigure{
      \includegraphics[width=\textwidth]{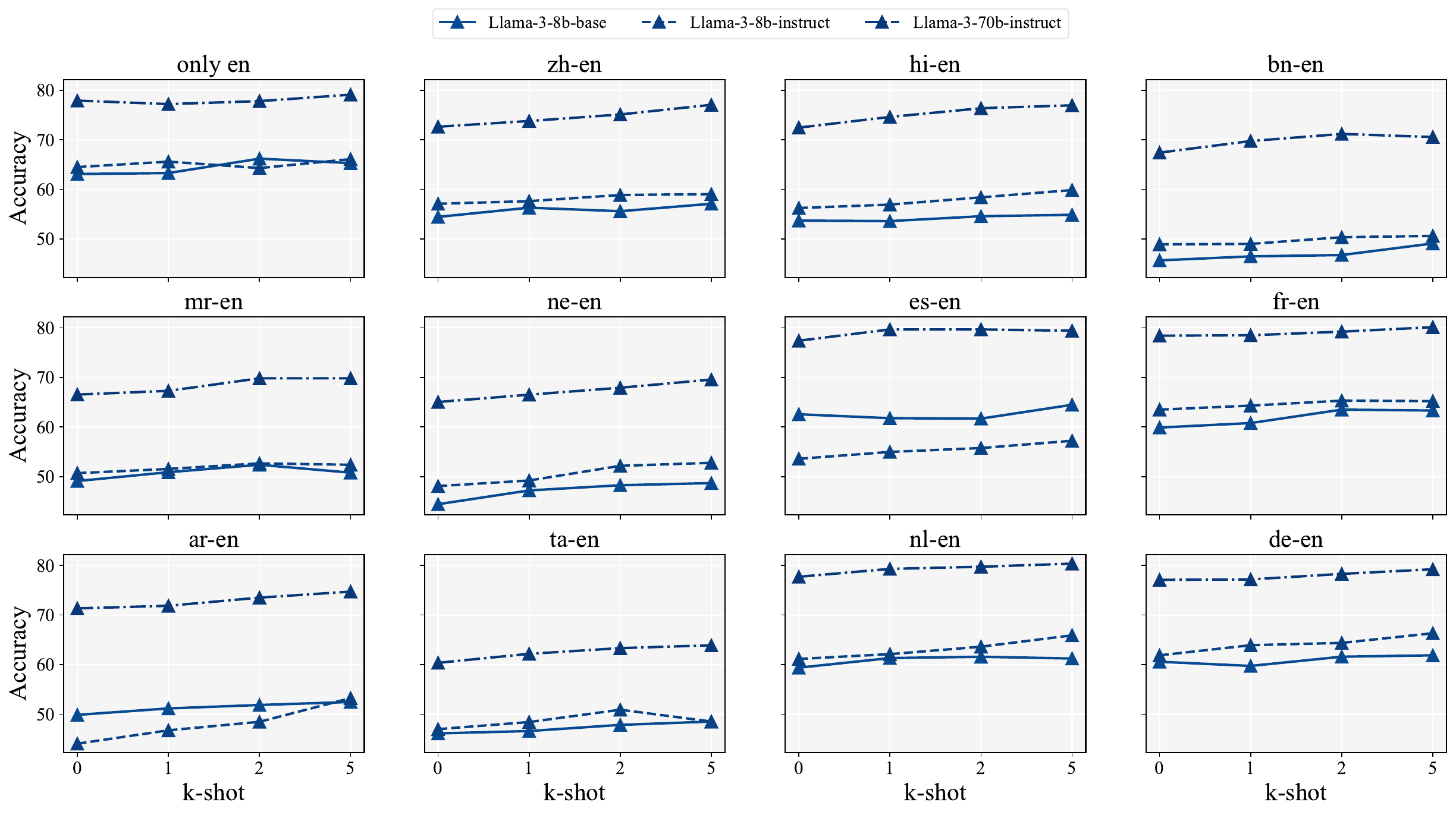}
      \label{fig:mmlu_llama}
    }
    \subfigure{
      \includegraphics[width=\textwidth]{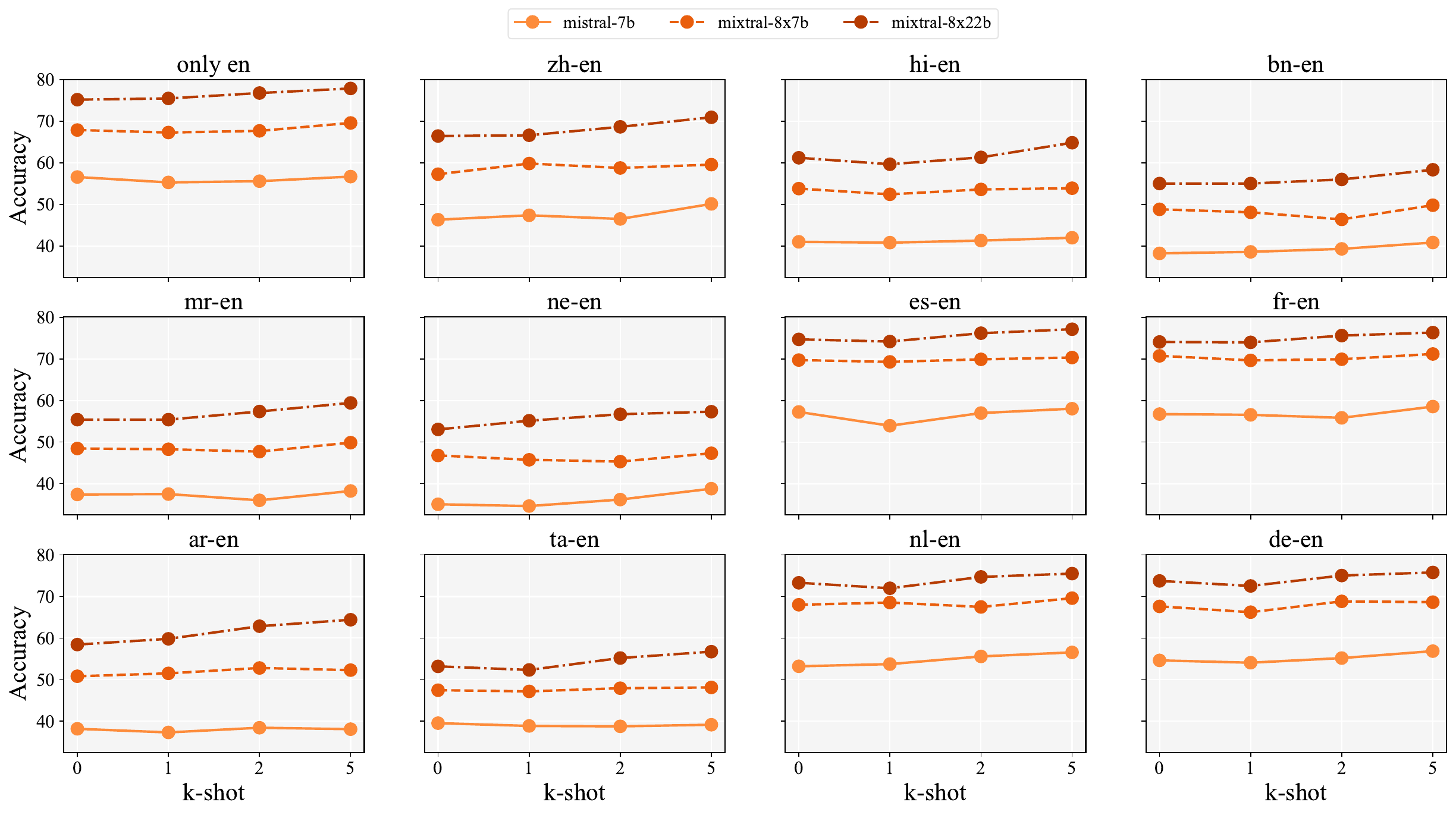}
      \label{fig:mmlu_mistral}
    }
    \caption {\textbf{Accuracy of $K$-shot evaluation across three model families on CM-MMLU.}}
    \label{fig:mmlu_fewshot}
\end{figure*}

\begin{figure*}[h]
    \subfigure{
      \includegraphics[width=\textwidth]{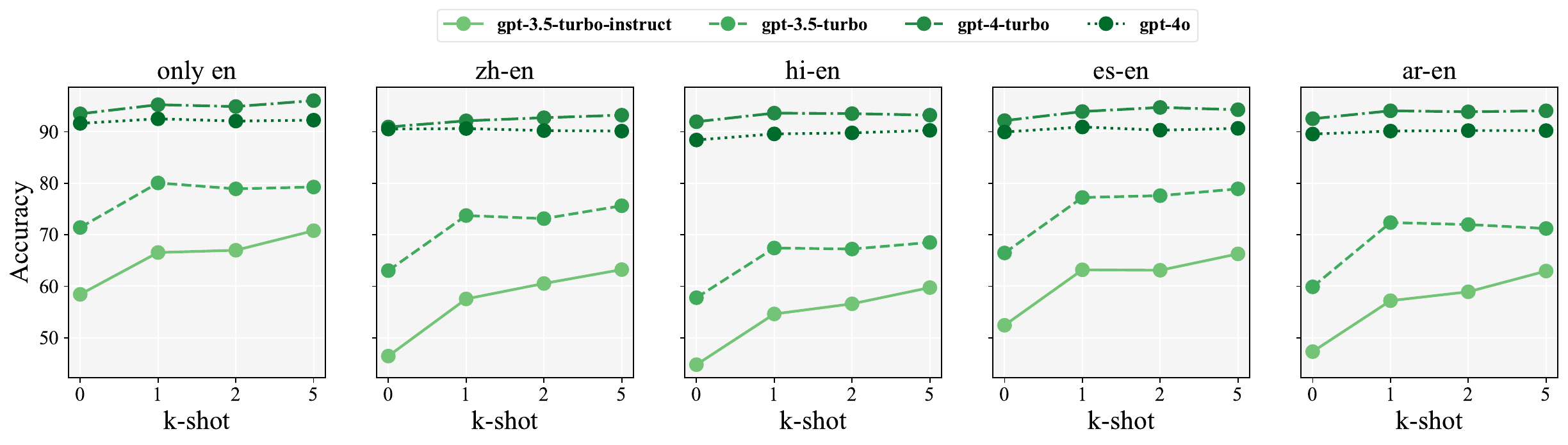}
      \label{fig:gsm8k_gpt}
    }
    \subfigure{
      \includegraphics[width=\textwidth]{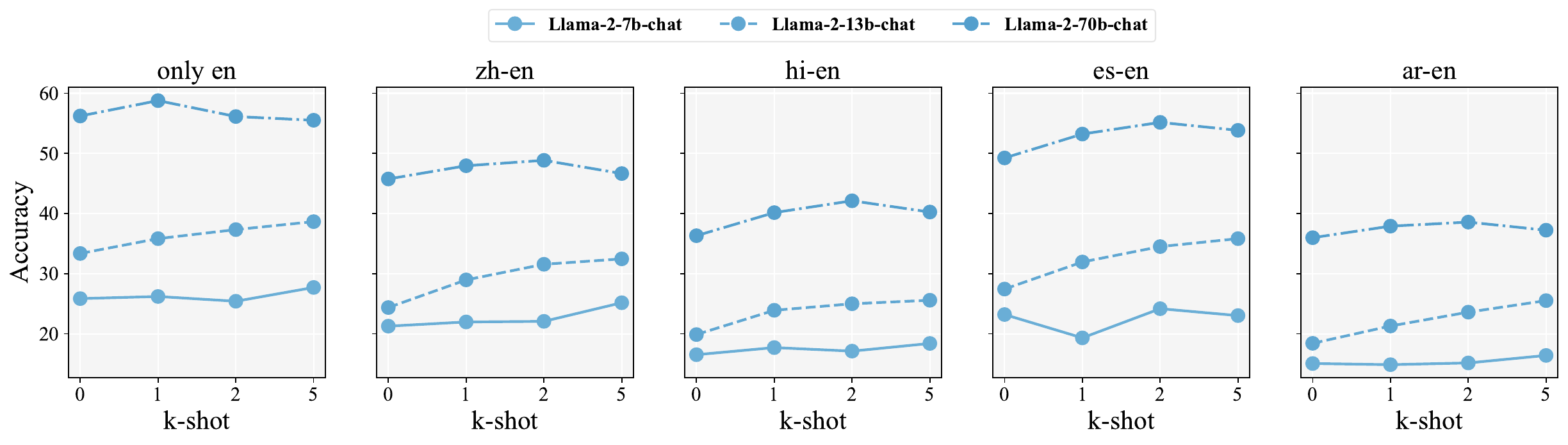}
      \label{fig:gsm8k_llama}
    }
    \subfigure{
      \includegraphics[width=\textwidth]{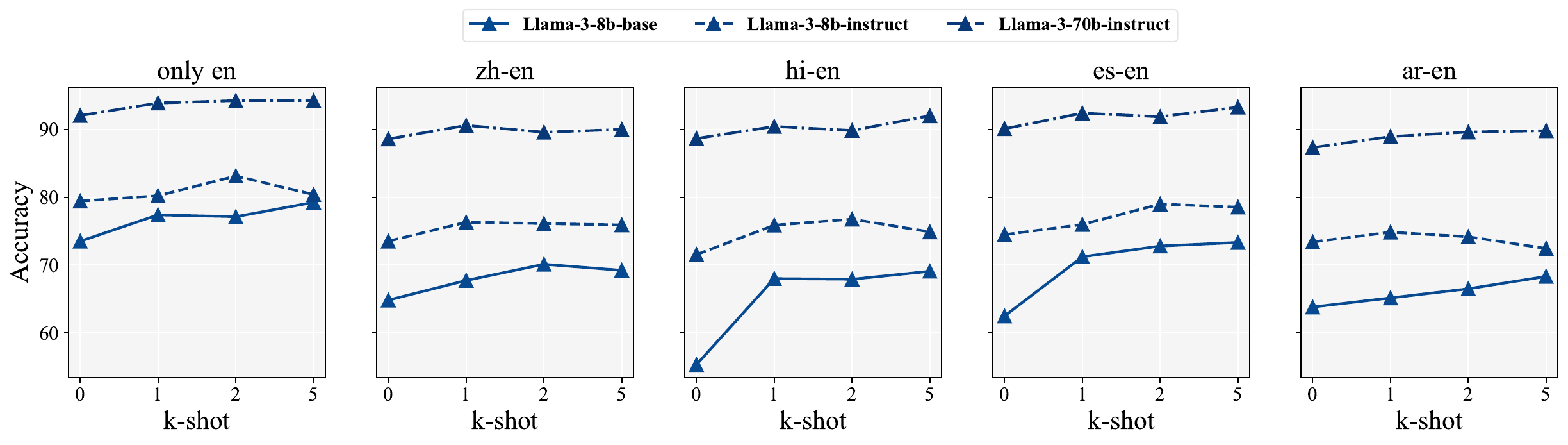}
      \label{fig:gsm8k_llama}
    }
    \subfigure{
      \includegraphics[width=\textwidth]{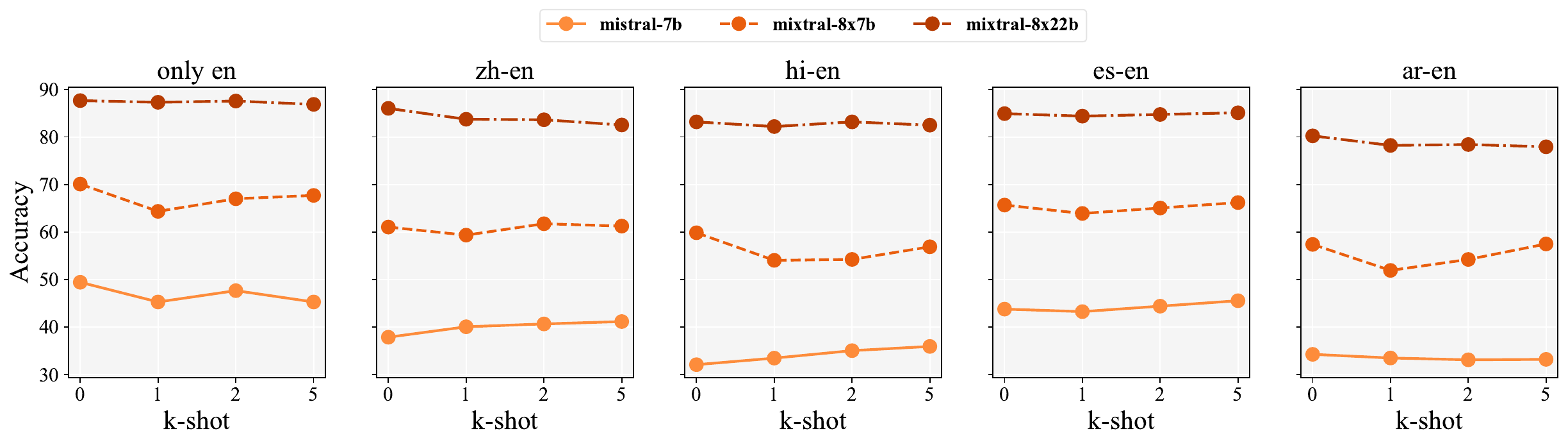}
      \label{fig:gsm8k_mistral}
    }
    \caption {\textbf{Accuracy of $K$-shot evaluation across three model families on CM-GSM8K.}}
    \label{fig:gsm8k_fewshot}
\end{figure*}

\begin{figure*}[h]
    \subfigure{
      \includegraphics[width=\textwidth]{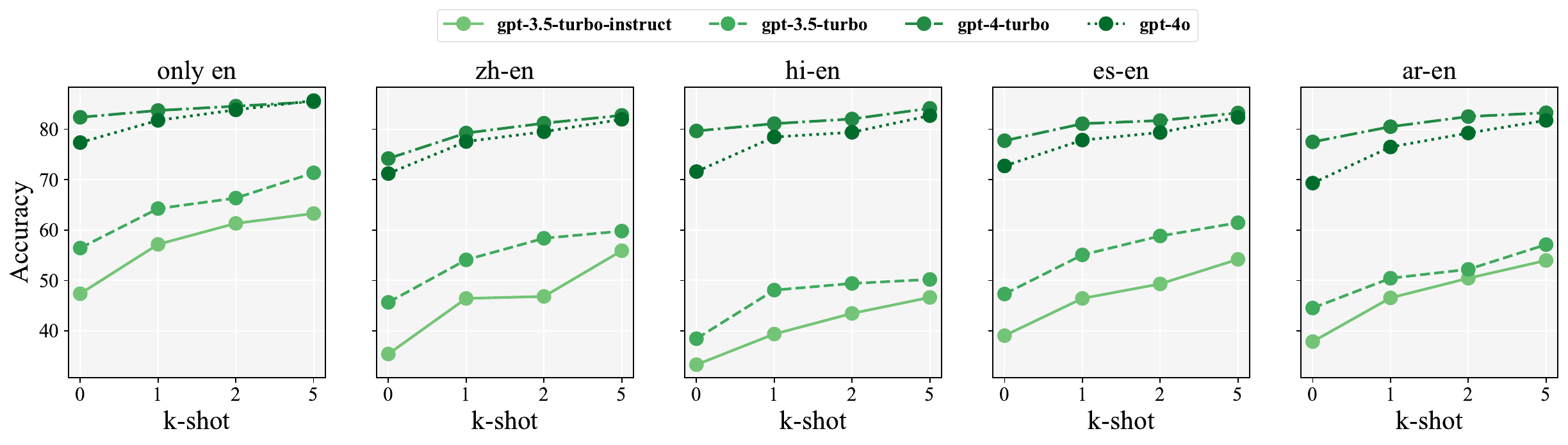}
      \label{fig:truthfulqa_gpt}
    }
    \subfigure{
      \includegraphics[width=\textwidth]{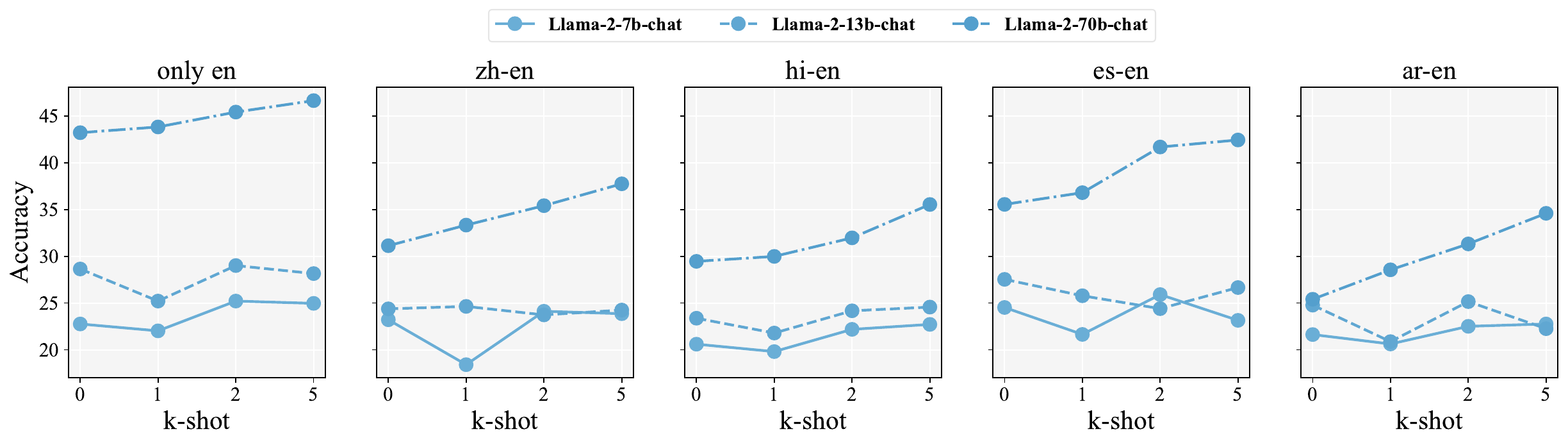}
      \label{fig:truthfulqa_llama}
    }
    \subfigure{
      \includegraphics[width=\textwidth]{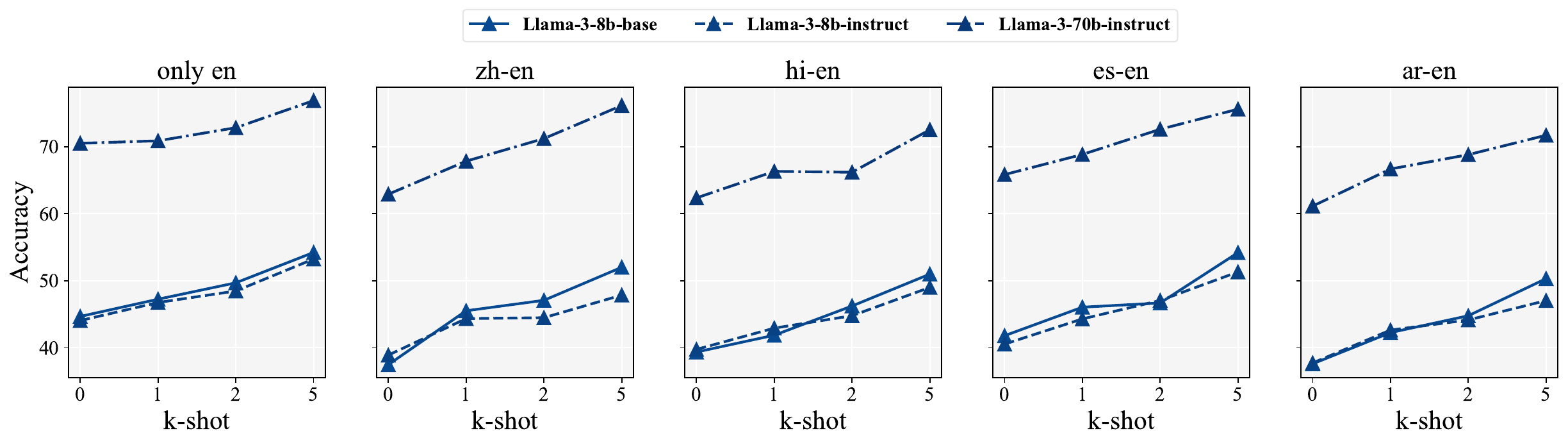}
      \label{fig:truthfulqa_llama}
    }
    \subfigure{
      \includegraphics[width=\textwidth]{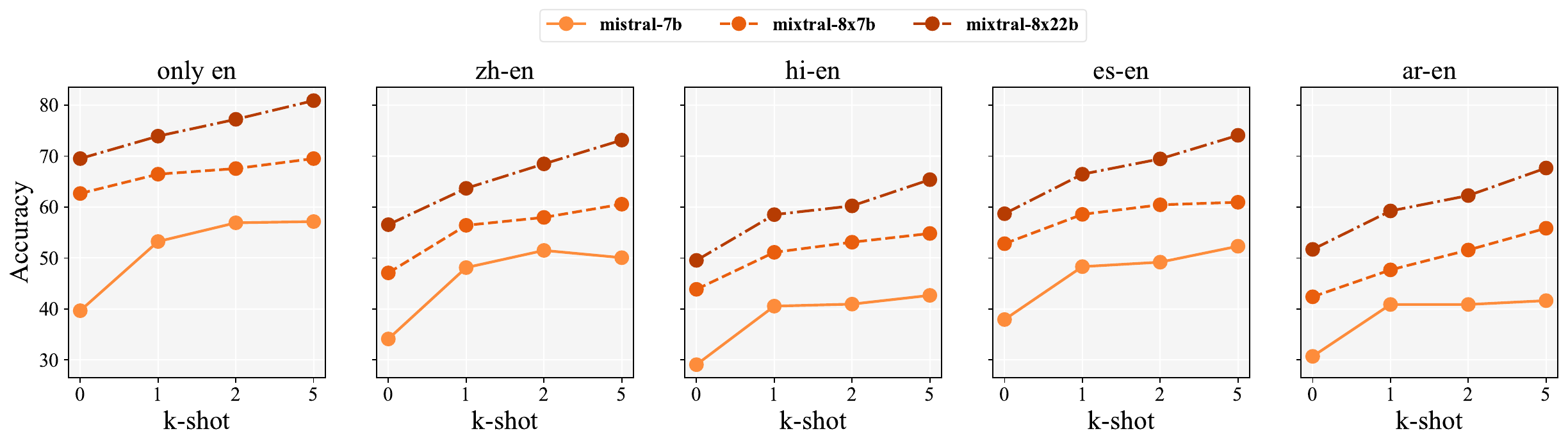}
      \label{fig:truthfulqa_gpt}
    }
    \caption {\textbf{Accuracy of $K$-shot evaluation across three model families on CM-TruthfulQA.}}
    \label{fig:truthfulqa_fewshot}
\end{figure*}

%% file: tables/versus.tex

\begin{table*}[t]
\centering
\vspace{-2em}
\resizebox{0.95\textwidth}{!}{%
\begin{tabular}{@{}lcccccccccc@{}}\toprule
\textbf{Language Pairs} & \textbf{LID} & \textbf{POS} & \textbf{NER} & \textbf{SA} & \textbf{MT} & \textbf{QA} & \textbf{NLI} & \textbf{Multi-Choice} & \textbf{Math} & \textbf{Truthfulness} \\ \midrule

\multicolumn{11}{c}{\textit{LinCE}} \\ 
Spanish-English & \checkmark & \checkmark & \checkmark & \checkmark & \checkmark & - & - & - & - & - \\
Hindi-English & \checkmark & \checkmark & \checkmark & - & \checkmark & - & - & - & - & - \\
Nepali-English & \checkmark & - & - & - & - & - & - & -  & - & -\\
MS Arabic-Egyptian Arabic & \checkmark & - & \checkmark & - & \checkmark & - & - & -  & - & - \\ \midrule
\multicolumn{11}{c}{\textit{GLUECoS}} \\
Spanish-English & \checkmark  & \checkmark & \checkmark & \checkmark & - & - & - & -  & - & - \\
Hindi-English & \checkmark  & \checkmark & \checkmark & \checkmark & - & \checkmark & \checkmark & -  & - & - \\ \midrule
\multicolumn{11}{c}{\textit{CodeMixBench}} \\
Spanish-English & \checkmark  & \checkmark & \checkmark & \checkmark & \checkmark & - & - & \checkmark  & \checkmark & \checkmark \\
Hindi-English & \checkmark  & \checkmark & \checkmark & \checkmark & \checkmark & - & - & \checkmark  & \checkmark & \checkmark \\
Nepali-English & \checkmark  & - & - & \checkmark & - & - & - & \checkmark  & - & - \\
MS Arabic-Egyptian Arabic & \checkmark  & - & \checkmark & - & - & - & - & -  & - & - \\
Arabic-English & -  & - & - & - & \checkmark & - & - & \checkmark  & \checkmark & \checkmark \\ 
Chinese-English & \checkmark & \checkmark & - & - & \checkmark & - & - & \checkmark  & \checkmark & \checkmark \\
Bengali-English & - & - & - & \checkmark & \checkmark & - & - & \checkmark  & - & - \\
Marathi-English & \checkmark & - & - & - & \checkmark & - & - & \checkmark  & - & - \\
Tamil-English & - & - & - & \checkmark & - & - & - & \checkmark  & - & - \\
Malayalam-English & - & - & - & \checkmark & - & - & - & -  & - & - \\
French-English & - & - & - & - & - & - & - & \checkmark  & - & - \\
Dutch-English & - & - & - & - & - & - & - & \checkmark  & - & - \\
German-English & \checkmark & - & - & - & - & - & - & \checkmark  & - & - \\
Frisian-Dutch & \checkmark & \checkmark & - & - & - & - & - & - & - & - \\
Hokkien-Chinese & \checkmark & - & - & - & \checkmark & - & - & - & - & - \\
Guarani-Spanish & \checkmark & - & \checkmark & - & - & - & - & -  & - & - \\

\bottomrule
\end{tabular}%
}
\caption{
\textbf{Overview of the CodeMixbench language pairs and tasks compared to LinCE and GLUECoS.}
}
\label{tab:versus}
\vspace{-1em}
\end{table*}

%% file: tables/collected_datasets.tex
\begin{table*}[t]
\centering
\begin{tabular}{ccccccc} \toprule

 & \textbf{Languages} & \textbf{Size} & \textbf{All Tokens} & \textbf{M-index} & \textbf{I-index} & \textbf{Source}  \\ \midrule
\multirow{10}{*}{LID}
    & \multicolumn{1}{c}{zh-en} & \multicolumn{1}{c}{3,022} & \multicolumn{1}{c}{37,064} & 0.538 & 0.399 & \citet{calvilloSurprisalPredictsCodeSwitching2020b} \\ 
    & \multicolumn{1}{c}{hok-zh} & \multicolumn{1}{c}{3,800} & \multicolumn{1}{c}{44,022} & 0.557 & 0.173 & \citet{luExploringMethodsBuilding2022a} \\ 
    & \multicolumn{1}{c}{hi-en} & \multicolumn{1}{c}{744} & \multicolumn{1}{c}{15,446} & 0.224 & 0.137 & \citet{maveLanguageIdentificationAnalysis2018b} \\ 
    & \multicolumn{1}{c}{ne-en} & \multicolumn{1}{c}{1,332} & \multicolumn{1}{c}{19,273} & 0.388 & 0.220 & \citet{solorioOverviewFirstShared2014} \\ 
    & \multicolumn{1}{c}{mr-en} & \multicolumn{1}{c}{1,340} & \multicolumn{1}{c}{15,945} & 0.347 & 0.241 & \citet{chavanMyBoliCodemixed2023} \\ 
    & \multicolumn{1}{c}{es-en} & \multicolumn{1}{c}{1,133} & \multicolumn{1}{c}{40,391} & 0.160 & 0.077 & \citet{molinaOverviewSecondShared2016} \\ 
    & \multicolumn{1}{c}{msa-ea} & \multicolumn{1}{c}{1,116} & \multicolumn{1}{c}{21,978} & 0.073 & 0.031 & \citet{molinaOverviewSecondShared2016} \\ 
    & \multicolumn{1}{c}{de-en} & \multicolumn{1}{c}{1,252} & \multicolumn{1}{c}{37,511} & 0.232 & 0.077 & \citet{sternerTongueSwitcherFineGrainedIdentification2023} \\ 
    & \multicolumn{1}{c}{fy-nl} & \multicolumn{1}{c}{250} & \multicolumn{1}{c}{2,356} & 0.381 & 0.278 & \citet{braggaarChallengesAnnotatingParsing2021} \\ 
    & \multicolumn{1}{c}{gn-es} & \multicolumn{1}{c}{180} & \multicolumn{1}{c}{2,857} & 0.558 & 0.327 & \citet{chiruzzoOverviewGUASPAIberLEF2023a} \\ \midrule
\multirow{4}{*}{POS}
    & \multicolumn{1}{c}{zh-en} & \multicolumn{1}{c}{2,909} & \multicolumn{1}{c}{35,600} & - & - & \citet{calvilloSurprisalPredictsCodeSwitching2020b} \\
    & \multicolumn{1}{c}{hi-en} & \multicolumn{1}{c}{160} & \multicolumn{1}{c}{3,476} & - & - & \citet{singhTwitterCorpusHindiEnglish2018} \\ 
    & \multicolumn{1}{c}{es-en} & \multicolumn{1}{c}{1,000} & \multicolumn{1}{c}{7,712} & - & - & \citet{sotoCrowdsourcingUniversalPartofSpeech2017a} \\  
    & \multicolumn{1}{c}{fy-nl} & \multicolumn{1}{c}{250} & \multicolumn{1}{c}{2,356} & 0.381 & 0.278 & \citet{braggaarChallengesAnnotatingParsing2021} \\ \midrule
\multirow{4}{*}{NER}
    & \multicolumn{1}{c}{hi-en} & \multicolumn{1}{c}{314} & \multicolumn{1}{c}{5,364} & - & - & \citet{singhLanguageIdentificationNamed2018a} \\
    & \multicolumn{1}{c}{es-en} & \multicolumn{1}{c}{1,000} & \multicolumn{1}{c}{12,139} & - & - & \citet{aguilarNamedEntityRecognition2018a} \\ 
    & \multicolumn{1}{c}{msa-ea} & \multicolumn{1}{c}{1,122} & \multicolumn{1}{c}{22,742} & - & - & \citet{aguilarNamedEntityRecognition2018a} \\ 
    & \multicolumn{1}{c}{gn-es} & \multicolumn{1}{c}{180} & \multicolumn{1}{c}{2,857} & 0.558 & 0.327 & \citet{chiruzzoOverviewGUASPAIberLEF2023a} \\ \midrule
 \multirow{7}{*}{SA}
    & \multicolumn{1}{c}{hi-en} & \multicolumn{1}{c}{1,261} & \multicolumn{1}{c}{-} & - & - & \citet{patraSentimentAnalysisCodeMixed2018} \\ 
    & \multicolumn{1}{c}{bn-en} & \multicolumn{1}{c}{1,000} & \multicolumn{1}{c}{-} & - & - & \citet{raihanOffensiveLanguageIdentification2023c} \\ 
    & \multicolumn{1}{c}{mr-en} & \multicolumn{1}{c}{1,250} & \multicolumn{1}{c}{-} & - & - & \citet{chavanMyBoliCodemixed2023} \\ 
    & \multicolumn{1}{c}{ne-en} & \multicolumn{1}{c}{1,070} & \multicolumn{1}{c}{-} & - & - & \citet{pahariLanguagePreferenceExpression2023} \\ 
    & \multicolumn{1}{c}{es-en} & \multicolumn{1}{c}{1,859} & \multicolumn{1}{c}{28,202} & - & - & \citet{patwaSemEval2020TaskOverview2020a} \\ 
    & \multicolumn{1}{c}{ta-en} & \multicolumn{1}{c}{3,049} & \multicolumn{1}{c}{-} & - & - & \citet{chakravarthi-etal-2020-corpus} \\ 
    & \multicolumn{1}{c}{ml-en} & \multicolumn{1}{c}{1,171} & \multicolumn{1}{c}{-} & - & - & \citet{chakravarthiSentimentAnalysisDataset2020} \\ \midrule
    \multirow{5}{*}{MT}
    & \multicolumn{1}{c}{zh-en->zh} & \multicolumn{1}{c}{3,022} & \multicolumn{1}{c}{37,064} & 0.538 & 0.399 & \citet{calvilloSurprisalPredictsCodeSwitching2020b} \\ 
    & \multicolumn{1}{c}{hok-zh->zh} & \multicolumn{1}{c}{3,800} & \multicolumn{1}{c}{44,022} & 0.557 & 0.173 & \citet{luExploringMethodsBuilding2022a} \\ 
    & \multicolumn{1}{c}{hi-en->en} & \multicolumn{1}{c}{942} & \multicolumn{1}{c}{11,849} & 0.90 & 0.53 & \citet{chenCALCS2021Shared2022} \\ 
    & \multicolumn{1}{c}{bn-en->en} & \multicolumn{1}{c}{2,000} & \multicolumn{1}{c}{-} & - & - & \citet{vavreAdaptingMultilingualModels2022} \\ 
    & \multicolumn{1}{c}{mr-en->en} & \multicolumn{1}{c}{2,000} & \multicolumn{1}{c}{-} & - & - & \citet{vavreAdaptingMultilingualModels2022} \\ \bottomrule
\end{tabular}
  \caption{
    \textbf{The statistics of collected datasets.} 
  }
  \label{tab:collected}
\end{table*}

%% file: tables/labse.tex
\begin{table}[]
\centering
\resizebox{0.9\columnwidth}{!}{%
\begin{tabular}{lccc}\toprule
\textbf{Language Pair} & \multicolumn{1}{c}{\textbf{CM \& L1}} & \multicolumn{1}{c}{\textbf{CM \& L2}} & \multicolumn{1}{c}{\textbf{L1 \& L2}} \\ \midrule
zh-en & 0.958 & 0.936 & 0.914 \\
hi-en & 0.951 & 0.918 & 0.887 \\
bn-en & 0.938 & 0.907 & 0.883 \\
mr-en & 0.910 & 0.894 & 0.854 \\
ne-en & 0.929 & 0.905 & 0.879 \\
es-en & 0.972 & 0.967 & 0.939 \\
fr-en & 0.974 & 0.962 & 0.935 \\
ar-en & 0.956 & 0.941 & 0.911 \\
ta-en & 0.892 & 0.873 & 0.838 \\
nl-en & 0.967 & 0.952 & 0.923 \\
de-en & 0.968 & 0.945 & 0.914 \\ \bottomrule
\end{tabular}%
}
\caption{
\textbf{LaBSE scores for synthetic code-mixed data across different language pairs.} \( L1 \) indicates non-English languages. \( L2 \) denotes English. \( CM \) denotes synthesized code-mixing data.
}
\label{tab:labsescores}
\end{table}

%% file: tables/gpt4valid.tex

\begin{sidewaystable*}[]
\centering
\resizebox{\textwidth}{!}{%
\begin{tabular}{cp{6cm}p{18cm}cc} \toprule
\textbf{Index} & \multicolumn{1}{c}{\textbf{Code-Mix}} & \multicolumn{1}{c}{\textbf{Chain of Thought Analysis}} & \textbf{GPT-4} & \textbf{Human} \\
1 & \_\_\_\_\_\_\_\_\_\_\_\_ 在向目标受众传递信息方面非常 effective, 因为它可以以大多数其他 media 无法做到的方式进行 explanation。(A): Digital 媒体。(B): 印刷媒体。(C): Broadcast 媒体。(D): 户外 media。 & The multi-choice question integrates English and Chinese in a manner that maintains logical flow and coherence, especially for readers familiar with both languages. The use of English words such as "effective," "media," and "explanation" within a predominantly Chinese sentence structure does not disrupt the overall understanding. This indicates a good level of coherence. The naturalness of the code-mixing might vary depending on the reader's fluency in both languages, but given the context of media and communication, the use of English terms feels appropriate and mirrors real-world usage in bilingual or multilingual environments. Therefore, the naturalness is fairly high. The readability is also good, as the structure of the question follows a logical pattern, and the code-mixing does not obscure the meaning. The bilingual reader should find the question accessible, with the English terms likely being familiar concepts within the context of media studies. & 3, 2, 3 & 3, 2, 3 \\ \midrule
2 & 在膳食脂肪酸类中，which have the most beneficial effect on plasma LDL cholesterol levels?(A): n-3和n-6 polyunsaturated fatty acids (B): 饱和脂肪酸和trans unsaturated fatty acids (C): Monounsaturated fatty acids (D): Both a和c & The multi-choice question integrates English and Chinese in a manner that maintains logical flow and coherence, especially for readers familiar with both languages. The terminology related to dietary fats and cholesterol levels is presented in a way that leverages English scientific terms within a Chinese sentence structure, which is a common practice in scientific communities with bilingual speakers. This integration feels natural to those accustomed to discussing specialized topics in a code-mixed language environment. The readability is somewhat impacted by the switch between languages, especially for readers who may not be equally proficient in both, but the overall structure and context help in understanding the question's intent. & 3, 3, 2 & 3, 3, 3 \\ \midrule
3 & 定价方法根据顾客所感知的价值来设定价格的是: (A): Cost-oriented approach。 (B): 需求导向的方法。 (C): Competitor-oriented approach。 (D): 价值导向的方法。 & The multi-choice question integrates English and Chinese languages, focusing on pricing strategies. The terms "Cost-oriented approach," "Competitor-oriented approach," and the Chinese equivalents for "demand-oriented method" and "value-oriented method" are mixed, targeting an audience familiar with business terminology in both languages. The question maintains coherence by clearly presenting the options related to pricing methods, despite the language switch. The naturalness is somewhat affected by the direct mixing of English and Chinese terms, which might not be standard practice in a purely academic or professional setting but could be natural in a bilingual educational context. The readability is fairly high, as the structure of the question is straightforward, and the bilingual reader can easily understand the options provided, even though they switch between languages. & 2, 2, 2 & 3, 2 ,3 \\ \midrule
4 & 语句1 | 传统机器学习结果假定训练集和测试集是独立且同分布的。语句2 | In 2017, COCO models were usually pretrained on ImageNet. (A): 正确，True (B): 错误，False (C): 正确，False (D): 错误，True & The multi-choice question integrates two languages (Chinese and English) to discuss a concept related to machine learning and model pretraining. The transition between the two languages is smooth, given the context is academic and technical, where English terms are commonly used in non-English speaking countries due to the global nature of the field. The statements are connected logically, discussing assumptions in machine learning and practices in model pretraining, which are relevant to each other. The use of both languages seems intentional and reflects a natural way of discussing specialized topics in a bilingual setting. The readability is high, assuming the audience is familiar with both languages and the technical terms used. The options provided mix languages for "correct" and "incorrect", which might slightly challenge readability but remains within the context of code-mixing practices. & 3, 3, 2 & 3, 3, 2 \\ \bottomrule
\end{tabular}%
}
\caption{\textbf{Four sampling validation examples in Model-Aligned Evaluation.} The scores in both the GPT-4 and Human columns are arranged in the order of Coherence, Naturalness, and Readability.}
\label{tab:gpt4valid}
\end{sidewaystable*}

%% file: tables/statistics.tex
\begin{table*}[h]
\centering
\resizebox{\textwidth}{!}{%
\begin{tabular}{cccccccccccc} \toprule

\multirow{2}{*}{} & \multirow{2}{*}{\textbf{Lang.}} & \multirow{2}{*}{\textbf{Size}} & \multirow{2}{*}{\begin{tabular}[c]{@{}c@{}}\textbf{L1 / L2 / Other}\\\textbf{tokens}\end{tabular}} & \multicolumn{2}{c}{\textbf{Word-Level}} & \multicolumn{3}{c}{\textbf{Semantic}} & \multicolumn{3}{c}{\textbf{Model-Aligned}} \\ \cline{5-12} 
 &  &  &  & \textbf{M} & \textbf{I} & \textbf{sim1} & \textbf{sim2} & \textbf{sim3} & \textbf{Co.} & \textbf{Na.} & \textbf{Re.} \\ \midrule
\multirow{11}{*}{\rotatebox{90}{MMLU}}
 & \multicolumn{1}{c}{zh-en} & \multicolumn{1}{c}{1133} & \multicolumn{1}{c}{32510 / 17765 / 2646} & 0.75 & \multicolumn{1}{c}{0.22} & 0.96 & 0.94 & \multicolumn{1}{c}{0.92} & 2.89 & 2.52 & 2.48 \\
 & \multicolumn{1}{c}{es-en} & \multicolumn{1}{c}{1146} & \multicolumn{1}{c}{26303 / 30492 / 3652} & 0.87 & \multicolumn{1}{c}{0.31} & 0.97 & 0.97 & \multicolumn{1}{c}{0.94} & 2.89 & 2.62 & 2.56 \\
 & \multicolumn{1}{c}{fr-en} & \multicolumn{1}{c}{1107} & \multicolumn{1}{c}{29412 / 27589 / 3549} & 0.86 & \multicolumn{1}{c}{0.27} & 0.97 & 0.96 & \multicolumn{1}{c}{0.94} & 2.80 & 2.49 & 2.42 \\ 
 & \multicolumn{1}{c}{de-en} & \multicolumn{1}{c}{1078} & \multicolumn{1}{c}{27856 / 22163 / 3701} & 0.85 & \multicolumn{1}{c}{0.28} & 0.97 & 0.95 & \multicolumn{1}{c}{0.91} & 2.79 & 2.44 & 2.39 \\
 & \multicolumn{1}{c}{nl-en} & \multicolumn{1}{c}{1135} & \multicolumn{1}{c}{28992 / 26243 / 3551} & 0.87 & \multicolumn{1}{c}{0.31} & 0.97 & 0.95 & \multicolumn{1}{c}{0.92} & 2.85 & 2.52 & 2.48 \\
 & \multicolumn{1}{c}{ar-en} & \multicolumn{1}{c}{1155} & \multicolumn{1}{c}{26977 / 18815 / 3346} & 0.78 & \multicolumn{1}{c}{0.22} & 0.96 & 0.94 & \multicolumn{1}{c}{0.92} & 2.85 & 2.54 & 2.45 \\
 & \multicolumn{1}{c}{hi-en} & \multicolumn{1}{c}{1024} & \multicolumn{1}{c}{30767 / 19174 / 3417} & 0.77 & \multicolumn{1}{c}{0.25} & 0.95 & 0.92 & \multicolumn{1}{c}{0.89} & 2.93 & 2.74 & 2.55 \\
 & \multicolumn{1}{c}{bn-en} & \multicolumn{1}{c}{1114} & \multicolumn{1}{c}{23912 / 22680 / 3667} & 0.82 & \multicolumn{1}{c}{0.25} & 0.93 & 0.91 & \multicolumn{1}{c}{0.87} & 2.86 & 2.63 & 2.50 \\
 & \multicolumn{1}{c}{mr-en} & \multicolumn{1}{c}{1067} & \multicolumn{1}{c}{21402 / 21956 / 4380} & 0.84 & \multicolumn{1}{c}{0.24} & 0.93 & 0.91 & \multicolumn{1}{c}{0.86} & 2.81 & 2.57 & 2.45 \\
 & \multicolumn{1}{c}{ne-en} & \multicolumn{1}{c}{1150} & \multicolumn{1}{c}{26268 / 21434 / 3737} & 0.82 & \multicolumn{1}{c}{0.25} & 0.93 & 0.91 & \multicolumn{1}{c}{0.87} & 2.83 & 2.58 & 2.44 \\
 & \multicolumn{1}{c}{ta-en} & \multicolumn{1}{c}{1047} & \multicolumn{1}{c}{18477 / 23521 / 5570} & 0.81 & \multicolumn{1}{c}{0.23} & 0.97 & 0.98 & \multicolumn{1}{c}{0.94} & 2.76 & 2.57 & 2.44 \\ \midrule
\multirow{4}{*}{\rotatebox{90}{GSM8K}}
 & \multicolumn{1}{c}{zh-en} & \multicolumn{1}{c}{825} & \multicolumn{1}{c}{22244 / 15934 / 3036} & 0.77  & \multicolumn{1}{c}{0.19} & 0.96 & 0.94 & \multicolumn{1}{c}{0.92} & 2.46 & 2.18 & 2.21 \\ 
 & \multicolumn{1}{c}{es-en} & \multicolumn{1}{c}{1231} & \multicolumn{1}{c}{24208 / 26113 / 5902} & 0.86 & \multicolumn{1}{c}{0.34} & 0.98 & 0.97 & \multicolumn{1}{c}{0.95} & 2.44 & 2.20 & 2.19 \\ 
 & \multicolumn{1}{c}{ar-en} & \multicolumn{1}{c}{1141} & \multicolumn{1}{c}{23506 / 20578 / 5229} & 0.84 & \multicolumn{1}{c}{0.23} & 0.96 & 0.94 & \multicolumn{1}{c}{0.92} & 2.26 & 2.12 & 2.09 \\ 
 & \multicolumn{1}{c}{hi-en} & \multicolumn{1}{c}{1170} & \multicolumn{1}{c}{28128 / 22778 / 6285} & 0.80 & \multicolumn{1}{c}{0.26} & 0.96 & 0.93 & \multicolumn{1}{c}{0.91} & 2.51 & 2.26 & 2.26 \\ \midrule
\multirow{4}{*}{\rotatebox{90}{TruthfulQA}}
 & \multicolumn{1}{c}{zh-en} & \multicolumn{1}{c}{771} & \multicolumn{1}{c}{30461 / 15663 / 1589} & 0.72 & \multicolumn{1}{c}{0.20} & 0.97 & 0.94 & \multicolumn{1}{c}{0.92} & 2.80 & 2.36 & 2.42 \\ 
 & \multicolumn{1}{c}{es-en} & \multicolumn{1}{c}{799} & \multicolumn{1}{c}{24467 / 20517 / 1953} & 0.85 & \multicolumn{1}{c}{0.31} & 0.98 & 0.96 & \multicolumn{1}{c}{0.94} & 2.74 & 2.38 & 2.36 \\ 
 & \multicolumn{1}{c}{ar-en} & \multicolumn{1}{c}{795} & \multicolumn{1}{c}{23311 / 16260 / 2810} & 0.82 & \multicolumn{1}{c}{0.24} & 0.97 & 0.95 & \multicolumn{1}{c}{0.93} & 2.77 & 2.43 & 2.35 \\ 
 & \multicolumn{1}{c}{hi-en} & \multicolumn{1}{c}{757} & \multicolumn{1}{c}{28447 / 16764 / 2206} & 0.75 & \multicolumn{1}{c}{0.25} & 0.97 & 0.93 & \multicolumn{1}{c}{0.90} & 2.87 & 2.67 & 2.47 \\ \midrule
\multirow{3}{*}{\rotatebox{90}{MT}}
 & \multicolumn{1}{c}{zh-en} & \multicolumn{1}{c}{850} & \multicolumn{1}{c}{15934 / 9763 / 825} &  0.78 & \multicolumn{1}{c}{0.17} & 0.92 & 0.89 & \multicolumn{1}{c}{0.85} & 2.60 & 2.44 & 2.31 \\ 
 & \multicolumn{1}{c}{es-en} & \multicolumn{1}{c}{1059} & \multicolumn{1}{c}{15047 / 13006 / 1155} & 0.86 & \multicolumn{1}{c}{0.29} & 0.95 & 0.93 & \multicolumn{1}{c}{0.89} & 2.59 & 2.46 & 2.29 \\ 
 & \multicolumn{1}{c}{ar-en} & \multicolumn{1}{c}{802} & \multicolumn{1}{c}{11319 / 8527 / 1063} & 0.85 & \multicolumn{1}{c}{0.22} & 0.93 & 0.91 & \multicolumn{1}{c}{0.87} & 2.50 & 2.37 & 2.19 \\
 \midrule
\multicolumn{2}{c}{Average} & \multicolumn{1}{c}{1016} & \multicolumn{1}{c}{24543 / 19897 / 3330} & 0.81 & \multicolumn{1}{c}{0.25} & 0.96 & 0.94 & \multicolumn{1}{c}{0.91} & 2.72 & 2.46 & 2.38 \\ 
 \bottomrule
\end{tabular}%
}
  \caption{
    \textbf{The statistics of synthesized datasets.}
    The column \( Lang. \) indicates the two languages code-mixed in the dataset.
    The column \( Size \) indicates the size of the dataset.
    The column \( L1 / L2 / Other\) \(tokens \) shows token counts for the first language, the second language, and other language-independent tokens.
    In the column \( Word \)-\( Level \), \( M \) indicates the M-index, and \( I \) indicates the I-index.
    In the column \( Semantic \), \( sim1 \) represents the similarity between code-mixed text and the monolingual text in the first language, \( sim2 \) the similarity with the text in the second language, and \( sim3 \) the similarity between the monolingual texts in the first and second languages. 
    In the \( Model \)-\( Aligned \) column, \( Co. \), \( Na. \), and \( Re. \) denote coherence, naturalness, and readability respectively.
  }
  \label{tab:statistics}
\end{table*}

%% file: tables/one-shot_2.tex
\begin{table*}[htbp]
\centering
\resizebox{0.65\textwidth}{!}{%
\begin{tabular}{@{}lcccc@{}}\toprule
 & \textbf{GPT-3.5-Turbo-Instruct} & \textbf{GPT-3.5-Turbo} & \textbf{GPT-4-Turbo} & \textbf{GPT-4o} \\ \midrule

\multicolumn{5}{c}{\textit{Language Identification (Accuracy)}} \\
zh-en & 89.57 & \textbf{93.38} & 93.35 & 93.31  \\
hok-en & 46.43 & 43.62 & 45.58 & \textbf{58.57}  \\
hi-en & 75.03 & 83.41 & 89.81 & \textbf{89.84}  \\
ne-en & 68.46 & 84.47 & 83.63 & \textbf{85.87}  \\
mr-en & 78.87 & 88.88 & 89.63 & \textbf{92.15}  \\
es-en & 72.26 & 85.28 & 87.47 & \textbf{88.26}  \\
msa-ea & 57.86 & 68.18 & 75.26 & \textbf{76.30}  \\
de-en & 71.27 & \textbf{89.70} & 84.45 & 86.06  \\
fy-nl & 62.02 & 71.11 & \textbf{77.72} & 70.97  \\
gn-es & 76.82 & 85.67 & 89.02 & \textbf{89.99}  \\
Average & 69.86 & 79.37 & 81.59 & \textbf{83.13}  \\ \midrule
\multicolumn{5}{c}{\textit{Part Of Speech (Accuracy)}} \\
zh-en & 71.21 & 74.83 & 76.47 & \textbf{76.91}  \\
hi-en & 70.69 & 70.56 & \textbf{72.23} & 71.70  \\
es-en & 81.68 & 83.02 & \textbf{89.32} & 87.58  \\
fy-nl & 79.84 & 81.73 & 84.39 & \textbf{85.62}  \\
Average & 75.85 & 77.53 & \textbf{80.60} & 80.45 \\ \midrule
\multicolumn{5}{c}{\textit{Named Entity Recognition (F1)}} \\
hi-en & 79.92 & 93.56 & 93.45 & \textbf{93.82}  \\
es-en & 77.12 & \textbf{92.84} & 86.21 & 92.00  \\
msa-ea & 77.95 & 87.70 & \textbf{88.12} & 86.11  \\
gn-es & 86.74 & 91.59 & 94.28 & \textbf{94.51}  \\
Average & 80.43 & 91.42 & 90.51 & \textbf{91.61} \\ \midrule
\multicolumn{5}{c}{\textit{Sentiment Analysis (Accuracy)}} \\
hi-en & 61.46 & 33.78 & \textbf{66.69} & 63.60  \\
bn-en & 62.20 & 53.30 & 69.90 & \textbf{76.70}  \\
mr-en & 54.88 & 32.24 & \textbf{69.52} & 60.56  \\
ne-en & 59.81 & 36.07 & 70.28 & \textbf{71.68}  \\
es-en & 46.21 & 46.21 & \textbf{57.18} & 50.89  \\
ta-en & 51.49 & 38.70 & \textbf{55.10} & 47.65  \\
ml-en & 46.88 & 37.83 & 31.77 & 32.11 \\
Average & 54.71 & 39.73 & \textbf{60.06} & 57.60 \\ \midrule
\multicolumn{5}{c}{\textit{Machine Translation (BLUE)}} \\
zh-en $\rightarrow$ zh & 67.28 & 68.19 & 76.69 & \textbf{79.35}  \\
zh-en $\rightarrow$ en* & 45.47 & 49.00 & \textbf{53.21} & 52.78  \\
hok-zh $\rightarrow$ zh & 52.92 & 50.08 & 60.48 & \textbf{67.95}  \\
hi-en $\rightarrow$ en & 31.08 & 30.68 & 31.17 & \textbf{32.61}  \\
bn-en $\rightarrow$ en & 16.96 & 17.99 & 22.91 & \textbf{23.59}  \\
mr-en $\rightarrow$ en & 13.46 & 14.51 & 18.57 & \textbf{19.84}  \\
es-en $\rightarrow$ en* & 63.38 & 65.94 & 68.20 & \textbf{68.40}  \\
ar-en $\rightarrow$ en* & 54.35 & 57.04 & 61.90 & \textbf{62.35}  \\
Average & 43.11 & 44.18 & 49.14 & \textbf{50.86} \\ \bottomrule

 \end{tabular}%
}
\caption{
\textbf{One-shot evaluation of GPT models on LID, POS, NER, SA and MT.} The \textit{Average} represents the mean score of each model across various datasets from a given task. For each model family, the scores of the top-performing models are highlighted in bold. "*" indicates the datasets we synthesized.
}
\label{tab:one-shot_2}
\end{table*}